\documentclass[journal]{IEEEtran}

\ifCLASSINFOpdf
\else
\fi

\RequirePackage{amsmath}
\usepackage{amsmath,amsfonts,amssymb}
\usepackage{algpseudocode}
\usepackage{algorithm}
\usepackage{algorithmicx}
\usepackage{graphicx}
\usepackage{siunitx}
\usepackage{cite}
\usepackage{textcomp}
\usepackage{gensymb}
\usepackage{bbm}
\usepackage{etoc}
\newcommand{\loss}{\mathcal{L}}
\DeclareMathOperator*{\argmin}{argmin}
\DeclareMathOperator*{\argmax}{argmax}
\usepackage{boldline,multirow}
\usepackage{color}
\usepackage{soul} 
\usepackage[table]{xcolor}
\newcommand{\tabincell}[2]{\begin{tabular}{@{}#1@{}}#2\end{tabular}}
\DeclareMathAlphabet\mathbfcal{OMS}{cmsy}{b}{n}
\usepackage{nameref,zref-xr}
\usepackage[colorlinks=true,citecolor=blue,urlcolor=black]{hyperref}

\begin{document}
%


\title{Joint Optimization of Class-Specific Training- and Test-Time Data Augmentation in Segmentation}

\author{Zeju~Li,
        Konstantinos~Kamnitsas,
        Qi~Dou,
        Chen~Qin
        and~Ben~Glocker
\thanks{Z. Li, K. Kamnitsas and B. Glocker are with the BioMedIA Group, Department of Computing, Imperial College London, SW7 2AZ, United Kingdom. K. Kamnitsas is also with Department of Engineering Science, University of Oxford, OX3 7DQ, United Kingdom and School of Computer Science, University of Birmingham, B15 2TT, United Kingdom. Q. Dou is with Department of Computer Science and Engineering, The Chinese University of Hong Kong, 999077, Hong Kong. C. Qin is with I-X and Department of Electrical and Electronic Engineering, Imperial College London, SW7 2AZ, United Kingdom. E-mail: zeju.li18@imperial.ac.uk.}
}


\maketitle

\begin{abstract}

This paper presents an effective and general data augmentation framework for medical image segmentation. We adopt a computationally efficient and data-efficient gradient-based meta-learning scheme to explicitly align the distribution of training and validation data which is used as a proxy for unseen test data. We improve the current data augmentation strategies with two core designs. First, we learn class-specific training-time data augmentation (TRA) effectively increasing the heterogeneity within the training subsets and tackling the class imbalance common in segmentation. Second, we jointly optimize TRA and test-time data augmentation (TEA), which are closely connected as both aim to align the training and test data distribution but were so far considered separately in previous works. We demonstrate the effectiveness of our method on four medical image segmentation tasks across different scenarios with two state-of-the-art segmentation models, DeepMedic and nnU-Net. Extensive experimentation shows that the proposed data augmentation framework can significantly and consistently improve the segmentation performance when compared to existing solutions. Code is publicly available\footnote{\url{https://github.com/ZerojumpLine/JCSAugment}}.

\end{abstract}


%

\IEEEpeerreviewmaketitle


\section{Introduction}
\label{sec:introduction}

\begin{figure}[t]
\centering
\includegraphics[width=0.48\textwidth]{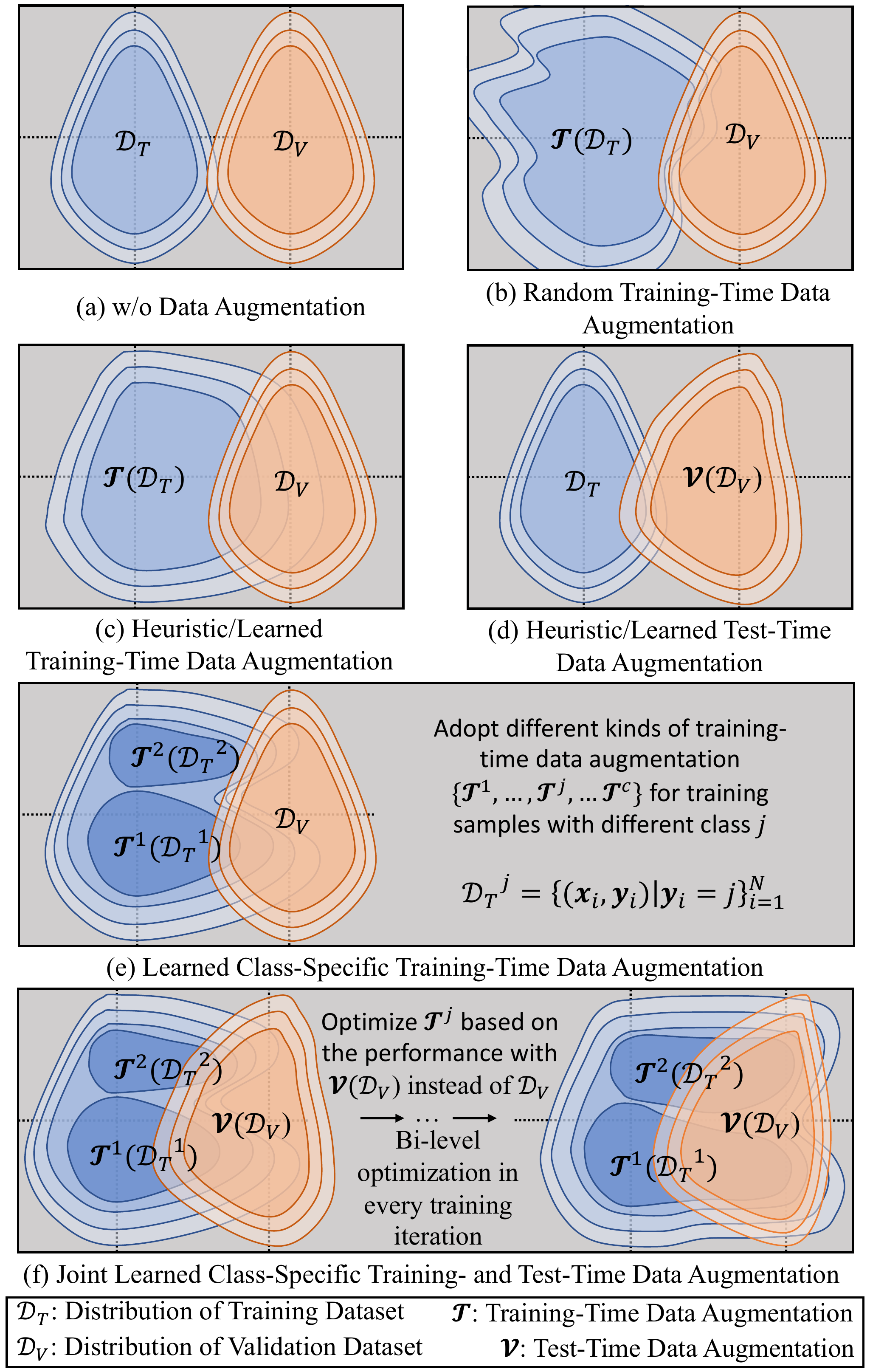}

\caption{Data augmentation improves segmentation model performance by aligning the training and validation/test data distribution. As illustrated in (b)~\cite{cubuk2019randaugment}, (c)\cite{kamnitsas2017efficient, isensee2021nnu,cubuk2018autoaugment,lim2019fast,li2020dada} and (d)~\cite{shanmugam2020and, kim2020learning}, current methods optimize the data distributions by using a validation set as a proxy for unseen test data. Our framework brings improvements by integrating two conceptually simple and intuitive ideas: (e) We adopt different kinds of training-time data augmentation (TRA) for training samples from different classes, effectively extending the training data distribution and alleviating the class imbalance issue. (f) We jointly optimize TRA and test-time data augmentation (TEA) during every training iteration, making the data distributions overlaps more. } \label{fig1}
\end{figure}

\IEEEPARstart{D}{ata} augmentation is a de facto technique in neural networks and has shown to improve model generalization~\cite{van2001art}. It is essential for medical image segmentation algorithms to perform well on unseen test data. Depending on when it is performed, we can divide data augmentation into training-time data augmentation (TRA) and test-time data augmentation (TEA). TRA aims to increase the variation captured by the training dataset by adding perturbed samples with the goal to capture the unseen test data distribution. TEA robustifies the final prediction by averaging predictions of predefined, assumed non-causal variations of test data, to which the model should be robust~\cite{wang2019aleatoric}. An alternative approach for TEA is to modify the test data to achieve higher accuracy with the pretrained model by transforming the test samples to match the distribution of the training data, which is the opposite direction of TRA. Here, we consider to make TRA and TEA complement each other towards the goal of more accurate and robust predictions.

Common data augmentation strategies are usually designed based on heuristics and manually tuned configurations with respect to reducing validation error~\cite{kamnitsas2017efficient, isensee2021nnu}. However, strategies designed for one task may not be optimal for another task or dataset. Consequently, data augmentation without considering data and task characteristics may not always improve the model performance. In particular, different medical image segmentation tasks may require different data augmentation settings, due to changes in image acquisition protocols, modalities, and anatomical structures of interest~\cite{isensee2021nnu}. It is tedious to hand-engineer suitable augmentation strategies for each individual task. Therefore, methods have been proposed to automatically learn effective augmentations directly from the available training data~\cite{cubuk2018autoaugment,lim2019fast,li2020dada, xu2020automatic, yang2019searching, shanmugam2020and, kim2020learning}.

However, we argue that there are two major limitations constraining the performance of current data augmentation strategies. First, previous studies~\cite{cubuk2018autoaugment,lim2019fast, shanmugam2020and, kim2020learning} mostly focus on either TRA or TEA separately, without considering their connections, despite the two being closely linked. This could lead to suboptimal results as the test condition can be adapted through TEA which is not taken into account by TRA when the two are considered in isolation. Second, most TRAs adopt the same transformations for all the samples without considering the different properties existing in different classes. Specifically, the foreground samples in segmentation are more prone to overfitting than background samples because they underrepresented due to class imbalance~\cite{li2020analyzing}. Current data augmentation strategies fail to model the heterogeneity of samples from different classes and the resulting model performance may suffer from overfitting under class imbalance.

In this study, we aim to bridge the gap between TRA and TEA by presenting a gradient-based meta-learning framework to automatically discover optimal TRA and TEA strategies, simultaneously. As illustrated in Fig.~\ref{fig1}(a,b,c,d), data augmentation improves model generalization by aligning the training and underlying test data distribution. Our data augmentation framework (c.f. Fig.~\ref{fig1}(e, f)) further takes class properties and test condition into account, fundamentally restructuring the data distributions aiming for an increased overlap. We validate our method with medical image segmentation because of its imbalanced nature and clinical importance.

The contributions of this study can be summarized as follows: 1) We build a bridge between TRA and TEA through joint optimization of data augmentation policies during the training process, which improves alignment of training and test sample distributions and yields better generalization. 2) We introduce a method that automatically finds different TRA policies for training samples from different classes, implicitly addressing the class imbalance problem. 3) We design a transformation set for TRA with 15 cascaded transformations and 47 operations in total, as well as a transformation set with 83 operations for TEA. These transformation sets cover most transformations in medical image segmentation and can also be easily extended and applied to other applications. 4) Extensive experiments performed on four datasets with two state-of-the-art segmentation models show that our method can consistently improve segmentation performance in various applications and demonstrate the potential to replace the heuristically chosen augmentation policies currently used in most previous works.


\section{Related work}
\label{sec:relatedwork}

\subsection{Data augmentation model}

The majority of data augmentation strategies consist of a set of transformations defined based on domain knowledge to represent the heterogeneity of the test data. Examples include rotations, flipping, and intensity shifts~\cite{krizhevsky2012imagenet, cirecsan2013mitosis}. On the other hand, there are also heuristic perturbation techniques such as cutout~\cite{devries2017improved} and mixup~\cite{zhang2017mixup} that, even though they lead to unrealistic synthetic samples, have been empirically found to improve model generalization. More realistic transformations can be generated based on properties matching~\cite{zhao2019data} or generative adversarial networks~\cite{gupta2019generative}. Although these techniques showed promising performance, the design of data augmentation is difficult because it requires prior knowledge about the task at hand. Optimal strategies, however, may differ significantly between different tasks, datasets and types of input modalities, and thus will be difficult to hand engineer~\cite{cubuk2018autoaugment,isensee2021nnu}. In this study, we aim to automate the process of designing data augmentation.

Currently most TRA methods adopt the same transformations to all the training samples except~\cite{li2020analyzing}, which proposed to increase the variance of foreground samples by heuristically reducing the number of transformed samples for the background classes in order to alleviate class imbalance. However, they found different hyper-parameters are optimal for different datasets, and the chosen transformations and hyper-parameters were based on heuristics. In contrast, our method automatically learns different transformations for different classes and discovers the rules from the training data by itself.

\subsection{Learning based training-time data augmentation}

There have been many attempts to optimize TRA along with the training process to obtain task-specific TRA policies. Most of the studies are developed based on the idea of adversarial training~\cite{goodfellow2014explaining}. The basic adversarial augmentation might not improve generalization on real data as the constructed samples are not realistic. Recent methods attempt to improve real data heterogeneity by adopting an advanced augmentation model~\cite{chen2020realistic} or restricting the search space~\cite{paschali2019data}, which require strong prior knowledge. Different from these, some methods were proposed to generate artificial samples with task constraints~\cite{chaitanya2021semi}, which encourage a generative model to produce additional well-classified images with class properties to enlarge the training data distribution. However, the well-classified samples might not be very useful when the training data is sufficient as they would not make significant changes to the learning of the decision boundary.

Our method is closely related to the line of research which optimizes TRA based on the validation performance such that the learned model can best generalize. Those methods find the sets of augmentation policies that are optimal for a specific training database, out of a pool of possible transformations, based on reinforcement learning~\cite{cubuk2018autoaugment,yang2019searching}, meta-learning~\cite{li2020dada, xu2020automatic}, or density matching~\cite{lim2019fast}. In our study, we consider to learn the parameters of a probability distribution over TRA with a meta-learning scheme. The meta-learner parameters are optimized with the aim of enabling the task segmentation network to perform better on a validation set. In this way, the meta-learner is explicitly trained to select augmentations that improve generalization. Our method improves existing solutions by the joint optimization of TRA and TEA as well as learning a separate augmentation per class. In addition, the defined transformation pool in our work is more comprehensive than previous studies for medical image segmentation, making it more practical to improve upon current heuristic solutions.

\subsection{Learning based test-time data augmentation}
In TEA, class-posterior probabilities from multiple predictions are averaged after applying predefined transformations to the test sample, which was found to be effective to improve accuracy. Recently, some methods were proposed to learn TEA by choosing the transformations obtaining low loss values on the validation set based on a pre-trained model~\cite{shanmugam2020and, kim2020learning}. Test-time adaptation is another kind of learning based TEA where the pre-trained model is adapted to fit a single test sample based on denoising autoencoder~\cite{karani2021test} or self-supervision~\cite{he2021autoencoder}. These learning based TEA strategies can be seen as a post-processing to the segmentation and do not contribute to the learning of the model. In contrast, our method proposes to combine the optimization of TRA and TEA during training, which leads to not only learning the optimal TRA and TEA transformations that complement each other, but also learning optimal model parameters given the specific set of data transformations.

\section{Method}
\label{sec:method}

\begin{figure*}[t]
\centering
\includegraphics[width=\textwidth]{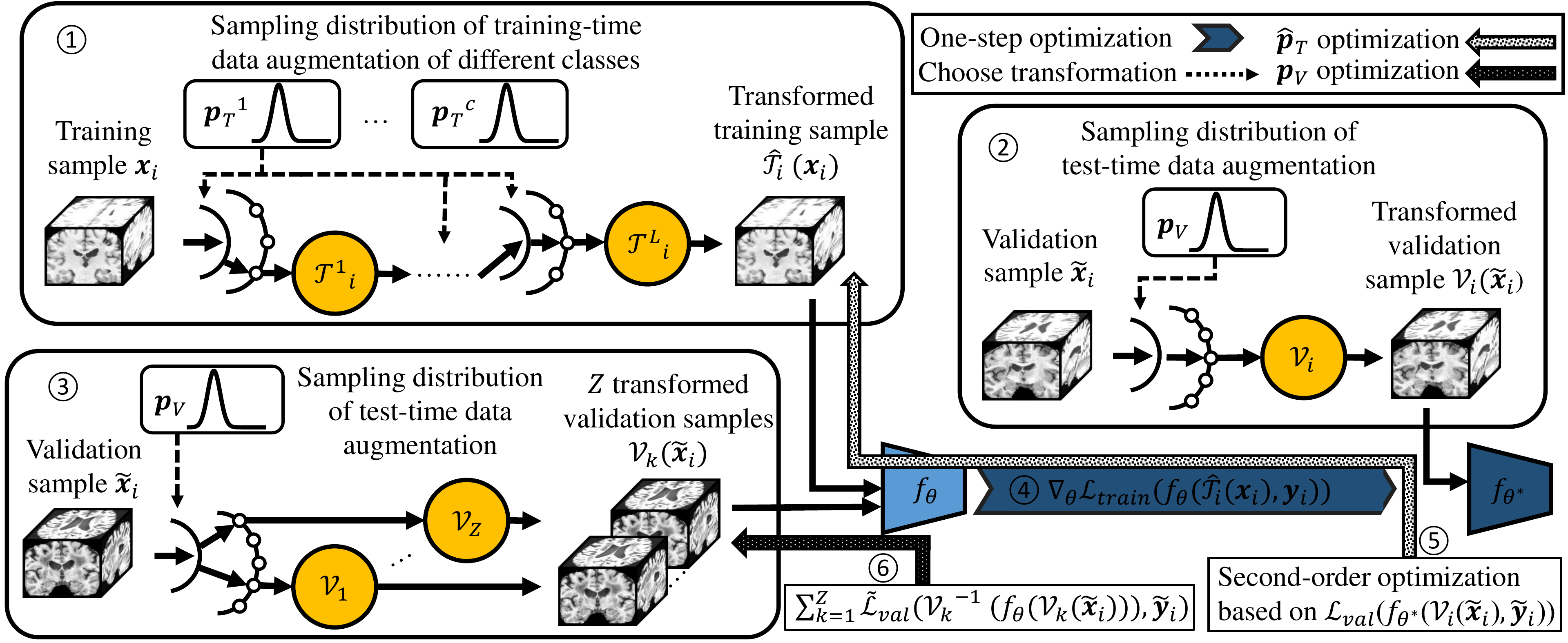}

\caption{The optimization process of the proposed method. In this study, data augmentation is formulated as the probability distribution of multiple predefined transformations, as demonstrated in \textcircled{1}, \textcircled{2} and \textcircled{3}. During the same iteration, class-specific TRA is optimized based on meta-gradients with \textcircled{5} while TEA is optimized based on the validation losses of $Z$ transformed samples with \textcircled{6}.} \label{fig2}
\end{figure*}

\subsection{Preliminaries}

We consider the image segmentation problem with $c$ total number of classes. A training dataset $\mathcal{D}_T = \{ (\boldsymbol{x}_i, \boldsymbol{y}_i) \}_{i = 1}^N$ with $N$ samples is given, where $\boldsymbol{x}_i$ is a training image and $\boldsymbol{y}_i$ corresponds to the segmentation label map with individual labels $y_{ip} \in \{1, ..., c\}$ for each image pixel $p$. Assuming a segmenter $f_\theta$ parameterized by $\theta$, our aim is to learn optimal $\theta^*$ parameters, such that $f_{\theta^*}(\cdot$) minimizes the empirical risk over the training data. For any training loss function $\mathcal{L}_{train}$, the empirical risk of the segmentation model $f_\theta$ is defined as $R_{\mathcal{L}_{train}}(f_\theta)=\frac{1}{N}\sum_{i=1}^{N}\mathcal{L}_{train}(f_\theta(\boldsymbol{x}_i), \boldsymbol{y}_i)$. Apart from $\mathcal{D}_T$, we usually have a validation dataset $\mathcal{D}_V = \{ (\tilde{\boldsymbol{x}}_i, \tilde{\boldsymbol{y}}_i) \}_{i = 1}^M$ with $M$ samples along with a validation loss $\mathcal{L}_{val}$, which is taken as a proxy for unseen test data and used to tune the hyper-parameters including learning rates~\cite{li2017hyperband}, network architecture~\cite{zoph2016neural} and data augmentation policies~\cite{cubuk2018autoaugment}. Note that $\mathcal{D}_V$ could come from a different distribution from $\mathcal{D}_T$, based on different assumptions of unseen test data.

\subsection{Sampling transformations}

For a sample $\boldsymbol{x}_i$ (or $\tilde{\boldsymbol{x}}_i$), we will apply transformation $\mathcal{T}_i(\cdot)$ which is specific to the $i$-th sample. $\mathcal{T}_i$ is obtained by sampling from a set of $K$ operations $\{\mathcal{O}^1, ...,\mathcal{O}^K\}$ based on the corresponding probability distribution $\boldsymbol{p}$ = $[p_1, ..., p_K]^\intercal$. In this study we represent with a different $\mathcal{O}^j$ operation not only transformations of different type (for example rotations, contrast enhancement, etc.) but also transformations of the same type but different magnitudes (for example rotations of different degree). We do not further optimize the predefined magnitudes of transformations during training.
Our method will optimize during training the sampling distribution $\boldsymbol{p}$ of different transformations for data augmentation, so that we learn which transformations are most appropriate for the given dataset and task.

In order to include the distribution into the gradient based optimization through the non-differentiable sampling process, we reparameterize the categorical distribution using the Gumbel-Softmax trick~\cite{jang2016categorical}. We calculate the probability of assigning sample $\boldsymbol{x}_i$ (or $\tilde{\boldsymbol{x}}_i$) with operation $\mathcal{O}^j$ as:

\begin{equation}
    s_{ij} = \frac{\mathrm{e}^{p_j+g_{ij}}}
    {\sum_{v=1}^K \mathrm{e}^{p_v+g_{iv}}}, \qquad \text{for } j=1, ..., K,
    \label{eq:func1}
\end{equation}

where $g_{ij}$ is a sample drawn from the Gumbel distribution, i.e., $g_{ij}$ = --$\log$(--$\log$($\varepsilon$)), in which $\varepsilon$ is a random number by drawing $\varepsilon$ $\sim$ Uniform(0, 1). It then holds that $\sum_{j=1}^{K}s_{ij}=1$ and $0\leq s_{ij} \leq 1$, $\forall j$. In this way, the stochasticity involved in the sampling process is removed from the computational graph of network's training and the process of choosing augmentation $\mathcal{T}_i$ based on probability distribution $\boldsymbol{p}$ now becomes differentiable. Specifically, $\mathcal{T}_i$ is chosen as $\mathcal{O}^{j^*}$ where $j^*$ = $\argmax_{j}(s_{ij})$. The sampling probability $\boldsymbol{p}$, which we would like to optimize, can   still not be updated via backpropagation, both due to the non differentiable $\argmax$ and because the transformations are non-differentiable in the general case. To work around this, we also calculate a weight $w_i$ that corresponds to the sample $\boldsymbol{x}_i$ (or $\tilde{\boldsymbol{x}}_i$) with:

\begin{equation}
    w_i=\max_{j}(s_{ij})+\underbrace{(1-\max_{j}(s_{ij}))}_{\text{does not require gradient}},
    \label{eq:wcalculate}
\end{equation}

which is a function of the sampling probability $s_{ij}$. We then incorporate the weight into the empirical risk as $R^{\prime}_{\mathcal{L}_{train}}(f_\theta) = \frac{1}{N}\sum_{i=1}^{N}w_i\mathcal{L}_{train}(f_\theta(\mathcal{T}_i(\boldsymbol{x}_i)), \boldsymbol{y}_i)$. In this manner, $w_i$ and $s_{ij}$ are part of the total loss and hence can be straightforwardly optimized. During the forward propagation, we utilize $w_i$ to evaluate the chosen transformation $\mathcal{T}_i$ without affecting the training procedure, as $w_i$ is always equal to 1. During backpropagation, $w_i$ that is associated with a relatively effective $\mathcal{T}_i$ is prone to be increased. As we enforce the computation of the second term in Eq.~\ref{eq:wcalculate} to never require gradient, we can use $w_i$ as a means to optimize $s_{ij}$ and thus $\boldsymbol{p}$ with gradient descend.

For distinction between TRA and TEA, in the following paragraphs we define the probability distribution and transformation of TRA as $\boldsymbol{p}_T$ and $\hat{\mathcal{T}}_i$ while denoting the ones of TEA as $\boldsymbol{p}_V$ and $\mathcal{V}_i$ unless otherwise noted. Note that we are considering the problem of image segmentation, therefore the spatial transformations are always applied to $\boldsymbol{y}_i$ simultaneously but we omit this for simplicity.

\subsection{Overview of the training process}

We aim to reduce the generalization gap explicitly by optimizing a probability distribution over data augmentations $\boldsymbol{p}_T$ (or $\boldsymbol{p}_V$) based on the gradient from the validation data $\nabla_{\boldsymbol{p}_{T}}\mathcal{L}_{val}$ (or $\nabla_{\boldsymbol{p}_{V}}\mathcal{L}_{val}$). Thus, $\boldsymbol{p}_T$ (or $\boldsymbol{p}_V$) is automatically adapted to the underlying task-specific characteristics.

We develop a training framework based on meta-learning via second-order optimization to accomplish this. The optimization process of the proposed method is illustrated in Fig.~\ref{fig2}. During the sampling process, we first obtain the transformed training data with \textcircled{1} and optimize the model to $f_{\theta^*}$ with \textcircled{4} based on a single optimization step; then we pass the transformed validation data with \textcircled{2} through $f_{\theta^*}$ to compute the second-order gradients \textcircled{5} and backprop to \textcircled{1}, to learn TRA that leads to learning $\theta^*$ that best generalizes on validation data transformed with TEA; meanwhile, we also apply varied transformations of TEA to a single validation sample with \textcircled{3} and we try to learn TEA $\boldsymbol{p}_V$ which can transform a single validation sample $\tilde{\boldsymbol{x}}_i$ to have the lowest validation error with \textcircled{6}.

\begin{algorithm}
\small{
\caption{Joint Optimization of Class-Specific Training- and Test-Time Data Augmentation in Segmentation}
\label{alg:AugmentDistribution}
\begin{algorithmic}[1]
\Require
      \Statex $\mathcal{D}_T = \{ (\boldsymbol{x}_i, \boldsymbol{y}_i) \}_{i = 1}^N$: training data, $\mathcal{D}_V = \{ (\tilde{\boldsymbol{x}}_i, \tilde{\boldsymbol{y}}_i) \}_{i = 1}^M$: validation data; $f_\theta$($\cdot$): the segmentation model, $\hat{\mathcal{T}}_i$($\cdot$): TRA which is determined by drawing from class-specific probability $\hat{\boldsymbol{p}}_T$, $\mathcal{V}_i$($\cdot$): TEA which is determined by drawing from probability $\boldsymbol{p}_V$.
      \Statex $\alpha$, $\beta$, $\gamma$: learning rate to update $\theta$, $\hat{\boldsymbol{p}}_T$ and $\boldsymbol{p}_V$.
\State Initialize $\hat{\boldsymbol{p}}_T$, $\boldsymbol{p}_V$ with heuristic policies referring to the ones in DeepMedic~\cite{kamnitsas2017efficient} or nnU-Net~\cite{isensee2021nnu}.
\For{each iteration} 

\State Sample a batch of training data $\mathcal{B}_T = \{ (\boldsymbol{x}_i, \boldsymbol{y}_i) \}_{i = 1}^n$ from $\mathcal{D}_T$ and a batch of validation data $\mathcal{B}_V = \{(\tilde{\boldsymbol{x}}_i, \tilde{\boldsymbol{y}}_i) \}_{i = 1}^m$ from $\mathcal{D}_V$.

\For{a number of steps} \Comment{\emph{\small{Note: One step is sufficient in our experiments.}}}

\State Sample a set of $\{\hat{\mathcal{T}}_i(\cdot)\}_{i = 1}^n$ with Gumbel-Softmax distribution parameterized by ${\boldsymbol{p}_T}^j$ based on sample class.

\State Sample $\{\mathcal{V}_i(\cdot)\}_{i = 1}^m$ and $\{\mathcal{V}_k(\cdot)\}_{k = 1}^Z$ with Gumbel-Softmax distribution parameterized by $\boldsymbol{p}_V$.

\State Calculate $\theta^*$ with an optimization step via Eq.~\ref{eq:func2}. 

\State Optimize $\hat{\boldsymbol{p}}_T$ based on normalized meta-gradients: $\hat{\boldsymbol{p}}_T^{t+1} = \hat{\boldsymbol{p}}_T^{t} - \beta \nabla_{\hat{\boldsymbol{p}}_T^{t}} \frac{1}{m}\sum_{i=1}^{m} \mathcal{L}_{val}( f_{\theta^*}(\mathcal{V}_i(\tilde{\boldsymbol{x}}_i)),\tilde{\boldsymbol{y}}_i)$. \Comment{\emph{\small{Learning of TRA.}}}

\State Optimize $\boldsymbol{p}_V$ based on normalized gradient via Eq.~\ref{eq:tslearning}. \Comment{\emph{\small{Learning of TEA.}}}

\EndFor

\State Update $\theta$ to $\theta^*$. \Comment{\emph{\small{Training the segmentation model.}}}

\EndFor
\end{algorithmic}
}
\end{algorithm}

\subsection{Learning of class-specific training-time data augmentation}

\subsubsection{The design of predefined transformations}

Following the design of data augmentation in many medical image segmentation frameworks~\cite{kamnitsas2017efficient, isensee2021nnu}, we design the transformation set with $L$=15 cascaded operations including rotation, mirroring, gamma correction, histogram transformations, blurring, sharpening, adding noise, and simulating low resolution. The operation magnitudes are decided by uniformly sampling from predefined ranges. We summarize the detailed information about the operations in supplementary material. Specifically, the probability distribution and transformations of TRA is extended as $\boldsymbol{p}_T=(\boldsymbol{p}^1, ..., \boldsymbol{p}^L)$ and $\hat{\mathcal{T}}_i=\{{\mathcal{T}^1}_i, ..., {\mathcal{T}^L}_i\}$.

We ensure that our design of TRA is able to accomplish the same functionality with the built-in data augmentation in prevailing frameworks such as DeepMedic~\cite{kamnitsas2017efficient} and nnU-Net~\cite{isensee2021nnu}, therefore our method can act as a replacement for heuristic TRA. We initialize $\boldsymbol{p}_T$ with heuristic policies provided by these frameworks, as shown in Fig.~\ref{fig_policy}.

\subsubsection{Class-specific data augmentation}

We adopt different TRAs for training samples from different classes. Specifically, we extend the probability distribution to $\hat{\boldsymbol{p}}_T=({\boldsymbol{p}_T}^1, ..., {\boldsymbol{p}_T}^c)$ which contains different probability distributions for $c$ classes. In this way, TRA becomes more flexible and powerful as it gains the ability to draw $\hat{\mathcal{T}}_i$ from different distributions for different classes.

In practice, we determine the class of a training patch with the central pixel of the patch. Note that in this study we only regard the training samples to come from 2 classes consisting of foreground (tumor, lesion, and organs) and background.

\subsubsection{Policy optimization with meta-gradients}

Similar to previous works on learning TRA~\cite{cubuk2018autoaugment, li2020dada}, we aim to learn TRA based on the performance of validation data and formulate the optimization of TRA as a bi-level optimization problem:

\begin{align}
    &\min_{\hat{\boldsymbol{p}}_T}\frac{1}{M} \sum_{i=1}^{M}\mathcal{L}_{val}(f_{\theta^*}(\tilde{\boldsymbol{x}}_i), \tilde{\boldsymbol{y}}_i) \label{eqn:objectivefunA} \\
    &s.t.\quad \theta^* = \argmin_{\theta}\frac{1}{N} \sum_{i=1}^{N}w_i\mathcal{L}_{train}(f_\theta(\hat{\mathcal{T}}_i(\boldsymbol{x}_i)), \boldsymbol{y}_i).
    \label{eq:objectivefunB}
\end{align}

We propose to solve this based on gradient descent following~\cite{pedregosa2016hyperparameter, franceschi2018bilevel}. We train the model with a training batch containing $n$ samples and a validation batch consisting of $m$ samples. For simplicity, we shorten $\frac{1}{n}\sum_{i=1}^{n}w_i\mathcal{L}_{train}(f_\theta(\hat{\mathcal{T}}_i(\boldsymbol{x}_i), \boldsymbol{y}_i))$ as $\mathcal{L}_{train}(\theta, \hat{\boldsymbol{p}}_T)$ and $\frac{1}{m}\sum_{i=1}^{m}\mathcal{L}_{val}( f_{\theta^*}(\tilde{\boldsymbol{x}}_i),\tilde{\boldsymbol{y}}_i))$ as $\mathcal{L}_{val}(\theta^*)$ in the following paragraphs. Based on the chain rule, the gradient of validation loss w.r.t. $\hat{\boldsymbol{p}}_T$ is derived as:

\begin{equation}
     \nabla_{\hat{\boldsymbol{p}}_T}\mathcal{L}_{val}(\theta^*) = (\frac{\partial \theta^*}{\partial \hat{\boldsymbol{p}}_T})^\intercal \nabla_{\theta}\mathcal{L}_{val}(\theta^*),
    \label{eq:funcchain}
\end{equation}

where $\nabla_{\hat{\boldsymbol{p}}_T} = (\frac{\partial}{\partial \hat{\boldsymbol{p}}_T})^\intercal$ and $\nabla_{\theta} = (\frac{\partial}{\partial \theta})^\intercal$. The calculation of $\frac{\partial \theta^*}{\partial \hat{\boldsymbol{p}}_T}$ can be derived based on implicit function theorem~\cite{bengio2000gradient}. However, the calculation would introduce a Hessian which is not practical to calculate with the parameters of deep nerual network as the number of parameters is too large. There are many methods to approximate the gradient without Hessian calculation~\cite{pedregosa2016hyperparameter, franceschi2018bilevel, liu2018darts}, in this study we choose to approximate $\theta^*$ by using a single training step~\cite{finn2017model, liu2018darts}. Specifically, we approximate the optimal $\theta^*$ via a standard training step with: 

\begin{equation}
    \theta^* \approx \theta-\alpha \nabla_\theta \mathcal{L}_{train}(\theta, \hat{\boldsymbol{p}}_T).
    \label{eq:func2}
\end{equation}

Here, $\alpha$ is the step length which we set equal to the learning rate of the task model. Eq.~\ref{eq:func2} defines the approximated optimal $\theta^*$ when trained using the training data with sampled data augmentation $\hat{\mathcal{T}}_i$. In this manner, we can evaluate the effectiveness of the data augmentation policy $\hat{\boldsymbol{p}}_T$ based on the performance of the updated model $f_{\theta^*}$ on a held-out validation dataset. We differentiate this equation w.r.t. $\hat{\boldsymbol{p}}_T$ from both sides and yield:

\begin{equation}
    \frac{\partial \theta^*}{\partial \hat{\boldsymbol{p}}_T} = -\alpha\nabla^2_{\theta,\hat{\boldsymbol{p}}_T}\mathcal{L}_{train}(\theta^*, \hat{\boldsymbol{p}}_T),
    \label{eq:omegasimple}
\end{equation}

where $\nabla^2_{\theta,\hat{\boldsymbol{p}}_T} = \frac{\partial \nabla_{\theta}}{ \partial \hat{\boldsymbol{p}}_T}$. By substituting Eq.~\ref{eq:omegasimple} into Eq.~\ref{eq:funcchain}, now we can update $\hat{\boldsymbol{p}}_T$ with:

\begin{equation}
\begin{split}
    & \hat{\boldsymbol{p}}_T^{t+1} = \hat{\boldsymbol{p}}_T^t - \beta \nabla_{\hat{\boldsymbol{p}}_T^t} \mathcal{L}_{val}(\theta^*) \\
    & = \hat{\boldsymbol{p}}_T^t + \alpha\beta\nabla^2_{\hat{\boldsymbol{p}}_T^t,\theta}\mathcal{L}_{train}(\theta^*, \hat{\boldsymbol{p}}_T^{t}) \nabla_{\theta}\mathcal{L}_{val}(\theta^*),
    \label{eq:func3}
\end{split}
\end{equation}

which can be interpreted as a gradient of the gradient from the task-driven training. In the above, $\beta$ is the learning rate for determining the probability distribution. In this way, we can optimize the distribution $\hat{\boldsymbol{p}}_T$ explicitly with the aim to improve generalization of the segmentation model using the validation data. Eq.~\ref{eq:func3} includes a second-order gradient. As the distribution $\hat{\boldsymbol{p}}_T$ is represented with only a few parameters ($K$, which is in the order of 10-100), we find the complexity of the gradient computation to be $O(| \hat{\boldsymbol{p}}_T||\theta|)$ which is feasible and can be handled by prevailing toolboxes such as PyTorch and Tensorflow. 

After updating $\hat{\boldsymbol{p}}_T$, we update $f_\theta$ to $f_{\theta^*}$ to fit the updated TRA policy for the next iteration. We optimize $\hat{\boldsymbol{p}}_T$ along with the training of the task model, and at the end of the training we may have higher performance than any model learned with a random or manually configured augmentation policy.

\subsubsection{Gradient normalization}

We normalize the gradient from different classes of training samples, as we notice that the contributions of training samples from different classes to the reduction of the validation loss varied a lot. For example, the foreground samples are more effective for reducing the validation loss, resulting in increased probabilities of the policies associated with the foreground samples. Specifically, if we rewrite the gradient in Eq.~\ref{eq:funcchain} by the chain rule as:

\begin{equation}
    \nabla_{\hat{\boldsymbol{p}}_T} \mathcal{L}_{val}(\theta^*) = \sum_{i=1}^{n} \nabla_{w_i} \mathcal{L}_{val}(\theta^*)(\frac{\partial w_i}{\partial \hat{\boldsymbol{p}}_T})^\intercal,
    \label{eq:funcbeforenormalize}
\end{equation}

we would find the magnitude of $\nabla_{w_i} \mathcal{L}_{val}(\theta^*)$ is significantly larger for the foreground samples than the background samples. To resolve this, we rewrite Eq.~\ref{eq:funcbeforenormalize} as $\nabla_{\hat{\boldsymbol{p}}_T} \mathcal{L}_{val}(\theta^*) = \sum_{i=1}^{n} h_i (\frac{\partial w_i}{\partial \hat{\boldsymbol{p}}_T})^\intercal$ with the normalized gradient $h_i$:

\begin{equation}
    h_i = \nabla_{w_i} \mathcal{L}_{val}(\theta^*) - \frac{\sum_{j=1}^{n} \mathbbm{1}_{[y_{jc}=y_{ic}]}\nabla_{w_j} \mathcal{L}_{val}(\theta^*)}{\sum_{j=1}^{n} \mathbbm{1}_{[y_{jc}=y_{ic}]}} 
\end{equation}

where $y_{ic}$ is the central pixel label of the segmentation label map $\boldsymbol{y}_i$ and $\mathbbm{1}_{y_{jc}=y_{ic}} \in \{0,1\} $ is an indicator function which is equal to 1 if and only if $y_{jc}=y_{ic}$. Thus, the gradients are normalized for different classes. Another benefit of gradient normalization is that we can guarantee that the probability of transformation which is not sampled in one iteration would remain unchanged. 

\subsubsection{Sampling normalization}

We notice that the optimization process could also be biased towards transformations with high probability. Because the more frequently one transformation is sampled, its probability would be increased more as long as it is more effective than the majority of the transformations in the same batch. As a consequence, it would be likely to be trapped in a local minimum and the probability of any preferable transformation which is frequently sampled by chance could be increased a lot. Therefore, we also normalize $h_i$ with the sampling frequency and obtain $\hat{h}_i$ as:

\begin{equation}
    \hat{h}_i=\frac{h_i}{\sum_{v=1}^{n} \mathbbm{1}_{[\argmax_{j}(s_{vj})=\argmax_{j}(s_{ij})]}}. \\
    \label{eq:samplenormalize}
\end{equation}

\subsection{Learning of test-time data augmentation}
\label{sec:learnTSA}

\subsubsection{The design of predefined transformations}

We design the transformation set for TEA with $K$=84 kinds of deterministic operations including identity along with 41 spatial transformations, 30 intensity transformations, and 12 noise transformations. We summarize the detailed information about those transformations in supplementary material.
We initialize $\boldsymbol{p}_V$ referring to the heuristic policies used in nnU-Net~\cite{isensee2021nnu} which comprises mirroring and \ang{180} rotation in three directions, as shown in Fig.~\ref{fig_policy}.

\subsubsection{Policy optimization based on reverted predictions}

The optimization of TEA is straightforward. We aim to optimize the function with normal gradient descend:

\begin{equation}
    \min_{\boldsymbol{p}_V}\frac{1}{Z}\sum_{k=1}^{Z}\tilde{w}_k\tilde{\mathcal{L}}_{val}({\mathcal{V}_k}^{-1}(f_{\theta}(\mathcal{V}_i(\tilde{\boldsymbol{x}}_i))), \tilde{\boldsymbol{y}}_i), \\
    \label{eq:teaoptimization}
\end{equation}

where $Z$ is the number of samples TEA transformations in a batch. We update $\boldsymbol{p}_V$ by choosing the transformations which have the lowest validation loss with the same validation sample:

\begin{equation}
    \boldsymbol{p}_V^{t+1} = \boldsymbol{p}_V^{t} - \gamma \nabla_{\boldsymbol{p}_V}\frac{1}{Z} \sum_{k=1}^{Z}\tilde{w}_k\tilde{\mathcal{L}}_{val}({\mathcal{V}_k}^{-1}(f_\theta(\mathcal{V}_k(\tilde{\boldsymbol{x}}_i))), \tilde{\boldsymbol{y}}_i),
    \label{eq:tslearning}
\end{equation}

where $\gamma$ is the learning rate to update the probability and $\tilde{\mathcal{L}}_{val}$ is the validation loss function for TEA optimization, which can differ from $\mathcal{L}_{val}$.

\subsubsection{Sampling normalization}

Similar to TRA, we notice that the optimization of $\boldsymbol{p}_V$ would be biased due to the sampling results. We simplify $\frac{1}{Z} \sum_{k=1}^{Z}\tilde{w}_k\tilde{\mathcal{L}}_{val}({\mathcal{V}_k}^{-1}(f_\theta(\mathcal{V}_k(\tilde{\boldsymbol{x}}_i))), \tilde{\boldsymbol{y}}_i)$ to $\tilde{\mathcal{L}}_{val}(\theta)$, and derive the gradient based on the chain rule:

\begin{equation}
    \nabla_{\boldsymbol{p}_V}\tilde{\mathcal{L}}_{val}(\theta)=\sum_{k=1}^{Z} \tilde{h}_k(\frac{\partial \tilde{w}_k}{\partial \boldsymbol{p}_V})^\intercal.
    \label{eq:tslearningsimpl}
\end{equation}

Similarly, we normalize the gradients based on sampling frequency and calculate $\tilde{h}_k$ as:

\begin{equation}
    \tilde{h}_k = \frac{\nabla_{\tilde{w}_k}\tilde{\mathcal{L}}_{val}(\theta)}{\sum_{v=1}^{Z} \mathbbm{1}_{[\argmax_{j}(s_{vj})=\argmax_{j}(s_{kj})]}}.
    \label{eq:tslearningnormalize}
\end{equation}

\subsection{Inference of test-time data augmentation}

Given unseen test data, we adopt the learned TEA policy to transform the image. Specifically, in order to simplify the inference process, we do TEA at test time with the weighted sum of operations that have the highest $z$ probability, where $z$ is a hyper-parameter indicating the number of operations to be selected for aggregation. We choose $z$=8 for 3D U-Net while $z$=4 for DeepMedic. The weight of operation $\mathcal{O}^{j}$ is set as the corresponding sampling probability which is calculated as $\mathrm{e}^{p_j} / \sum_{v=1}^K \mathrm{e}^{p_v}$.

\subsection{Joint learning of training- and test-time data augmentation}

We propose to jointly optimize TRA and TEA, and specifically we optimize $\hat{\boldsymbol{p}}_T$ with the transformed validation data based on $\boldsymbol{p}_V^{t}$ and rewrite Eq.~\ref{eqn:objectivefunA} and \ref{eq:objectivefunB} as:

\begin{align}
    &\min_{\hat{\boldsymbol{p}}_T}\frac{1}{M}\sum_{i=1}^{M}\mathcal{L}_{val}(f_{\theta^*}(\mathcal{V}_i(\tilde{\boldsymbol{x}}_i)), \tilde{\boldsymbol{y}}_i) \label{eqn:objectivefunAs} \\
    &s.t.\quad \theta^* = \argmin_{\theta}\frac{1}{N} \sum_{i=1}^{N}w_i\mathcal{L}_{train}(f_\theta(\hat{\mathcal{T}}_i(\boldsymbol{x}_i)), \boldsymbol{y}_i).
    \label{eq:objectivefunBs}
\end{align}

Note that we optimize both $\hat{\boldsymbol{p}}_T$ and $\boldsymbol{p}_V$ in one training iteration. Bridging the optimization process of $\hat{\boldsymbol{p}}_T$ and $\boldsymbol{p}_V$ has two advantages: First, we can reduce the risk of overfitting on the validation data as it is extended with augmented samples. Second, the model can generalize well to the transformations we adopt at test time. The full procedure is summarized in Algorithm~\ref{alg:AugmentDistribution}.

Additional implementation details are provided in the supplementary material. We find that the policies do not need to be updated in every iteration, making training more efficient. In practice, we observe the training time would only increase by about 20\% compared to standard training. Typically, when a model takes 4 days to train with an NVIDIA 1080TI GPU for the segmentation task, our method costs 20 hours to find the optimal data augmentation strategies. This is computational efficient than AutoAugment~\cite{cubuk2018autoaugment} which could takes thousands of hours.

\section{Experiments, Results, and Discussion}
\label{sec:experiments}

\subsection{Experimental setup}

\subsubsection{Data pre-processing}
We normalize all datasets using the pipeline of nnU-Net. Specifically,  we adopt case-wise Z-score normalization for magnetic resonance (MR) images, and we normalize computed tomography (CT) images with dataset-wise Z-score normalization based on foreground samples after clipping the Hounsfield units (HU) values from 0.5\% to 99.5\%.

\subsubsection{Network configurations}
Our experiments are performed with DeepMedic~\cite{kamnitsas2017efficient} and a well configured 3D U-Net~\cite{isensee2021nnu}. We choose cross-entropy (CE) as $\mathcal{L}_{train}$ for DeepMedic and an equal combination of CE and soft Dice similarity coefficient (DSC) for 3D U-Net. We find that the optimal choice of $\mathcal{L}_{val}$ varies for different datasets and summarize the information in supplementary material. We adopt soft DSC for $\tilde{\mathcal{L}}_{val}$ for all the experiments. We choose the batch sizes $n$ and $m$ to be 10. We set the primary patch size as 37$\times$37$\times$37 for all the experiments with DeepMedic and a patch size of 64$\times$64$\times$64 for all the applications with 3D U-Net except prostate segmentation. We choose a patch size of 64$\times$64$\times$32 for prostate segmentation with 3D U-Net because images in this dataset have fewer slices. We train the networks for 1,000 epochs except for kidney and kidney tumor segmentation where we train for 2,000 epochs,  as we observed that the networks need more iterations to converge on this task. All the reported results are the average of two runs with different random seeds.

\subsubsection{Brain stroke lesion segmentation}

We firstly evaluate the proposed method with binary brain stroke lesion segmentation using the dataset of Anatomical Tracings of Lesions After Stroke (ATLAS)~\cite{liew2018large}. The images have a voxel spacing of 1.0$\times$1.0$\times$1.0 mm. With a total of 220 T1-weighted MR images, we randomly select 73 (50\%) or 145 (100\%) for training, 31 for validation, and 44 for test.

\subsubsection{Kidney and kidney tumor segmentation}

Secondly, we evaluate the proposed method with kidney and kidney tumor segmentation using the training dataset of Kidney Tumor Segmentation Challenge (KiTS)~\cite{heller2019kits19} which contains 210 CT images. We resample all images to voxel spacing of 1.6$\times$1.6$\times$3.2 mm. We randomly select 70 (50\%) or 140 (100\%) for training, 28 for validation, and 42 for test. We omit the segmentation results of kidney as find most methods perform well (DSC $>$ 95.0) on the task of kidney segmentation.

\subsubsection{Abdominal organ segmentation}

We also evaluate the proposed method with the task of abdominal organ segmentation~\cite{xu2015efficient} which contains 14 classes including spleen (SP), right kidney (RK), left kidney (LK), gallbladder (GB), esophagus (E), liver (LIV), stomach (STO), aorta (AO), inferior vena cava (IVC), portal vein and splenic vein (V), pancreas (PA), right adrenal gland (RA) and left adrenal gland (LA). We resample all images to a voxel spacing of 1.6$\times$1.6$\times$3.2 mm. We randomly select 20 for training, 4 for validation, and 6 for test. 

\subsubsection{Cross-site prostate segmentation}

Additionally, we utilize our method to align training data and validation data of prostate segmentation from different domains~\cite{liu2020saml}. Specifically, we utilize 30 T2-weighted MR images from site A~\cite{bloch2015nci} which were collected with 1.5T Philips MRI machine with endorectal coil and 19 T2-weighted MR images from site B~\cite{lemaitre2015computer} which were collected with 3T Siemens MRI machines without endorectal coil. We resample all the images to a voxel spacing of 0.8$\times$0.8$\times$1.5 mm. We investigate the scenario where the target domain (site B) has limited labeled data. We select 20 cases from site A for training and 6 cases for test. We select 1 case from site B for validation and use 18 cases for testing. Note that for cross-site prostate segmentation, we report results with models trained with both training data and validation data as this serves as a fairer baseline compared to using training data only.

\subsection{Compared methods}

\subsubsection{Heuristic}

We compare with the heuristic TRA and TEA which are set as the default configurations in DeepMedic~\cite{kamnitsas2017efficient} and nnU-Net~\cite{isensee2021nnu}. We also report a few results based on models trained using both the training and validation data with heuristic TRA.

\subsubsection{Learned TRA}
We compare with methods that adopt the data augmentation policies based on the validation performance without considering class dependency~\cite{cubuk2018autoaugment, lim2019fast, li2020dada}.

\subsubsection{TRA with different transformation magnitudes}
We also compare with RandAugment~\cite{cubuk2019randaugment}, which only changes the transformation magnitudes based on grid searching. Specifically, we keep the data augmentation probability and replace the operations of the same type with different magnitudes, yielding RandAugment-S, RandAugment-M and RandAugment-L. We summarize the results with RandAugment in supplementary material. 

\subsubsection{Learned TEA}
We compare with methods which optimize TEA based on a pretrained segmentation model~\cite{shanmugam2020and, kim2020learning}. Specifically, after training the model with proposed TRA, we refine TEA as described in Section~\ref{sec:learnTSA}.

\subsection{Quantitative results}

\begin{figure*}[t]
\centering
\includegraphics[width=\textwidth]{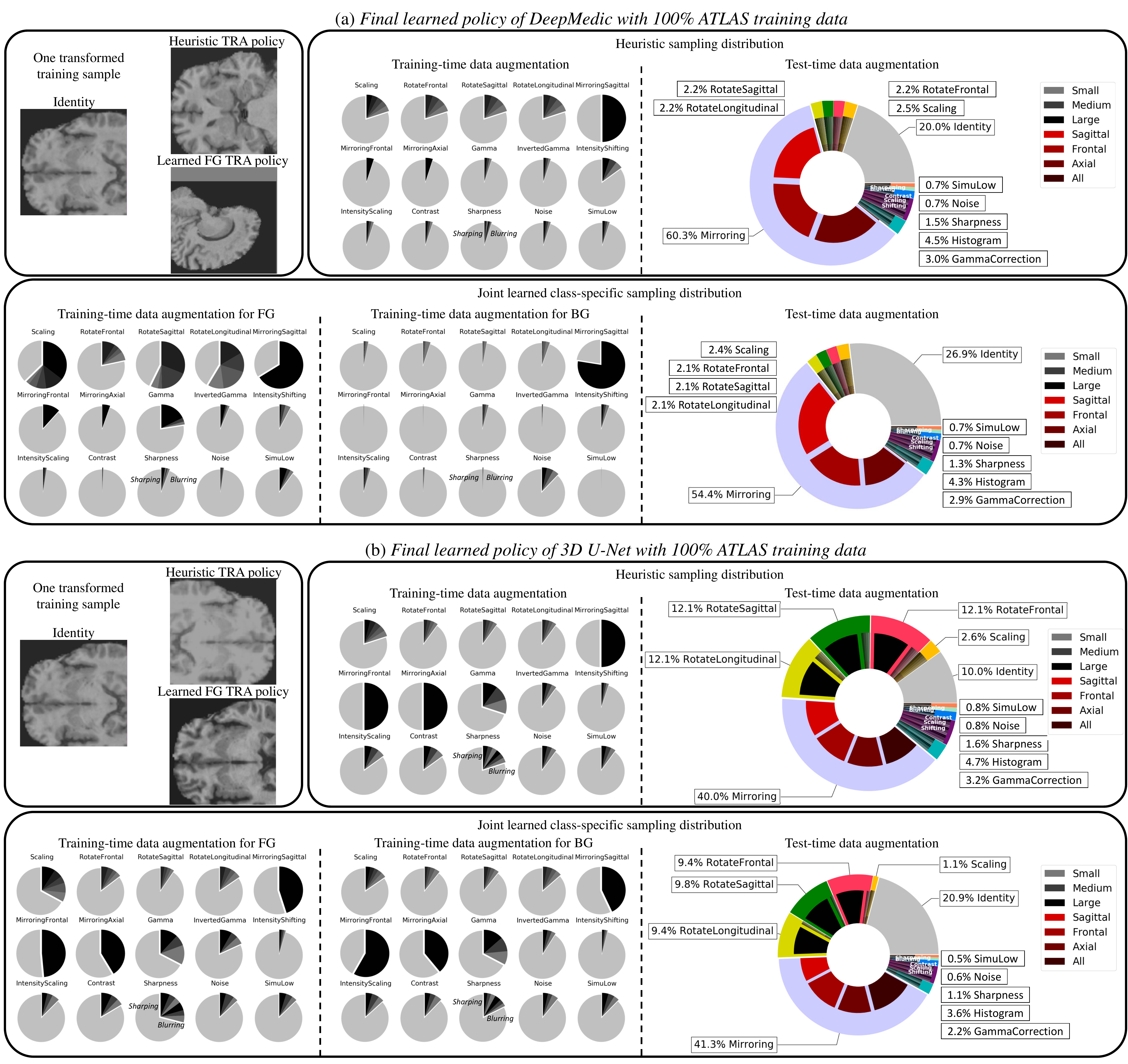}

\caption{The heuristic data augmentation policy and the learned probability distributions over augmentations based on different segmentation models for brain stroke lesion segmentation with 100\% ATLAS training data. We also visualize an example of the transformed foreground (FG) training sample with different sampling distributions for TRA. Our framework provides application-specific and class-specific data augmentation policies. We find the learned policies would adopt larger transformations to the FG than the background (BG) samples, implicitly alleviating the class imbalance issue.} \label{fig_policy}
\end{figure*}

Taking the manual segmentation as the ground truth, we calculate evaluation metrics including DSC, sensitivity (SEN), precision (PRC), 95\% Hausdorff distance (HD) (mm). We calculate the mean DSC results of different models under different settings for different datasets in Table~\ref{tablelvalsum}. We summarize more detailed results in Table~\ref{tabatlas}, \ref{tabkits}, \ref{taborganDM} and \ref{tabprostate}, separately. In order to assess the overall segmentation performance of different methods, we rank the methods according to different metrics under the same experiment setting and report the average rank (AVG rank) of the four metrics. The learned probability distributions over augmentations for brain lesion segmentation based on different models with 100\% ATLAS training data are summarized in Fig.~\ref{fig_policy}. As shown for TRA policies, the darkness of different pie chart segments stands for the magnitudes of the operations. For example, the lightest grey segments refer to the operation without any transformations and the darkest (black) segments represent the operations with large transformations. We summarize all the learned policies under different settings in supplementary material.

\begin{table}[ht]
\centering
\caption{Average DSC results of both DeepMedic and 3D U-Net for different segmentation tasks under varied settings using different data augmentation methods. Organ$_{\boldsymbol{r}}$ is the average performance of all rare organ classes.}\label{tablelvalsum}
\newsavebox{\tablelvalsum}
\begin{lrbox}{\tablelvalsum}

\begin{tabular}{m{20mm}<{\centering}|m{9mm}<{\centering}|m{16mm}<{\centering}|m{12mm}<{\centering}|m{15mm}<{\centering}}
\hlineB{3}
Dataset & Heuristic TRA w/o TEA & \textbf{Learned Class-Specific} w/o TEA & Heuristic TRA w/ Heuristic TEA & \textbf{Joint Learned Class-Specific} \\
\hlineB{1}
ATLAS~\cite{liew2018large} & 58.5 & \textbf{60.9} (+2.4)$^{**}$ & 61.1 & \textbf{62.4} (+1.3)$^{*}$ \\
KiTS~\cite{heller2019kits19}  & 70.0 & \textbf{74.9} (+4.9)$^{**}$ & 75.5 & \textbf{76.8} (+1.3)$^{*}$ \\
Organ~\cite{xu2015efficient} & 79.6 & \textbf{80.7} (+1.1)$^{*}$ & 80.3 & \textbf{81.1} (+0.8)$^{*}$ \\
Organ$_{\boldsymbol{r}}$~\cite{xu2015efficient} & 72.8 & \textbf{74.1} (+1.3)$^{*}$ & 73.5 & \textbf{74.6} (+1.1)$^{*}$ \\
Prostate~\cite{bloch2015nci,lemaitre2015computer} & 71.6 & \textbf{71.7} (+0.1)$^\sim$ & 74.0 & \textbf{76.4} (+2.4)$^{*}$ \\
\hline
\end{tabular}
\end{lrbox}
\scalebox{0.94}{\usebox{\tablelvalsum}}

{\raggedright \quad $^*p$-value $<$ 0.05; $^{**}p$-value $<$ 0.01; $^\sim p$-value $	\geq$ 0.05 (compared to Heuristic TRA w/o TEA or Heuristic TRA w/ Heuristic TEA) \par}
\end{table}

\begin{table*}[t]
\centering
\caption{Evaluation of brain stroke lesion segmentation on ATLAS based on different network architectures with different amounts of training data using different data augmentation methods. Best and second best results are in \textbf{bold}, with best also \underline{\textbf{underlined}}.}\label{tabatlas}
\newsavebox{\tableboxatlas}
\begin{lrbox}{\tableboxatlas}
\begin{tabular}{c|c|c|m{16mm}<{\centering}|m{6mm}<{\centering}|m{6mm}<{\centering}|m{6mm}<{\centering}|m{16mm}<{\centering}|m{6mm}<{\centering}|m{6mm}<{\centering}|m{6mm}<{\centering}|m{6mm}<{\centering}}
\hlineB{3}
\multirow{3}{*}{Model} & \multirow{3}{*}{\tabincell{c}{Training-time \\ augmentation}} & \multirow{3}{*}{\tabincell{c}{Test-time \\ augmentation}} & \multicolumn{4}{c|}{50\% training data} & \multicolumn{4}{c|}{100\% training data} & \multirow{3}{*}{\tabincell{c}{AVG \\ Rank \\ $\downarrow$}} \\
& & & \tabincell{c}{DSC \\ $\uparrow$} & SEN $\uparrow$ & PRC $\uparrow$ & 
HD $\downarrow$ & \tabincell{c}{DSC \\ $\uparrow$} & SEN $\uparrow$ & PRC $\uparrow$ & 
HD $\downarrow$ &  \\
\hlineB{1}
\multirow{9}{*}{DeepMedic~\cite{kamnitsas2017efficient}} & None & None & 51.7 & 50.6 & 65.0 & \textbf{\underline{20.4}} & 55.2 & 57.9 & 63.1 & \textbf{24.3} & 4.0  \\
& Heuristic~\cite{kamnitsas2017efficient} & None &  58.2 & 62.3 & 65.3 & 26.8 & 58.9 & \textbf{65.5} & 64.9 & 32.6 & 3.6 \\
& Heuristic$^\dagger$~\cite{kamnitsas2017efficient} & None & \textbf{59.1} & \underline{\textbf{64.7}} & 65.0 & 30.3 & \textbf{59.5}  & 63.0 & \underline{\textbf{67.5}} & 29.3 & 2.9 \\
& Learned~\cite{cubuk2018autoaugment, lim2019fast, li2020dada} & None & \textbf{\underline{59.5}} & 62.7 & \underline{\textbf{68.5}} & \textbf{25.1} & 58.4 & 62.9 & 67.0 & 26.2 & \textbf{2.6} \\
& \textbf{Learned Class-Specific} & None & \textbf{\underline{59.5}} (+1.3)$^{\sim}$ & \textbf{62.9} & \textbf{68.1} & 26.1 & \textbf{\underline{60.2}} (+1.3)$^*$ & \underline{\textbf{66.1}} & \textbf{67.3} & \textbf{\underline{22.7}} & \textbf{\underline{1.6}} \\
\cline{2-12}
& Heuristic~\cite{kamnitsas2017efficient} & Heuristic~\cite{isensee2021nnu} & 60.1 & 63.7 & 68.7 & \textbf{\underline{23.5}} & 60.6 & 65.7 & 67.7 & 27.6 & 3.5 \\
& \textbf{Learned Class-Specific} & Heuristic~\cite{isensee2021nnu} & \textbf{61.1} & \underline{\textbf{64.4}} & 70.2 & 25.6 & \textbf{61.6} & \underline{\textbf{66.2}} & 68.6 & \textbf{24.4} & 2.4 \\
& \textbf{Learned Class-Specific} & Learned~\cite{shanmugam2020and, kim2020learning} & \textbf{\underline{61.3}} & 64.2 & \textbf{70.8} & 25.3 & \textbf{61.6} & \textbf{66.0} & \textbf{69.7} & \textbf{\underline{23.5}} & \textbf{2.0} \\
& \multicolumn{2}{c|}{\textbf{Joint Learned Class-Specific}} & \textbf{\underline{61.3}} (+1.2)$^{\sim}$ & \textbf{64.3} & \underline{\textbf{71.2}} & \textbf{24.5} & \textbf{\underline{61.9}} (+1.3)$^\sim$ & 64.5 & \underline{\textbf{71.7}} & 25.0 & \textbf{\underline{1.9}} \\
\hline
\multirow{9}{*}{3D U-Net\cite{cciccek20163d}} & None & None &  54.6 & 56.3 & \underline{\textbf{67.2}} & \textbf{\underline{32.6}} & 56.7 & 58.8 & \underline{\textbf{69.8}} & \textbf{\underline{23.0}} & \textbf{3.0} \\
& Heuristic~\cite{isensee2021nnu} & None & 58.4 & 66.9 & 61.4 & 39.0 & 58.9 & 67.9 & 60.6 & 44.1 & 3.4 \\
& Heuristic$^\dagger$~\cite{isensee2021nnu} & None & 58.8 & \textbf{67.8} & 59.8 & 52.2 & 58.3 & \underline{\textbf{69.5}} & 56.2 & 58.6 & 3.8 \\
& Learned~\cite{cubuk2018autoaugment, lim2019fast, li2020dada} & None & \textbf{59.3} & 66.6 & 61.1 & 40.9 & \textbf{59.5} & \textbf{69.2} & 61.1 & 48.2 & 3.1 \\
& \textbf{Learned Class-Specific} & None & \textbf{\underline{62.0}} (+3.6)$^{**}$ & \underline{\textbf{68.8}} & \textbf{66.2} & \textbf{37.8} & \textbf{\underline{62.1}} (+3.2)$^{**}$ & 68.9 & \textbf{65.8} & \textbf{34.8} & \textbf{\underline{1.8}} \\
\cline{2-12}
& Heuristic~\cite{isensee2021nnu} & Heuristic~\cite{isensee2021nnu} & 61.7 & \textbf{67.0} & 69.6 & 22.0 & 62.3 & \underline{\textbf{68.6}} & 68.4 & 31.4 & 3.1 \\
& \textbf{Learned Class-Specific} & Heuristic~\cite{isensee2021nnu} & 61.8 & 66.4 & \underline{\textbf{70.2}} & \textbf{\underline{20.3}} & \textbf{63.9} & 68.4 & \underline{\textbf{72.2}} & \textbf{\underline{23.5}} & \textbf{2.1} \\
& \textbf{Learned Class-Specific} & Learned~\cite{shanmugam2020and, kim2020learning} & \textbf{62.2} & 66.9 & \textbf{69.9} & \textbf{20.5} & \textbf{63.9} & \textbf{68.5} & 72.0 & \textbf{24.5} & \textbf{2.3} \\
& \multicolumn{2}{c|}{\textbf{Joint Learned Class-Specific}} & \textbf{\underline{62.3}} (+0.6)$^\sim$ & \underline{\textbf{67.4}} & 69.2 & 28.9 & \textbf{\underline{64.0}} (+1.7)$^\sim$ & \textbf{68.5} & \textbf{72.1} & \textbf{24.5} & \textbf{\underline{2.1}} \\
\hline
\end{tabular}
\end{lrbox}
\scalebox{0.93}{\usebox{\tableboxatlas}}

{\raggedright \quad $^*p$-value $<$ 0.05; $^{**}p$-value $<$ 0.01; $^\sim p$-value $	\geq$ 0.05 (compared to Heuristic TRA w/o TEA or Heuristic TRA w/ Heuristic TEA) \par}

{\raggedright \quad $^\dagger$We train these models with both training and validation data. \par}

\end{table*}

\begin{table*}[t]
\centering
\caption{Evaluation of kidney tumor segmentation based on different network architectures with different amounts of training data using different data augmentation methods. Best and second best results are in \textbf{bold}, with best also \underline{\textbf{underlined}}.}\label{tabkits}
\newsavebox{\tableboxkits}
\begin{lrbox}{\tableboxkits}

\begin{tabular}{c|c|c|m{16mm}<{\centering}|m{6mm}<{\centering}|m{6mm}<{\centering}|m{6mm}<{\centering}|m{16mm}<{\centering}|m{6mm}<{\centering}|m{6mm}<{\centering}|m{6mm}<{\centering}|m{6mm}<{\centering}}
\hlineB{3}

\multirow{3}{*}{Model} & \multirow{3}{*}{\tabincell{c}{Training-time \\ data augmentation}} & \multirow{3}{*}{\tabincell{c}{Test-time \\ data augmentation}} & \multicolumn{4}{c|}{50$\%$} & \multicolumn{4}{c|}{100$\%$} & \multirow{3}{*}{\tabincell{c}{AVG \\ Rank \\ $\downarrow$}} \\
& & & DSC $\uparrow$ & SEN $\uparrow$ & PRC $\uparrow$ & HD $\downarrow$ & \tabincell{c}{DSC \\ $\uparrow$} & SEN $\uparrow$ & PRC $\uparrow$ & HD $\downarrow$ \\
\hlineB{1}
\multirow{8}{*}{DeepMedic~\cite{kamnitsas2017efficient}} & None &  None  & 40.1 & 35.4 & 56.2 & 93.0 & 51.1 & 50.0 & 62.0 & \textbf{\underline{72.8}} &  3.6 \\
& Heuristic~\cite{kamnitsas2017efficient} & None & 66.6 & 69.3 & 72.4 & 76.3 & \textbf{69.5} & 77.2 & \textbf{69.8} & \textbf{76.3}  & 2.6 \\
& Learned~\cite{cubuk2018autoaugment, lim2019fast, li2020dada} &  None & \textbf{69.1} & \textbf{71.3} & \textbf{75.1} & \textbf{\underline{61.8}} & \textbf{69.5} & \underline{\textbf{79.3}} & 67.6 & 89.4  & \textbf{2.1} \\
& \textbf{Learned Class-Specific} & None & \textbf{\underline{71.6}} (+5.0)$^{**}$ & \underline{\textbf{72.8}} & \underline{\textbf{76.8}} & \textbf{66.5} & \textbf{\underline{71.2}} (+1.7)$^\sim$ & \textbf{78.1} & \underline{\textbf{70.4}} & 88.3  & \textbf{\underline{1.5}} \\
\cline{2-12}
& Heuristic~\cite{kamnitsas2017efficient} & Heuristic~\cite{isensee2021nnu} & 70.5 & 70.5 & \textbf{78.5} & 58.7 & 72.9 & 77.9 & \underline{\textbf{74.1}} & 62.5  & 3.3 \\
& \textbf{Learned Class-Specific} & Heuristic~\cite{isensee2021nnu} & 72.5 & 73.4 & 78.3 & \textbf{48.1} & 73.1 & \textbf{79.5} & 73.2 & \textbf{\underline{57.2}}  & 2.8 \\
& \textbf{Learned Class-Specific} & Learned~\cite{shanmugam2020and, kim2020learning} & \textbf{72.8} & \underline{\textbf{73.6}} & \textbf{78.5} & \textbf{\underline{47.9}} & \textbf{73.3} & 79.3 & 73.6 & \textbf{60.5}  & \textbf{2.0} \\
& \multicolumn{2}{c|}{\textbf{Joint Learned Class-Specific}} & \textbf{\underline{73.3}} (+2.8)$^{**}$ & \textbf{73.5} & \underline{\textbf{79.7}} & 48.4  & \textbf{\underline{74.1}} (+1.2)$^\sim$ & \underline{\textbf{79.6}} & \textbf{74.0} & 71.7 & \textbf{\underline{1.9}} \\
\hline
\multirow{8}{*}{3D U-Net~\cite{cciccek20163d}}& None &  None & 43.5 & 39.5 & 60.9 & 104.6 & 60.3 & 57.1 & 71.6 & 78.4  & 4.0 \\
& Heuristic~\cite{isensee2021nnu} & None & 76.6 & 80.2 & \textbf{77.4} & \textbf{\underline{40.6}} & \textbf{77.6} & 82.1 & \textbf{77.6} & 59.5  & 2.5 \\
& Learned~\cite{cubuk2018autoaugment, lim2019fast, li2020dada} & None & \textbf{76.7} & \textbf{82.0} & 76.1 & 55.7 & \textbf{\underline{78.5}} & \underline{\textbf{84.2}} & 77.4 & \textbf{50.6}  & \textbf{2.1} \\
& \textbf{Learned Class-Specific} & None & \textbf{\underline{78.4}} (+1.8)$^\sim$ & \underline{\textbf{82.2}} & \textbf{\underline{78.0}} & \textbf{47.2} & \textbf{\underline{78.5}} (+0.9)$^\sim$ & \textbf{83.2} & \underline{\textbf{78.0}} & \textbf{\underline{48.8}}  & \textbf{\underline{1.3}} \\
\cline{2-12}
& Heuristic~\cite{isensee2021nnu} &  Heuristic~\cite{isensee2021nnu} & \textbf{78.8} & \textbf{82.1} & 79.4 & \textbf{\underline{37.2}} & 79.7 & 83.3 & 79.5 & 45.5  & 3.1 \\
& \textbf{Learned Class-Specific} &  Heuristic~\cite{isensee2021nnu} & 78.7 & 81.7 & \textbf{79.6} & \textbf{42.0} & \textbf{80.5} & \textbf{84.0} & 80.2 & 42.3  & 2.6 \\
& \textbf{Learned Class-Specific} & Learned~\cite{shanmugam2020and, kim2020learning} & \textbf{78.8} & 81.7 & \textbf{79.6} & \textbf{42.0} & \textbf{\underline{80.6}} & \underline{\textbf{84.1}} & \textbf{80.4} & \textbf{\underline{38.3}}  & \underline{\textbf{1.8}} \\
& \multicolumn{2}{c|}{\textbf{Joint Learned Class-Specific}} & \textbf{\underline{79.3}} (+0.5)$^\sim$ & \underline{\textbf{82.2}} & \underline{\textbf{79.7}} & 45.4 & 80.4 (+0.7)$^\sim$ & 83.5 & \underline{\textbf{80.6}} & \textbf{38.6}  & \textbf{2.0} \\
\hline
\end{tabular}
\end{lrbox}
\scalebox{0.93}{\usebox{\tableboxkits}}

{\raggedright \quad $^*p$-value $<$ 0.05; $^{**}p$-value $<$ 0.01; $^\sim p$-value $	\geq$ 0.05 (compared to Heuristic TRA w/o TEA or Heuristic TRA w/ Heuristic TEA) \par}

\end{table*}

\begin{table*}[t]
\centering
\caption{Evaluation of abdominal organ segmentation based on different network architectures using random data augmentation methods. Best and second best results are in \textbf{bold}, with the best also \underline{\textbf{underlined}}. AVG$_{\boldsymbol{r}}$ is the average performance of all rare classes including GB, E, AO, IVC, V, PA, RA and LA.}\label{taborganDM}
\newsavebox{\tableboxorganDM}
\begin{lrbox}{\tableboxorganDM}
\begin{tabular}{c|c|c|c|c|c|c|c|c|c|c|c|c|c|c|c|c|c}
\hlineB{3}
\multirow{2}{*}{Model} & \multirow{2}{*}{\tabincell{c}{Training-time \\ data augmentation}} & \multirow{2}{*}{\tabincell{c}{Test-time \\ data augmentation}} & \multicolumn{14}{c}{DSC} \\
& & & SP & RK & LK & GB & E & LIV & STO & AO & IVC & V & PA & RA & LA & AVG $\uparrow$ & {AVG}$_{\boldsymbol{r}}$ $\uparrow$ \\
\hlineB{1}
\multirow{8}{*}{DeepMedic~\cite{kamnitsas2017efficient}} & None &  None & 87.7 & 89.5 & 91.6 & 48.7 & 71.9 & 94.8 & 76.7 & 83.8 & 82.5 & 62.0 & 51.9 & 49.4 & 54.3 & 72.7 & 63.1 \\
& Heuristic~\cite{kamnitsas2017efficient} & None & 90.9 & 90.5 & 90.8 & 59.8 & 75.1 & 93.0 & 77.9 & 84.3 & 84.9 & 69.8 & 65.1 & 65.4 & 66.4 & \textbf{78.0} & 71.3  \\
& Learned~\cite{cubuk2018autoaugment, lim2019fast, li2020dada} & None & 93.4 & 90.8 & 93.1 & 63.6 & 76.0 & 94.1 & 81.7 & 86.2 & 84.2 & 71.5 & 66.7 & 68.2 & 64.6 & \textbf{\underline{79.5}} & \textbf{72.6} \\
& \textbf{Learned Class-Specific} & None & 91.8 & 89.7 & 91.0 & 65.7 & 76.6 & 93.9 & 79.6 & 85.2 & 85.2 & 69.9 & 68.9 & 69.7 & 66.8 & \tabincell{c}{\underline{\textbf{79.5}} \\ (+1.5)$^{**}$} & \tabincell{c}{\underline{\textbf{73.5}} \\ (+2.2)$^{**}$} \\
\cline{2-18}
& Heuristic~\cite{kamnitsas2017efficient} & Heuristic$^\dagger$~\cite{isensee2021nnu} & 91.4 & 90.6 & 92.2 & 60.3 & 78.6 & 93.1 & 78.1 & 84.4 & 85.3 & 70.5 & 66.4 & 65.4 & 69.2 & 78.9 & 72.5 \\
& \textbf{Learned Class-Specific} & Heuristic$^\dagger$~\cite{isensee2021nnu} & 92.1 & 90.2 & 92.1 & 65.3 & 77.3 & 94.0 & 80.8 & 86.6 & 86.1 & 69.2 & 68.5 & 66.7 & 64.9 & 79.5 & \textbf{73.1} \\
& \textbf{Learned Class-Specific} & Learned~\cite{shanmugam2020and, kim2020learning} & 92.1 & 89.9 & 91.5 & 65.6 & 77.5 & 94.0 & 80.4 & 86.3 & 86.0 & 69.7 & 68.8 & 68.7 & 66.6 & \textbf{\underline{79.8}} & \textbf{\underline{73.7}} \\
& \multicolumn{2}{c|}{\textbf{Joint Learned Class-Specific}} & 92.9 & 91.5 & 93.3 & 63.7 & 77.6 & 93.4 & 80.0 & 86.1 & 85.1 & 69.9 & 68.5 & 66.3 & 66.8 & \tabincell{c}{\textbf{79.6} \\ (+0.7)$^\sim$} & \tabincell{c}{73.0 \\ (+0.5)$^\sim$} \\
\hline
\multirow{8}{*}{3D U-Net~\cite{cciccek20163d}} & None &  None & 89.2 & 91.6 & 92.5 & 29.9 & 69.8 & 95.2 & 86.9 & 88.1 & 86.1 & 69.3 & 60.0 & 54.8 & 61.7 & 75.0 & 65.0 \\
& Heuristic~\cite{isensee2021nnu} & None & 94.9 & 92.7 & 92.6 & 55.4 & 75.9 & 96.0 & 89.3 & 91.6 & 88.0 & 72.9 & 75.4 & 67.7 & 67.8 & \textbf{81.5} & \textbf{74.3} \\
& Learned~\cite{cubuk2018autoaugment, lim2019fast, li2020dada} & None & 94.6 & 92.9 & 92.8 & 59.7 & 76.6 & 96.1 & 90.0 & 91.8 & 88.2 & 73.1 & 73.3 & 64.4 & 66.0 & \textbf{81.5} & 74.1 \\
& \textbf{Learned Class-Specific} & None & 95.0 & 93.4 & 92.9 & 59.1 & 76.0 & 96.2 & 88.9 & 91.9 & 88.0 & 73.8 & 74.3 & 61.9 & 72.6 & \tabincell{c}{\textbf{\underline{81.8}} \\ (+0.3)$^\sim$} & \tabincell{c}{\textbf{\underline{74.7}} \\ (+0.4)$^\sim$} \\
\cline{2-18}
& Heuristic~\cite{isensee2021nnu} & Heuristic~\cite{isensee2021nnu} & 95.3 & 93.6 & 92.8 & 58.4 & 73.2 & 96.2 & 90.4 & 91.8 & 88.6 & 74.6 & 75.2 & 64.5 & 69.0 & 81.8 & 74.4 \\
& \textbf{Learned Class-Specific} & Heuristic~\cite{isensee2021nnu} & 95.3 & 93.4 & 93.0 & 52.7 & 77.5 & 96.1 & 89.9 & 92.0 & 88.2 & 76.2 & 74.7 & 64.8 & 72.8 & 82.0 & 74.9 \\
& \textbf{Learned Class-Specific} & Learned~\cite{shanmugam2020and, kim2020learning} & 95.3 & 93.3 & 92.9 & 56.1 & 77.6 & 96.2 & 89.9 & 92.0 & 88.3 & 75.7 & 74.7 & 64.8 & 72.6 & \textbf{82.2} & \textbf{75.2} \\
& \multicolumn{2}{c|}{\textbf{Joint Learned Class-Specific}} & 94.4 & 93.5 & 93.1 & 61.8 & 80.8 & 96.0 & 89.3 & 90.9 & 87.7 & 72.7 & 77.6 & 66.7 & 71.4 & \tabincell{c}{\textbf{\underline{82.8}} \\ (+1.0)$^\sim$}  & \tabincell{c}{\textbf{\underline{76.2}} \\ (+1.8)$^\sim$}  \\
\hline
\end{tabular}
\end{lrbox}
\scalebox{0.79}{\usebox{\tableboxorganDM}}
{\raggedright \quad $^*p$-value $<$ 0.05; $^{**}p$-value $<$ 0.01; $^\sim p$-value $	\geq$ 0.05 (compared to Heuristic TRA w/o TEA or Heuristic TRA w/ Heuristic TEA) \par}

{\raggedright $^\dagger$We adopt a heuristic TEA policy with larger probability of identity transformation here because typical ones would decrease the performance. \par}
\end{table*}

\begin{table*}[t]
\centering
\caption{Evaluation of cross-site prostate segmentation based on different network architectures using different data augmentation methods. Best and second best results are in \textbf{bold}, with the best also \underline{\textbf{underlined}}.}\label{tabprostate}
\newsavebox{\tableboxprostate}
\begin{lrbox}{\tableboxprostate}
\begin{tabular}{c|c|c|m{16mm}<{\centering}|m{6mm}<{\centering}|m{6mm}<{\centering}|m{6mm}<{\centering}||m{6mm}<{\centering}|m{6mm}<{\centering}|m{6mm}<{\centering}|m{6mm}<{\centering}|m{6mm}<{\centering}}
\hlineB{3}
\multirow{3}{*}{Model} & Site A & Site A/B & \multicolumn{4}{c||}{Site B} & \multicolumn{4}{c|}{Site A} & \multirow{3}{*}{\tabincell{c}{AVG \\ Rank \\ $\downarrow$}} \\
& \multirow{2}{*}{\tabincell{c}{Training-time \\ data augmentation}} & \multirow{2}{*}{\tabincell{c}{Test-time \\ data augmentation}} & \multirow{2}{*}{\tabincell{c}{DSC \\ $\uparrow$}} & \multirow{2}{*}{\tabincell{c}{SEN \\ $\uparrow$}} & \multirow{2}{*}{\tabincell{c}{PRC \\ $\uparrow$}} & \multirow{2}{*}{\tabincell{c}{HD \\ $\downarrow$}} & \multirow{2}{*}{\tabincell{c}{DSC \\ $\uparrow$}} & \multirow{2}{*}{\tabincell{c}{SEN \\ $\uparrow$}} & \multirow{2}{*}{\tabincell{c}{PRC \\ $\uparrow$}} & \multirow{2}{*}{\tabincell{c}{HD \\ $\downarrow$}} & \\
& & & & & & & & & & \\
\hlineB{1}
\multirow{10}{*}{DeepMedic~\cite{kamnitsas2017efficient}} & None & None & 14.9 & 11.6 & 45.3 & 42.6 & 82.4 & 77.1 & 90.7 & 6.7 & 6.0 \\
& Heuristic~\cite{kamnitsas2017efficient} & None & 46.4 & 43.2 & 59.4 & 26.9 & 88.0 & \textbf{85.5} & 91.4 & 4.8 & 5.0 \\
& Heuristic Fine-Tuning$^\dagger$~\cite{kamnitsas2017efficient} & None & 56.7 & 46.4 & \underline{\textbf{77.5}} & \textbf{\underline{9.4}} & 27.6 & 20.0 & 81.4 & 18.9 & \textbf{2.5} \\
& Heuristic$^\ddagger$~\cite{kamnitsas2017efficient} & None & \textbf{69.3} & \textbf{67.2} & 73.5 & \textbf{15.1} & \textbf{88.1} & 84.8 & \textbf{92.3} & \textbf{\underline{4.5}} & \textbf{2.5} \\
& Learned$^\ddagger$~\cite{cubuk2018autoaugment, lim2019fast, li2020dada} & None & 65.8 & 62.8 & 75.1 & 21.9 & 87.5 & 83.4 & \underline{\textbf{92.6}} & \textbf{4.6} & 3.0 \\
& \textbf{Learned Class-Specific}$^\ddagger$ & None & \textbf{\underline{70.0}} (+0.7)$^\sim$ & \underline{\textbf{68.0}} & \textbf{75.9} & 18.7 & \textbf{\underline{88.2}} & \underline{\textbf{85.6}} & 91.5 & 4.7 & \textbf{\underline{1.8}} \\
\cline{2-12}
& Heuristic$^\ddagger$~\cite{kamnitsas2017efficient} & Heuristic~\cite{isensee2021nnu} & 69.4 & 66.3 & 76.5 & \textbf{8.0} & 88.2 & 84.6 & \underline{\textbf{92.6}} & 4.6 & 3.3 \\
& \textbf{Learned Class-Specific}$^\ddagger$ & Heuristic~\cite{isensee2021nnu} & 69.9 & 66.3 & \underline{\textbf{80.0}} & \textbf{8.0} & \textbf{\underline{88.8}} & \underline{\textbf{86.1}} & \textbf{92.4} & \textbf{\underline{4.4}} & \textbf{2.3} \\
& \textbf{Learned Class-Specific}$^\ddagger$ & Learned~\cite{shanmugam2020and, kim2020learning} & \textbf{70.2} & \textbf{67.7} & \textbf{77.6} & 15.3 & \textbf{88.5} & \textbf{86.0} & 91.7 & 4.6 & 2.5\\
& \multicolumn{2}{c|}{\textbf{Joint Learned Class-Specific}$^\ddagger$} & \textbf{\underline{72.8}} (+3.4)$^{**}$ & \underline{\textbf{71.0}} & 76.6 & \textbf{\underline{7.9}} & 88.2 & 85.4 & 91.8 & \textbf{4.5} & \textbf{\underline{1.5}} \\
\hline
\multirow{10}{*}{3D U-Net~\cite{cciccek20163d}} & None & None & 57.2 & 52.8 & \textbf{69.7} & \textbf{13.7} & 87.1 & 84.2 & \underline{\textbf{91.1}} & \textbf{\underline{5.3}} & 4.0\\
& Heuristic~\cite{isensee2021nnu} & None & 63.3 & 89.3 & 50.9 & 64.0 & \textbf{\underline{89.4}} & 88.4 & \textbf{90.8} & 16.0 & 5.0 \\
& Heuristic Fine-Tuning$^\dagger$~\cite{isensee2021nnu} & None & 68.7 & 63.9 & \underline{\textbf{84.1}} & \textbf{\underline{9.8}} & 55.6 & 48.6 & 77.2 & 25.1 & \textbf{2.8} \\
& Heuristic$^\ddagger$~\cite{isensee2021nnu} & None & \textbf{73.9} & 88.4 & 65.6 & 60.1 & \textbf{\underline{89.4}} & 88.9 & 90.0 & 18.7 & 3.8 \\
& Learned$^\ddagger$~\cite{cubuk2018autoaugment, lim2019fast, li2020dada} & None & \textbf{\underline{76.5}} & \textbf{89.8} & 67.9 & 44.2 & 87.0 & \textbf{89.2} & 85.3 & 26.9 & \textbf{\underline{2.5}} \\
& \textbf{Learned Class-Specific}$^\ddagger$ & None & 73.2 (-0.7)$^\sim$ & \underline{\textbf{90.1}} & 63.4 & 41.5 & \textbf{87.7} & \underline{\textbf{89.6}} & 86.6 & \textbf{15.4} & 3.0 \\
\cline{2-12}
& Heuristic$^\ddagger$~\cite{isensee2021nnu} & Heuristic~\cite{isensee2021nnu} & 78.7 & \underline{\textbf{90.5}} & 71.4 & 28.0 & \textbf{89.7} & 89.1 & \textbf{90.6} & \textbf{12.1} & 2.8 \\
& Learned$^\ddagger$~\cite{cubuk2018autoaugment, lim2019fast, li2020dada} & Heuristic~\cite{isensee2021nnu} & 78.5 & 89.2 & 71.6 & 31.1 & 88.9 & 89.2 & 89.0 & 12.8 & 3.5 \\
& Learned$^\ddagger$~\cite{cubuk2018autoaugment, lim2019fast, li2020dada} & Learned~\cite{shanmugam2020and, kim2020learning} & \textbf{79.6} & \textbf{89.9} & \textbf{72.7} & \textbf{23.6} & 89.0 & \underline{\textbf{89.3}} & 89.1 & 12.3 & \textbf{2.0} \\
& \multicolumn{2}{c|}{\textbf{Joint Learned}$^\ddagger$} & \textbf{\underline{80.0}} (+1.3)$^\sim$ & 88.2 & \underline{\textbf{74.1}} & \textbf{\underline{18.9}} & \textbf{\underline{90.0}} & \underline{\textbf{89.3}} & \underline{\textbf{91.0}} & \textbf{\underline{11.8}} & \textbf{\underline{1.8}} \\
\hline
\end{tabular}
\end{lrbox}
\scalebox{0.95}{\usebox{\tableboxprostate}}

{\raggedright \quad $^*p$-value $<$ 0.05; $^{**}p$-value $<$ 0.01; $^\sim p$-value $	\geq$ 0.05 (compared to Heuristic$^\ddagger$ TRA w/o TEA or Heuristic$^\ddagger$ TRA w/ Heuristic TEA) \par}

{\raggedright \quad $^\dagger$We pretrain these models with training data from site A and fine-tune with validation data from site B. \par}

{\raggedright \quad $^\ddagger$We train these models with both training data from site A and validation data from site B. \par}

\end{table*}

\begin{figure*}[t]
\centering
\includegraphics[width=\textwidth]{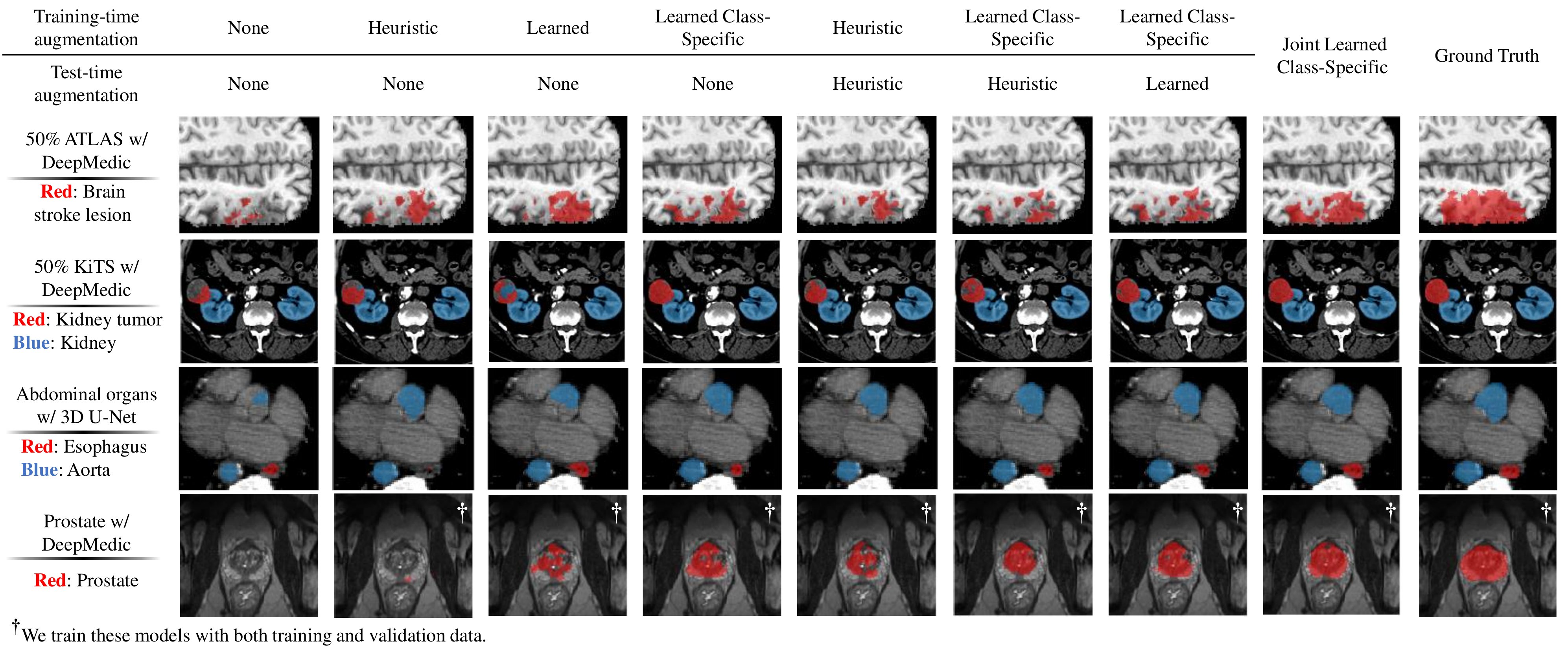}

\caption{Visualization of different datasets and segmentation results with different data augmentation methods. The proposed data augmentation framework can help the model produce overall better segmentation results with higher sensitivity. Best viewed in color.} \label{fig_example}
\end{figure*}

\subsubsection{The effectiveness of class-specific TRA}

Heuristic TRAs, which were tuned based on varied segmentation tasks~\cite{kamnitsas2017efficient, isensee2021nnu}, significantly help improve the segmentation performance in all cases compared with models trained without TRA. This indicates that TRA is vital for medical image segmentation as limited training data and class imbalance can easily lead to model overfitting~\cite{li2020analyzing}.

Learned TRA, which is optimized with validation data, can provide application-specific policies and is more effective than heuristic TRA in most cases. We find that the models trained with learned TRA can even outperform the ones trained with heuristic TRA that use both training and validation data, as shown in Table~\ref{tabatlas}. This may indicate that it will be more effective to increase the heterogeneity within the training data by adopting application-specific TRA than adding a small amount of training data. We find RandAugment with specific magnitude could be more effective than the learned one in some cases. Specifically, RandAugment-L is better than the learned ones for kidney tumor segmentation under specific setting, as shown in Table~\ref{tabkits}. This might indicate that learned TRA is prone to overfitting the validation data and the optimized policies are not guaranteed to be optimal for unseen test data, as also found in~\cite{cubuk2019randaugment}.

In contrast, class-specific TRA can better model the heterogeneity of the real data by taking class imbalance into account, and thus overfit less and perform better on unseen test data than alternative methods. We argue that class-specific TRA is important as it concerns the imbalanced nature of the segmentation datasets and directly regularizes the training data in an implicit way. As shown in Fig.~\ref{fig_policy}, compared with heuristic TRA, the learned policies tend to generate larger transformations for foreground samples while adopting smaller transformations to background samples. In segmentation, foreground classes are typically underrepresented and a learned baseline model would be biased towards the majority class. As a result, the model would map the foreground samples near the decision boundary and cause false negatives, as shown in~\cite{li2020analyzing}. Class-specific TRA can mitigate the class imbalance problem by inducing larger variance within the foreground samples, making the model learn a better decision boundary, consistently leading to better segmentation results with higher sensitivity. Particularly, we find class-specific TRA would improve the segmentation performance of rare classes more significantly (c.f. Table~\ref{taborganDM}) as it can enhance the rare class representation by increasing the heterogeneity of foreground sample variation. We also find that the probabilities of spatial transformations change more significantly compared to intensity transformations. This might indicate that spatial transformations are more effective in increasing the heterogeneity within training data. We validate our methods for prostate segmentation under domain shifts where the training and test data is collected under different conditions. We find directly fine-tuning the segmentation models with limited target data provides worse results than training with data from both domains. We report the segmentation results of both site B and site A with cross-site prostate segmentation in Table~\ref{tabprostate}. Although the learned data augmentation is optimized based on the validation data from site B (target domain), the models can still generalize well on site A. In addition, as we show in supplementary material, we find that our method can help the models generalize better on unseen test domains which are different from either site A or site B. This indicates that our method is robust to domain shifts and can be a safe choice to calibrate the segmentation performance of different domains within multi-domain learning.

\subsubsection{The effectiveness of joint optimization}

We find heuristic TEA can help the pretrained models produce better overall segmentation results with higher precision. This is because the ensemble of multiple predictions can reduce false positives as the models are unlikely to produce the same kind of false positives with all the transformed images. However, when TRA is optimized based on validation data without TEA, heuristic TEA might not work well as the model may overfit to the original data distribution and thus fail to generalize to the transformed data. Specifically, we observe that heuristic TEA would decrease the model performance for 3D U-Net trained with 50\% ATLAS training data (-0.3 in terms of DSC, c.f. Table~\ref{tabatlas}) and Deepmedic trained for prostate segmentation (-0.1 in terms of DSC, c.f. Table~\ref{tabprostate}) using learned class-specific TRA. In comparison, learned TEA can refine the transformations to fit the pretrained models and improve the results for most cases. 

However, learned TEA alone does not affect model training and cannot change the results significantly compared to heuristic TEA. In contrast, our method optimizes TRA based on TEA along the training process, jointly aligning the data distributions resulting in larger overlaps. For example, as illustrated in Fig.~\ref{fig_policy}(a), the learned TEA policy would increase the probability of flipping in sagittal planes for DeepMedic trained with 100\% ATLAS training data. It might be because the left and right hemispheres of human brains are generally symmetric. Correspondingly, TRA would tune the training data distribution with more samples flipped in the sagittal planes. In this way, the segmentation models not only have lower risks of making the same false positives but also generalize better on varied transformed samples. As a result, we find that the joint optimization further boosts the segmentation performance by achieving higher precision and sensitivity.

We argue that the joint optimization is crucial for data augmentation as it explicitly aligns the training and test-time conditions. Otherwise, the model may get stuck into a local minimum where we cannot find effective test-time transformations to fit the training data distribution. For example, we find that when compared with segmentation without TEA, learned TEA brings limited improvements for 3D U-Net trained with 50\% KiTS training data (0.4 in terms of DSC, c.f. Table~\ref{tabkits}) and DeepMedic trained for prostate segmentation (0.2 for site B in terms of DSC, c.f. Table~\ref{tabprostate}). This indicates that the predictions on most chosen transformations cannot contribute much to the results on top of the predictions of the original test images. In contrast, the joint optimization leverages the varied test-time transformations and improve the segmentation (0.9 and 2.8 separately in terms of DSC).

We notice that the learned TEA policies would generally prefer the original images (identity). In addition, the transformations which are not included in heuristic TEA are hardly useful. These findings indicate that we may not need to apply large transformations to the test data to improve generalization.

We visualize some segmentation results in Fig.~\ref{fig_example}. Similar to the findings in a previous study~\cite{li2020analyzing}, the model trained with imbalanced dataset would be prone to undersegment the foreground samples as a result of overfitting under class imbalance. Our class-specific TRA model can significantly reduce false negatives and improve the sensitivity of segmentation results. We observe heuristic TEA could cause under-segmentation while the joint optimization can further help the model improve segmentation performance by identifying more foreground samples. We further validate our methods with cardiac segmentation in MR images in supplementary material to prove that our methods can work well with anisotropic images under domain shifts.

\subsection{Limitations}

Our data augmentation algorithm aims to optimize the sampling distributions for TRA and TEA, and thus, automatically adapt data augmentation policies to given task. However, it might not be very effective when the predefined policies are already nearly optimal. For example, we observe that our methods do not bring much improvements for kidney tumor segmentation based on 3D U-Net when trained with 100\% training data. This is possibly because that the predefined policies were already optimized given it is the winning solution for the challenge.

We notice that class-specific TRA could be less effective with 3D U-Net on prostate segmentation. This may be due to the sampled patches always containing foreground, as the image size of this dataset is relatively small, and the structures-of-interest are relatively large. In this case, the optimization could be misled by the class-specific constraints. In practice, this could be alleviated by adopting a smaller patch size, and some investigations can be found in the supplementary material. Moreover, we could consider to restrict the regions of loss calculation to make our algorithms compatible with similar cases where the patch size is large up to the image size. This would need to be explored in future work.

The joint optimization of TRA and TEA will not be effective when TEA decreases the segmentation performance. For example, we find that the joint optimization cannot bring much improvements for DeepMedic with abdominal organ segmentation where most transformations for TEA do not seem to help much and the augmented validation data would improperly influence the TRA optimization. Therefore, we suggest validating the effectiveness of TEA before adopting the joint optimization.

Although we the proposed method can consistently improve the segmentation performance under varied scenarios, we observe that not all the results show statistical significance when compared to heuristic baselines. This might be due to the small size of the test set. We show that our methods show significant improvements when more test data is available (c.f. Table~\ref{tablelvalsum}). We observe that distance based metrics such as HD is unstable for the evaluation of imbalanced regions-of-interest (ROIs) because small false positive predictions could largely increase those metrics. After eliminating the false positive predictions with component-based post-processing, our method can always perform better in terms of both DSC and HD, as we demonstrate in supplementary material.

We present and validate our method in the context of medical image segmentation. We think that it has the potential to be extended to long-tailed image classification tasks where different classes have different properties and TEA is also important for better generalization. We show some initial experiments in supplementary material and will leave the in-depth investigation for future works.

\section{Conclusion}
\label{sec:conclusion}

We presented a general data augmentation framework for medical image segmentation. Compared with current solutions, our method aims to bridge the gap between training and test data distributions by class-specific TRA and joint optimization of TRA and TEA. We observe promising improvements in various tasks and models, making the proposed framework an attractive alternative to heuristic data augmentation strategies. We believe that the learned policies can provide valuable insights for practitioners to inform dynamic data collection and future designs of image transformations for data augmentation.



\section*{Acknowledgements}
Z.Li is grateful for the China Scholarship Council (CSC) Imperial Scholarship. This project has received funding from the ERC under the EU's Horizon 2020 research and innovation programme (grant No. 757173).

\newpage

\setcounter{section}{0}

\renewcommand\thesection{\Alph{section}}

\section*{Supplementary Material}

\localtableofcontents

\subsection{Learning Scheme}

\begin{figure*}[t]
\centering
\includegraphics[width=0.8\textwidth]{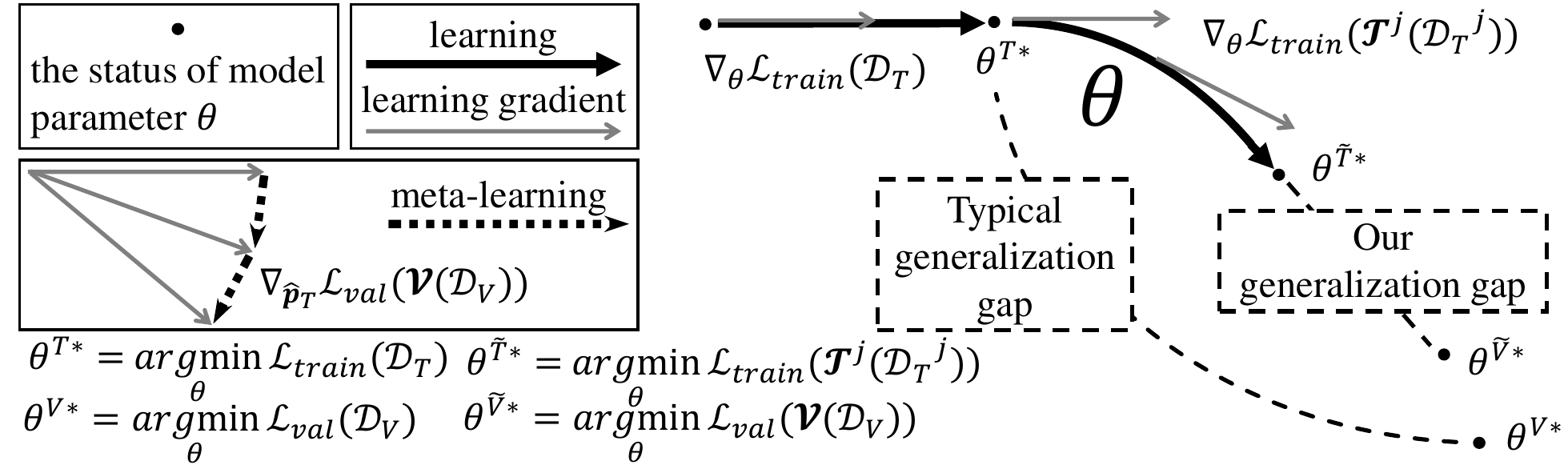}

\caption{The learning scheme of the proposed method. When we train the model with the training data $\mathcal{D}_T$ and the training criterion $\loss_{train}$, the model does not always generalize well on the validation data $\mathcal{D}_V$. We aim to close this generalization gap explicitly by the joint optimization of class-specific TRA $\mathbfcal{T}^j(\cdot)$ and TEA $\mathbfcal{V}(\cdot)$.} \label{figa1}
\end{figure*}

We describe the high-level learning scheme of the proposed method in Fig.~\ref{figa1}. During an optimization process (commonly stochastic gradient descent for neural networks), we minimize the $\mathcal{L}_{train}$ over all training data and yield the learned parameters $\theta^{T*}$. In practice, the learned $\theta^{T*}$ is sub-optimal for validation data because the training data cannot cover all the underlying data properties. Data augmentation is widely utilized to implicitly reduce the generalization gap $\theta^{T*}$ $\rightarrow$ $\theta^{V*}$ by using additional artificial training samples $\mathbfcal{T}(\mathcal{D}_T)$. However, it is not guaranteed that $\mathbfcal{T}(\cdot)$ is always effective during this process.

We propose to explicitly close the generalization gap by aligning the training and test data distribution. On the one hand, we optimize the model parameter from $\theta^{T*}$ to $\theta^{\tilde{T}*}$ by learning a TRA model with a meta-learning scheme. On the other hand, we transform the validation data in the way it can be easier to be recognized and change the target optimal parameters of validation from $\theta^{V*}$ to $\theta^{\tilde{V}*}$. By the joint optimization of class-specific TRA $\mathbfcal{T}^j(\cdot)$ and TEA $\mathbfcal{V}(\cdot)$, we are able to close the generalization gap from $\theta^{T*}$ $\rightarrow$ $\theta^{V*}$ to $\theta^{\tilde{T}*}$ $\rightarrow$ $\theta^{\tilde{V}*}$.

\subsection{Detailed Optimization Process}

\begin{figure*}[t]
\centering
\includegraphics[width=\textwidth]{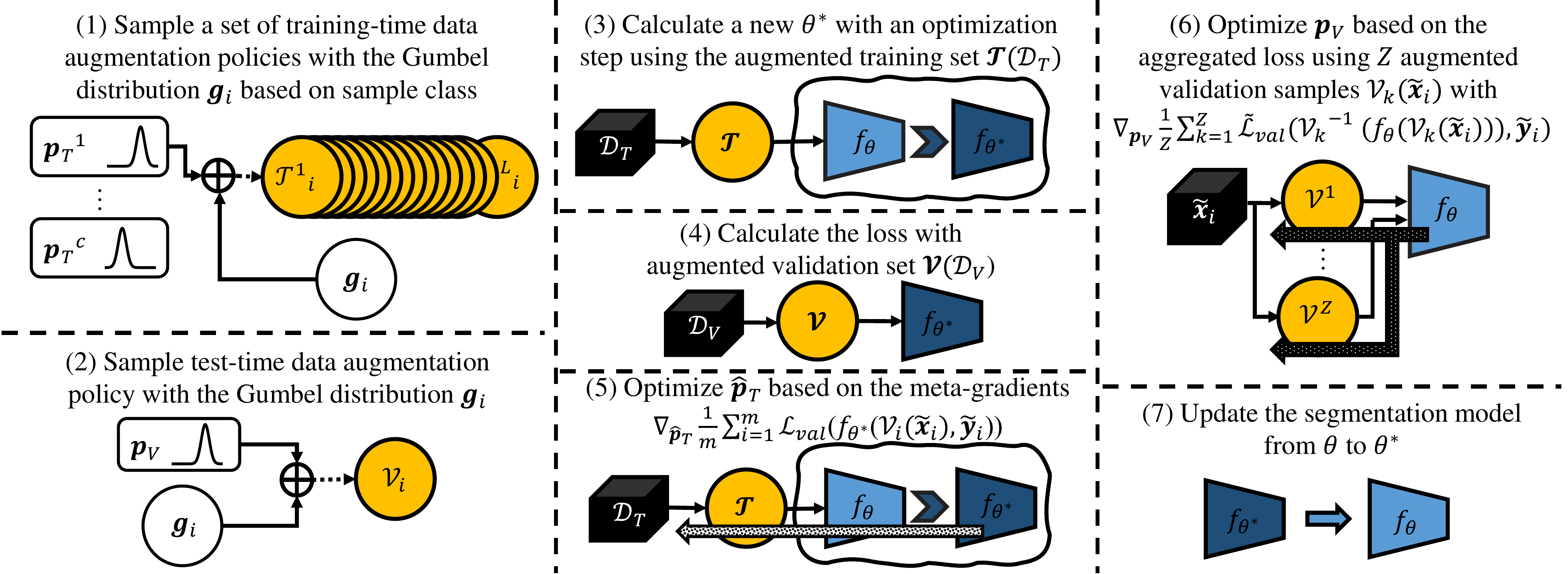}

\caption{Illustration of one iteration during the training process. It is consistent with Algorithm~\ref{alg:AugmentDistribution}.} \label{figa2}
\end{figure*}

The training process of one iteration is illustrated in detail in Fig.~\ref{figa2}.

\subsection{Implementation Details}

\subsubsection{Derivatives calculation based on implicit function theorem}

To compute $ \frac{\partial \theta^*}{\partial \hat{\boldsymbol{p}}_T}$, one can calculate the total derivatives on $\nabla_\theta  \mathcal{L}_{train}(\theta^*,\hat{\boldsymbol{p}}_T)=\boldsymbol{0}$ w.r.t. $\hat{\boldsymbol{p}}_T$ from both sides, assuming that $\nabla_\theta  \mathcal{L}_{train}(\theta^*,\hat{\boldsymbol{p}}_T)$ is continuously differentiable at $\boldsymbol{0}$~\cite{bengio2000gradient, zela2019understanding}:

\begin{equation}
    \frac{\partial \nabla_{\theta}\mathcal{L}_{train}(\theta^*,\hat{\boldsymbol{p}}_T)}{\partial \theta}\frac{\partial \theta^*}{\partial \hat{\boldsymbol{p}}_T} + \frac{\partial \nabla_{\theta}\mathcal{L}_{train}(\theta^*,\hat{\boldsymbol{p}}_T)}{ \partial \hat{\boldsymbol{p}}_T} = \boldsymbol{0}.
    \label{eq:funcimplicit}
\end{equation}

Then, with the assumption that the Hessain $\nabla^2_{\theta}\mathcal{L}_{train}(\theta^*,\hat{\boldsymbol{p}}_T)$ is invertable, we can yield:

\begin{equation}
    \frac{\partial \theta^*}{\partial \hat{\boldsymbol{p}}_T} = -(\nabla^2_{\theta}\mathcal{L}_{train}(\theta^*,\hat{\boldsymbol{p}}_T))^{-1} \nabla^2_{\theta,\hat{\boldsymbol{p}}_T}\mathcal{L}_{train}(\theta^*,\hat{\boldsymbol{p}}_T).
    \label{eq:omegacal}
\end{equation}

The results contain a Hessian $\nabla^2_{\theta,\hat{\boldsymbol{p}}_T}\mathcal{L}_{train}(\theta^*,\hat{\boldsymbol{p}}_T)$ and not practical to compute. Therefore, we follow the heuristics used in~\cite{finn2017model} to compute the derivatives.

\subsubsection{Meta-gradient calculation}

The gradient calculation in Eq.~\ref{eq:func3} can be further simplified based on finite difference approximation following~\cite{liu2018darts}. With some small $\epsilon = 0.01 / \left\Vert \nabla_{\theta}\mathcal{L}_{val}(\theta^*) \right\Vert_2$, we calculate two new parameters with $\theta^{\pm} = \theta \pm \epsilon\nabla_{\theta}\mathcal{L}_{val}(\theta^*)$.

With this notion, the second-order gradient can be written as:

\begin{equation}
\begin{split}
    & \nabla_{\hat{\boldsymbol{p}}_T^t,\theta}^{2} \mathcal{L}_{train}(\theta^*, \hat{\boldsymbol{p}}_T^t) \approx \\
    & \frac{\nabla_{\hat{\boldsymbol{p}}_T^t} \mathcal{L}_{train}(\theta^{+}, \hat{\boldsymbol{p}}_T^t) - \nabla_{\hat{\boldsymbol{p}}_T^t} \mathcal{L}_{train}(\theta^{-}, \hat{\boldsymbol{p}}_T^t)}{2\epsilon\nabla_{\theta}\mathcal{L}_{val}(\theta^*)} \\
    & \Leftrightarrow \quad \quad \nabla_{\hat{\boldsymbol{p}}_T^t,\theta}^{2} \mathcal{L}_{train}(\theta^*, \hat{\boldsymbol{p}}_T^t)\nabla_{\theta}\mathcal{L}_{val}(\theta^*) \approx \\
    & \frac{\nabla_{\hat{\boldsymbol{p}}_T^t} \mathcal{L}_{train}(\theta^{+}, \hat{\boldsymbol{p}}_T^t) - \nabla_{\hat{\boldsymbol{p}}_T^t} \mathcal{L}_{train}(\theta^{-}, \hat{\boldsymbol{p}}_T^t)}{2\epsilon}.
    \label{eqn:dartsfinitediffs}
\end{split}
\end{equation}

In this way, we reduce the calculation complexity from $O(|\hat{\boldsymbol{p}}_T||\theta|)$ to $O(|\hat{\boldsymbol{p}}_T|+|\theta|)$ and can approximate Eq.~\ref{eq:func3} with two forward processes of $f_{\theta^*}$.

\subsubsection{Efficient data augmentation sampling}

Attentive reader may find sampling and applying different transformations to $\boldsymbol{x}$ (or $\tilde{\boldsymbol{x}}_i$) during each iteration is time-consuming. In practice, we bypass this bottleneck by fetching a number of transformed samples in advance. Then when we input the transformed samples into the model, we sample from the probability again and get the corresponding $\boldsymbol{s}_i$ as if it is sampled from the current distribution. In this way, the sampling and data augmentation process can be done in parallel with network training and does not need additional time.

\subsubsection{Efficient optimization}

The proposed method described in Algorithm~\ref{alg:AugmentDistribution} would triple the training time compared with vanilla training process. However, the training time can be significantly reduced by updating the data augmentation policies (step 8 and 9 in Algorithm~\ref{alg:AugmentDistribution}) once several iterations. We find we can achieve similar results when the policies are updated once 10 iterations, only increasing the training time by 20\%.

\subsection{Proof of Concept on CIFAR-10}

In order to demonstrate that our method can effectively select useful augmentations, we first show results with a toy example on CIFAR-10~\cite{krizhevsky2009learning}. In this experiment, we optimize TRA and TEA separately. We use Wide-ResNet-40-2~\cite{BMVC2016_87} as the network backbone and pick 5120 images as the training set while another 5120 images as the validation set. We test on the official split of test set including 10000 images.

\subsubsection{Predefined transformations}

\begin{figure*}[t]
\centering
\includegraphics[width=0.9\textwidth]{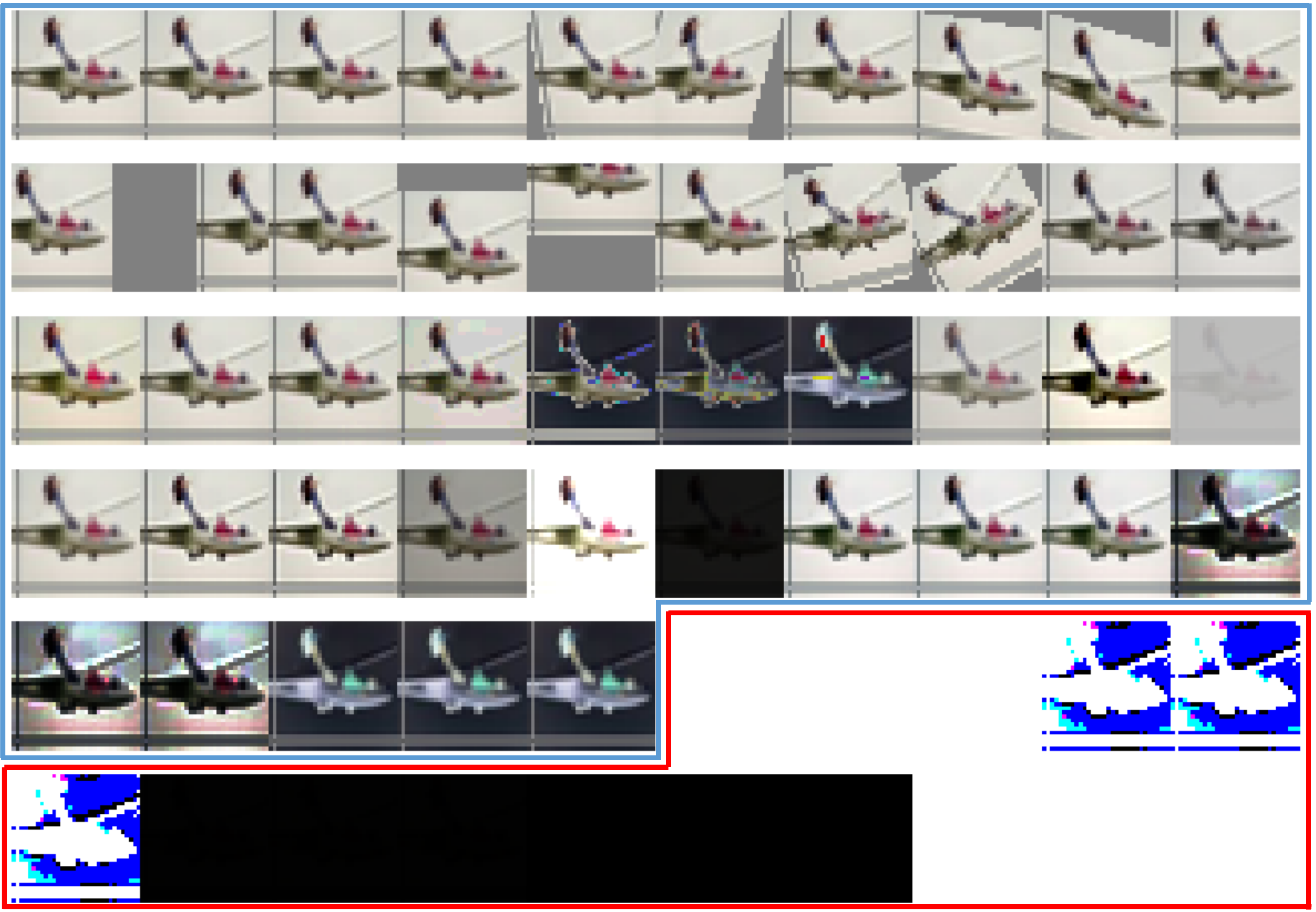}

\caption{Visualization of the CIFAR-10 transformation set, which contains 45 good transformations (marked with blue lines) and 15 bad transformations (marked with red lines).} \label{figa3}
\end{figure*}

We use the same transformation set for both TRA and TEA in this toy example. We initialize the augmentation distribution uniformly with 45 good transformations and 15 bad transformations. We adopt the good transformations from AutoAugment~\cite{cubuk2018autoaugment} including shearing, translation, rotation, color enhancement, posterization, solarization, contrast changing, sharpening, brightness changing, historgram equalization and inverting. We design the bad transformations as extremely low contrast and large intensity shifts. Given the original image $\boldsymbol{x}$, the bad transformations would apply $\mathcal{T}$($\boldsymbol{x}$) = $\boldsymbol{x}$$^2$, $\mathcal{T}$($\boldsymbol{x}$) = $\boldsymbol{x}$$^4$, $\mathcal{T}$($\boldsymbol{x}$) = $\boldsymbol{x}$ $\times$ 0.01, $\mathcal{T}$($\boldsymbol{x}$) = --$\boldsymbol{x}$ $\times$ 0.01 and $\mathcal{T}$($\boldsymbol{x}$) = $\boldsymbol{x}$ + 300. We show an example of totally 60 transformed images in Fig.~\ref{figa3}.

\subsubsection{Learning TRA}

The resulting probability distribution after training with the proposed scheme is visualized in Fig.~\ref{figa3}, the probabilities of all the bad transformations are low such that these are suppressed during training. The learned augmentations can improve the final accuracy from 83.8$\%$ to 85.1$\%$ on the validation set, compared to a policy with uniform probabilities of all 60 transformations. A model trained without any augmentation achieves 81.8$\%$. 

We apply the learned policy and train the same model with the total 10240 images from scratch, we find we can improve the performance from 88.2$\%$ to 89.0$\%$ on the test set, compared to a policy with uniform probabilities of all 60 transformations. A model trained without any augmentation achieves 87.1$\%$.

\begin{figure*}[t]
\centering
\includegraphics[width=0.7\textwidth]{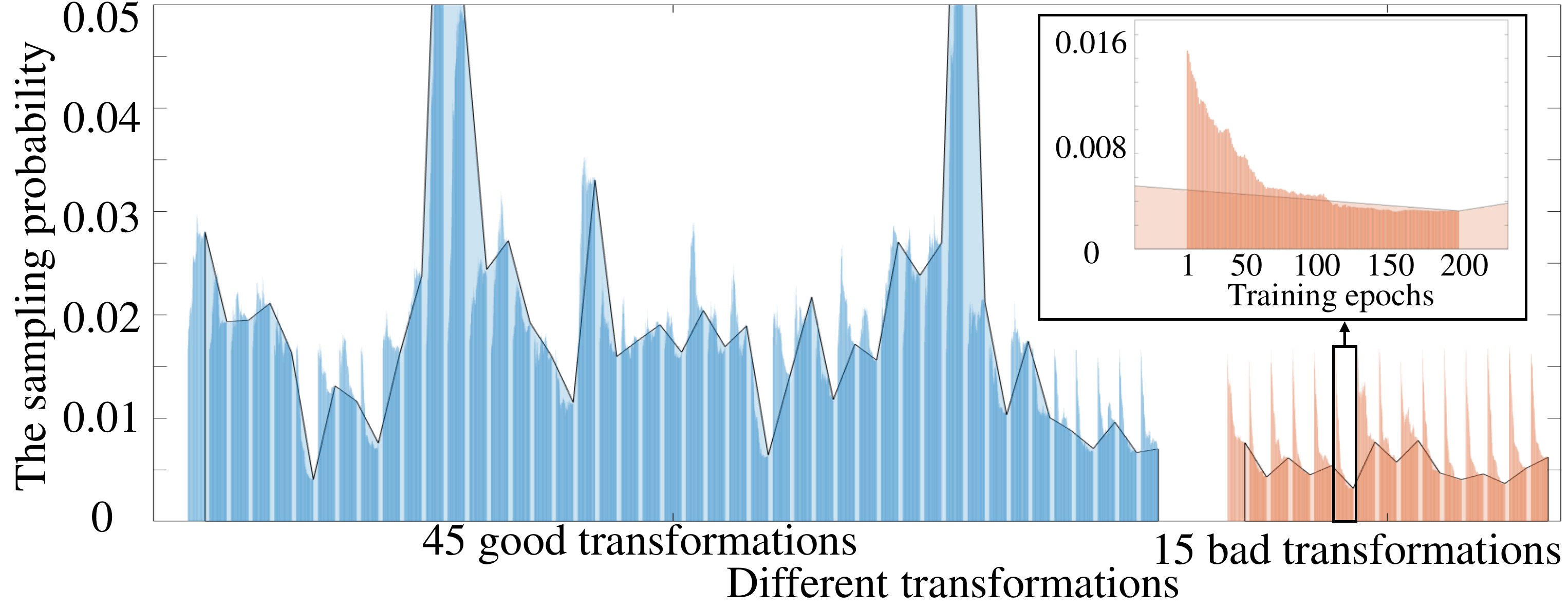}

\caption{Proof of concept on TRA with CIFAR-10. We manually add 15 bad transformations to the transformation set, as marked with orange color. Our method learns to decrease their probabilities during the training process.} \label{figa4}
\end{figure*}

\subsubsection{Learning TEA}

\begin{figure*}[t]
\centering
\includegraphics[width=0.7\textwidth]{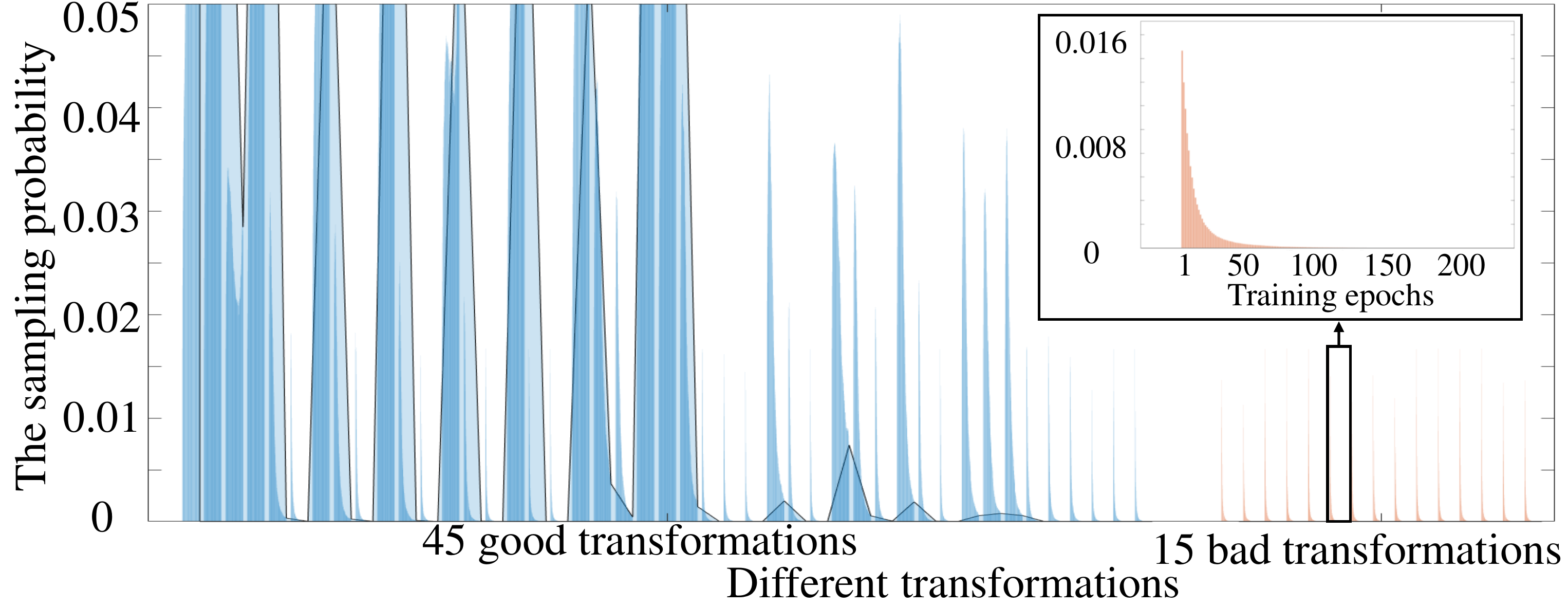}

\caption{Proof of concept on TEA with CIFAR-10. We  manually add the same 15 bad transformations to the transformation set and our method learns to decrease their probabilities during training.} \label{figa5}
\end{figure*}

The resulting TEA probability during the training process is demonstrated in Fig.~\ref{figa5}. Similarly, the probabilities of all the bad transformations are decreased during training.

\begin{figure*}[t]
\centering
\includegraphics[width=0.9\textwidth]{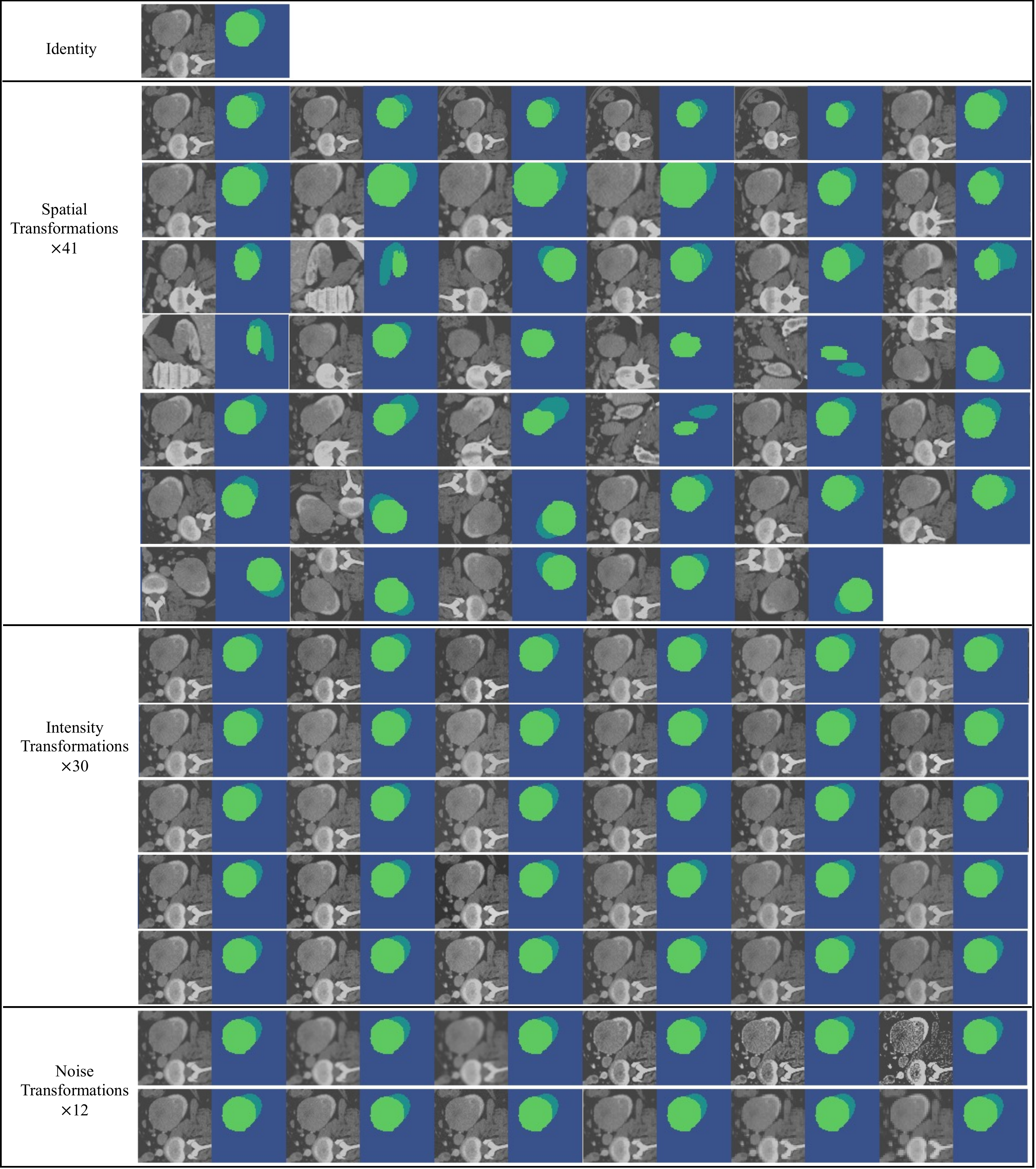}

\caption{Visualization of $K$=84 choices of transformations for TEA we use in this study. We apply different transformations to a test sample of KiTS. } \label{figa6}
\end{figure*}

\subsection{List of Operations for Training-Time Data Augmentation}

We list all the operations $\mathcal{O}$ we use for TRA in Table~\ref{tabtraug}. Note that we design most operations with stochastic magnitudes in a symmetric way. That is to say, the transformed image could be transformed back to the original image using the same operation, with the probability of 50\%. In this way, we can increase the variance of training dataset with more realistic and potentially useful samples.

Different from the operation set in AutoAugment which includes operations with deterministic magnitudes~\cite{cubuk2018autoaugment}, we design each operation with stochastic magnitudes which are sampled from a uniform sampling probability. In this way, we can cover transformations with larger variance and realize similar functionalities with TRA used in prevailing segmentation methods.

A straightforward question is whether we can further improve TRA by 1) adding the number of predefined transformations, 2) optimizing the magnitude of transformations as well or 3) including more complicated transformations such as generative models (i.e. utilizing a network to generate the augmented samples). In fact, we find that if we extend the transformation set with more choices (i.e. larger $K$), the performance would not be further improved. It is because that enlarging $K$ would make transformations being sampled less and make TRA harder to optimize. We find 2-10 is a reasonable range for $K$. It would be feasible to additionally optimize the magnitude of the operations. However, we find that it would make the optimization process unstable when we also optimize the operation magnitudes. This is might because controlling the operation magnitude is hard and the transformed images could become unrealistic. In addition, we are not sure if it is appropriate to optimize operation magnitude for the reduction of validation loss, as this might converge to local minimum. We also try to utilize a neural network to transform the training samples. Nevertheless, our initial experimental results show that the model would easily overfit the validation data but cannot perform well with unseen test data. The transformation model can easily align the training and validation data very well as the transformation model has a large mount of parameters. However, the relationship is too specific and does not generalize well on underlying test data. We observe that reducing model parameters of the transformation models or adding model regularization can alleviate the overfitting issue but it is still hard to achieve similar results with heuristic TRA strategies.

\begin{table*}[t]
\centering
\caption{List of all operations that our method can choose for TRA. We formulate TRA as a composition of $L$=15 types of operations, with each operation has $K$ choices of magnitudes. $K$ varies from 2 to 7.}\label{tabtraug}
\newsavebox{\tableboxtrauglist}
\begin{lrbox}{\tableboxtrauglist}

\begin{tabular}{m{15mm}<{\centering}m{12mm}<{\centering}m{32mm}<{\centering}m{66mm}<{\centering}m{35mm}<{\centering}}
\hlineB{3}
Category & Series ID & Operation Name & Description & Range of magnitudes \\
\hlineB{2}
\multirow{11}{*}{\tabincell{c}{Spatial \\ transformations}} & 0 & Scaling & Scale up/ down the image with factor 1+$M$ & [[0, 0.1), [0.1, 0.2), [0.2, 0.3), [0.3, 0.4), [0.4, 0.5)] \\
\cline{2-5}
& 1 & RotateFrontal & Rotate the image along frontal axis anticlockwise/ clockwise & [[0, 10$\degree$), [10$\degree$, 20$\degree$), [20$\degree$, 30$\degree$), 90$\degree$] \\
\cline{2-5}
& 2 & RotateSagittal & Rotate the image along sagittal axis anticlockwise/ clockwise & [[0, 10$\degree$), [10$\degree$, 20$\degree$), [20$\degree$, 30$\degree$), 90$\degree$] \\
\cline{2-5}
& 3 & RotateLongitudinal & Rotate the image along longitudinal axis anticlockwise/ clockwise  & [[0, 10$\degree$), [10$\degree$, 20$\degree$), [20$\degree$, 30$\degree$), 90$\degree$] \\
\cline{2-5}
& 4 & MirroringSagittal & Flip the sample in sagittal planes & None \\
\cline{2-5}
& 5 & MirroringFrontal & Flip the sample in frontal planes & None \\
\cline{2-5}
& 6 & MirroringAxial & Flip the sample in axial planes & None \\
\hline
\multirow{6}{*}{\tabincell{c}{Intensity \\ transformations}} & 7 & Gamma correction & Scale $\boldsymbol{x}$ to [0,1], then $\mathcal{O}$($\boldsymbol{x}$) = $\boldsymbol{x}$$^{(1 + \gamma)^{\pm 1}}$, and scale it back & [[0, 0.2), [0.2, 0.4), [0.4, 0.6)] \\
& 8 & Inverted gamma correction & Do the gamma correction with the inverted image & [[0, 0.2), [0.2, 0.4), [0.4, 0.6)] \\
\cline{2-5}
& 9 & Shifting intensity histogram & $\mathcal{O}$($\boldsymbol{x}$) = $\boldsymbol{x}$ $\pm$ shift & [[0, 0.1), [0.1, 0.2), [0.2, 0.3)] \\
& 10 & Scaling intensity histogram & $\mathcal{O}$($\boldsymbol{x}$) = $\boldsymbol{x}$ * (1 + scale)$^{\pm 1}$ & [[0, 0.1), [0.1, 0.2), [0.2, 0.3)] \\
& 11 & Contrast & Reduce the image mean, then $\mathcal{O}$($\boldsymbol{x}$) = $\boldsymbol{x}$ * (1 + scale)$^{\pm 1}$, and add the mean value back & [[0, 0.1), [0.1, 0.2), [0.2, 0.3)] \\
\hline
\multirow{6}{*}{\tabincell{c}{Noise \\ transformations}} & \multirow{4}{*}{12} & Blurring & Blur the image using Gaussian filter with different standard deviations & [[0.4, 0.6), [0.6, 0.8), [0.8, 1)] \\
& & Sharpening & Sharpen the image using Laplacian of Gaussian filter with different standard deviations & [[0.8, 1), [0.6, 0.8), [0.4, 0.6)] \\
\cline{2-5}
& 13 & Adding Gaussian noise & Add Gaussian noise with different standard deviations & [[0, 0.05), [0.05, 0.10), [0.10, 0.15)] \\
\cline{2-5}
& 14 & Simulating low resolution & Scale down the image, and scale it back & [[0.8, 1), [0.6, 0.8), [0.4, 0.6)] \\
\hline
\end{tabular}
\end{lrbox}
\scalebox{1}{\usebox{\tableboxtrauglist}}
\end{table*}

\subsection{List of Transformations for Test-Time Data Augmentation}

We list all the operations $\mathcal{O}$ we use for TEA in Table~\ref{tabteaug}. Note that the operations always have deterministic magnitudes. In this way, we can make sure we would apply the same set of transformation to a test sample.

\begin{table*}[t]
\centering
\caption{List of all operations that our method can choose for TEA. There are 24 types of operations with a total of $K$=84 choices.}\label{tabteaug}
\newsavebox{\tableboxauglist}
\begin{lrbox}{\tableboxauglist}

\begin{tabular}{m{15mm}<{\centering}m{18mm}<{\centering}m{40mm}<{\centering}m{61mm}<{\centering}m{35mm}<{\centering}}
\hlineB{3}
Category & Operation ID & Operation Name & Description & Range of magnitudes \\
\hlineB{2}
Identity & 0 &  Identity & No augmentation, $\mathcal{O}$($\boldsymbol{x}$) = $\boldsymbol{x}$  & None \\
\hline
\multirow{11}{*}{\tabincell{c}{Spatial \\ transformations}} 
& 1,2,3,4,5 & Scaling down & Scale down the image by factor 1 + $M$ & [0.05, 0.15, 0.25, 0.35, 0.45] \\
\cline{2-5}
& 6,7,8,9,10 & Scaling up & Scale up the image by factor 1 + $M$ & [0.05, 0.15, 0.25, 0.35, 0.45] \\
\cline{2-5}
& 11,12,13,14,15 & RotateFrontal ACW & Rotate the image along frontal axis anticlockwise & [5$\degree$, 15$\degree$, 25$\degree$, 90$\degree$, 180$\degree$] \\
\cline{2-5}
& 16,17,18,19 & RotateFrontal CCW & Rotate the image along frontal axis clockwise & [5$\degree$, 15$\degree$, 25$\degree$, 90$\degree$] \\
\cline{2-5}
& 20,21,22,23,24 & RotateSagittal ACW & Rotate the image along sagittal axis anticlockwise & [5$\degree$, 15$\degree$, 25$\degree$, 90$\degree$, 180$\degree$] \\
\cline{2-5}
& 25,26,27,28 & RotateSagittal CCW & Rotate the image along sagittal axis clockwise & [5$\degree$, 15$\degree$, 25$\degree$, 90$\degree$] \\
\cline{2-5}
& 29,30,31,32,33 & RotateLongitudinal ACW & Rotate the image along longitudinal axis anticlockwise & [5$\degree$, 15$\degree$, 25$\degree$, 90$\degree$, 180$\degree$] \\
\cline{2-5}
& 34,35,36,37 & RotateLongitudinal CCW & Rotate the image along longitudinal axis clockwise  & [5$\degree$, 15$\degree$, 25$\degree$, 90$\degree$] \\
\cline{2-5}
& 38,39,40,41 & Mirroring & Flip the sample in different planes & [Sagittal, Frontal, Axial, All] \\
\hline
\multirow{12}{*}{\tabincell{c}{Intensity \\ transformations}} & 42,43,44 & Gamma expansion & Scale $\boldsymbol{x}$ to [0,1], then $\mathcal{O}$($\boldsymbol{x}$) = $\boldsymbol{x}$$^{1 + \gamma}$, and scale it back & [0.1, 0.3, 0.5] \\
\cline{2-5}
& 45,46,47 & Gamma compression & Scale $\boldsymbol{x}$ to [0,1], then $\mathcal{O}$($\boldsymbol{x}$) = $\boldsymbol{x}$$^{1/(1 + \gamma)}$, and scale it back & [0.1, 0.3, 0.5] \\
\cline{2-5}
& 48,49,50 & Inverted gamma expansion & Do the gamma compression with the inverted image & [0.1, 0.3, 0.5] \\
\cline{2-5}
& 51,52,53 & Inverted gamma compression & Do the gamma expansion with the inverted image & [0.1, 0.3, 0.5] \\
\cline{2-5}
& 54,55,56 & Adding intensity & $\mathcal{O}$($\boldsymbol{x}$) = $\boldsymbol{x}$ + shift & [0.05, 0.15, 0.25] \\
\cline{2-5}
& 57,58,59 & Subtracting intensity & $\mathcal{O}$($\boldsymbol{x}$) = $\boldsymbol{x}$ - shift & [0.05, 0.15, 0.25] \\
\cline{2-5}
& 60,61,62 & Scaling up intensity histogram & $\mathcal{O}$($\boldsymbol{x}$) = $\boldsymbol{x}$ * (1 + scale) & [0.05, 0.15, 0.25] \\
\cline{2-5}
& 63,64,65 & Scaling down intensity histogram & $\mathcal{O}$($\boldsymbol{x}$) = $\boldsymbol{x}$ / (1 + scale) & [0.05, 0.15, 0.25] \\
\cline{2-5}
& 66,67,68 & Increasing contrast & Reduce the image mean, then $\mathcal{O}$($\boldsymbol{x}$) = $\boldsymbol{x}$ * (1 + scale), and add the mean value back & [0.05, 0.15, 0.25] \\
\cline{2-5}
& 69,70,71 & Decreasing contrast & Reduce the image mean, then $\mathcal{O}$($\boldsymbol{x}$) = $\boldsymbol{x}$ / (1 + scale), and add the mean value back & [0.05, 0.15, 0.25] \\
\hline
\multirow{7}{*}{\tabincell{c}{Noise \\ transformations}} & 72,73,74 & Blurring & Blur the image using Gaussian filter with different standard deviations & [0.5, 0.7, 0.9] \\
\cline{2-5}
& 75,76,77 & Sharpening & Sharpen the image using Laplacian of Gaussian filter with different standard deviations & [0.9, 0.7, 0.5] \\
\cline{2-5}
& 78,79,80 & Adding Gaussian noise & Add Gaussian noise with different standard deviations & [0.025, 0.075, 0.125] \\
\cline{2-5}
& 81,82,83 & Simulating low resolution & Scale down the image, and scale it back & [0.9, 0.7, 0.5] \\
\hline
\end{tabular}
\end{lrbox}
\scalebox{0.95}{\usebox{\tableboxauglist}}
\end{table*}

\subsection{Choices of Validation Loss for Different Settings}

We find $\loss_{val}$ should be chosen differently for different settings to achieve optimal results. $\loss_{val}$ can be cross entropy (CE), soft Dice similarity coefficient (DSC) or a combination of the two loss functions. We summarize the optimal choices for different settings in Table~\ref{tablelval}.

\begin{table*}[ht]
\centering
\caption{Choices of the validation loss function and initial learning rate for optimizing TRA with different settings.}\label{tablelval}
\newsavebox{\tablelval}
\begin{lrbox}{\tablelval}

\begin{tabular}{m{18mm}<{\centering}|m{15mm}<{\centering}|m{15mm}<{\centering}|m{15mm}<{\centering}|m{15mm}<{\centering}|m{15mm}<{\centering}}
\hlineB{3}
Dataset & Number of class & Model & Training data & $\loss_{val}$ & $\beta$
\\
\hlineB{1}
\multirow{5}{*}{\tabincell{c}{ATLAS}}& \multirow{5}{*}{\tabincell{c}{2}} &  \multirow{2}{*}{\tabincell{c}{DeepMedic}} & 50\% & DSC & 1e-3 \\
& &  & 100\% & DSC & 2e-3 \\
& & \multirow{2}{*}{\tabincell{c}{3D U-Net}} & 50\% & DSC & 5e-4 \\
& &  & 100\% & CE & 1e-3 \\
& & nnFormer & 50\% & DSC & 1e-3 \\
\hline
\multirow{5}{*}{\tabincell{c}{KiTS}}& \multirow{5}{*}{\tabincell{c}{3}} & \multirow{2}{*}{\tabincell{c}{DeepMedic}} & 50\% & DSC & 1e-3 \\
& &  & 100\% & DSC & 1e-3 \\
& & \multirow{2}{*}{\tabincell{c}{3D U-Net}} & 50\% & DSC+CE & 5e-4 \\
& &  & 100\% & CE & 5e-4 \\
& & nnFormer & 50\% & CE & 2e-3 \\
\hline
\multirow{2}{*}{\tabincell{c}{Organ}}&  \multirow{2}{*}{\tabincell{c}{14}} & DeepMedic & 100\% & CE & 2e-3 \\
& & 3D U-Net & 100\% & DSC & 5e-4 \\
\hline
\multirow{2}{*}{\tabincell{c}{Prostate}}& \multirow{2}{*}{\tabincell{c}{2}} & DeepMedic & 100\% & CE & 5e-4 \\
& & 3D U-Net & 100\% & CE & 5e-4 \\
\hline
Cardiac (Cross-sequence) & 4 & 3D U-Net & 100\% & CE & 5e-4 \\
\hline
Cardiac (Cross-site) & 4 & 3D U-Net & 100\% & DSC & 5e-4 \\
\hline
\end{tabular}
\end{lrbox}
\scalebox{1}{\usebox{\tablelval}}
\end{table*}

\subsection{The Policies and Results of RandAugment}

\begin{figure*}[ht]
\centering
\includegraphics[width=\textwidth]{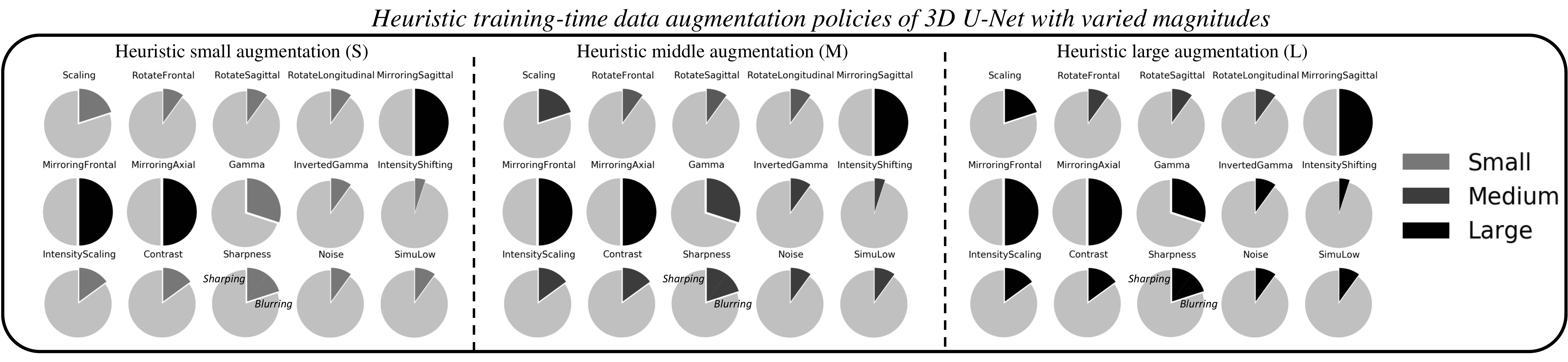}

\caption{The sampling distribution for three heuristic TRA policies which are created by only changing the transformation magnitudes, referring to~\cite{cubuk2019randaugment}.} \label{figa_predefined}
\end{figure*}

We implement RandAugment~\cite{cubuk2019randaugment} based on the heuristic TRA policies of nnU-Net~\cite{isensee2021nnu}. Specifically, we keep the probability of adopting operations and substitute all the heuristic operations within the same type with the ones having certain magnitude. We create three RandAugment policies based on varied magnitudes, denoted as RandAugment-S, RandAugment-M and RandAugment-L. The sampling distribution of these TRA policies are shown in Fig.~\ref{figa_predefined}.

We summarize the segmentation results when trained with RandAugment for kidney and kidney tumor segmentation based on 3D U-Net with 50$\%$ training data in Table~\ref{tabkitsrand}. We find that RandAugment with large transformation magnitudes performs better on kidney tumor segmentation. The proposed class-specific TRA can perform better than all the RandAugment variants.

\begin{table*}[t]
\centering
\caption{Evaluation of kidney and kidney tumor segmentation based on 3D U-Net with 50\% training data using different data augmentation methods. Best and second best results are in \textbf{bold}, with best also \underline{\textbf{underlined}}.}\label{tabkitsrand}
\newsavebox{\tableboxkitsrand}
\begin{lrbox}{\tableboxkitsrand}
\begin{tabular}{c|c|m{8mm}<{\centering}|m{8mm}<{\centering}|m{8mm}<{\centering}|m{8mm}<{\centering}|m{16mm}<{\centering}|m{8mm}<{\centering}|m{8mm}<{\centering}|m{8mm}<{\centering}|m{8mm}<{\centering}}
\hlineB{3}
\multirow{2}{*}{\tabincell{c}{Training-time \\ data augmentation}} & \multirow{2}{*}{\tabincell{c}{Test-time \\ data augmentation}} & \multicolumn{4}{c|}{Kidney} & \multicolumn{4}{c|}{Tumor} & \multirow{2}{*}{\tabincell{c}{AVG \\ Rank $\downarrow$}} \\
& & DSC $\uparrow$ & SEN $\uparrow$ & PRC $\uparrow$ & HD $\downarrow$ & DSC $\uparrow$ & SEN $\uparrow$ & PRC $\uparrow$ & HD $\downarrow$ &  \\ 
\hlineB{1}
None &  None & 95.3 & 94.0 & \underline{\textbf{97.3}} & 5.7 & 43.5 & 39.5 & 60.9 & 104.6 & 7.0 \\
Heuristic~\cite{isensee2021nnu} & None & \textbf{96.6} & \textbf{96.4} & 96.9 & \textbf{2.6} & 76.6 & 80.2 & 77.4 & \textbf{40.6} & 3.8 \\
RandAugment-S~\cite{cubuk2019randaugment} & None & 96.4 & 96.0 & 97.0 & 2.7 & 74.4 & 76.6 & \textbf{78.3} & 45.5 & 4.3 \\
RandAugment-M~\cite{cubuk2019randaugment} & None & 96.4 & 95.7 & \textbf{97.2} & 2.8 & 77.5 & 79.0 & \underline{\textbf{79.7}} & \textbf{\underline{34.2}} & \textbf{2.5} \\
RandAugment-L~\cite{cubuk2019randaugment} & None & 96.4 & 96.0 & 97.0 & 2.7 & \textbf{77.6} & \textbf{82.1} & 77.0 & 61.4 & 3.8 \\
Learned~\cite{cubuk2018autoaugment, lim2019fast, li2020dada} & None & 96.5 & 96.3 & 96.8 & 2.8 & 76.7 & 82.0 & 76.1 & 55.7 & 4.5 \\
\textbf{Learned Class-Specific} & None & \textbf{\underline{96.8}} & \underline{\textbf{96.6}} & 96.9 & \textbf{\underline{2.5}} & \textbf{\underline{78.4}} (+1.8)$^\sim$ & \underline{\textbf{82.2}} & 78.0 & 47.2 & \textbf{\underline{2.3}} \\
\hline
\end{tabular}
\end{lrbox}
\scalebox{1}{\usebox{\tableboxkitsrand}}

{\raggedright \quad $^*p$-value $<$ 0.05; $^{**}p$-value $<$ 0.01; $^\sim p$-value $	\geq$ 0.05 (compared to Heuristic$^\ddagger$ TRA w/o TEA) \par}

\end{table*}

\clearpage

\clearpage

\subsection{Learned Policies with Different Settings}

We summarize all the learned policies for different tasks based on different models in this section. We expect it to be taken as a reference for participants to collect datasets and design transformations for data augmentation. The data augmentation policies for different network architectures are initialized with the heuristic policies provided by these frameworks. These default policies are designed for general purpose and specifically to fit the properties of different segmentation models. Therefore, the learned policies also differ considerably between models. 

\subsubsection{ATLAS}

\begin{figure*}[t]
\centering
\includegraphics[width=\textwidth]{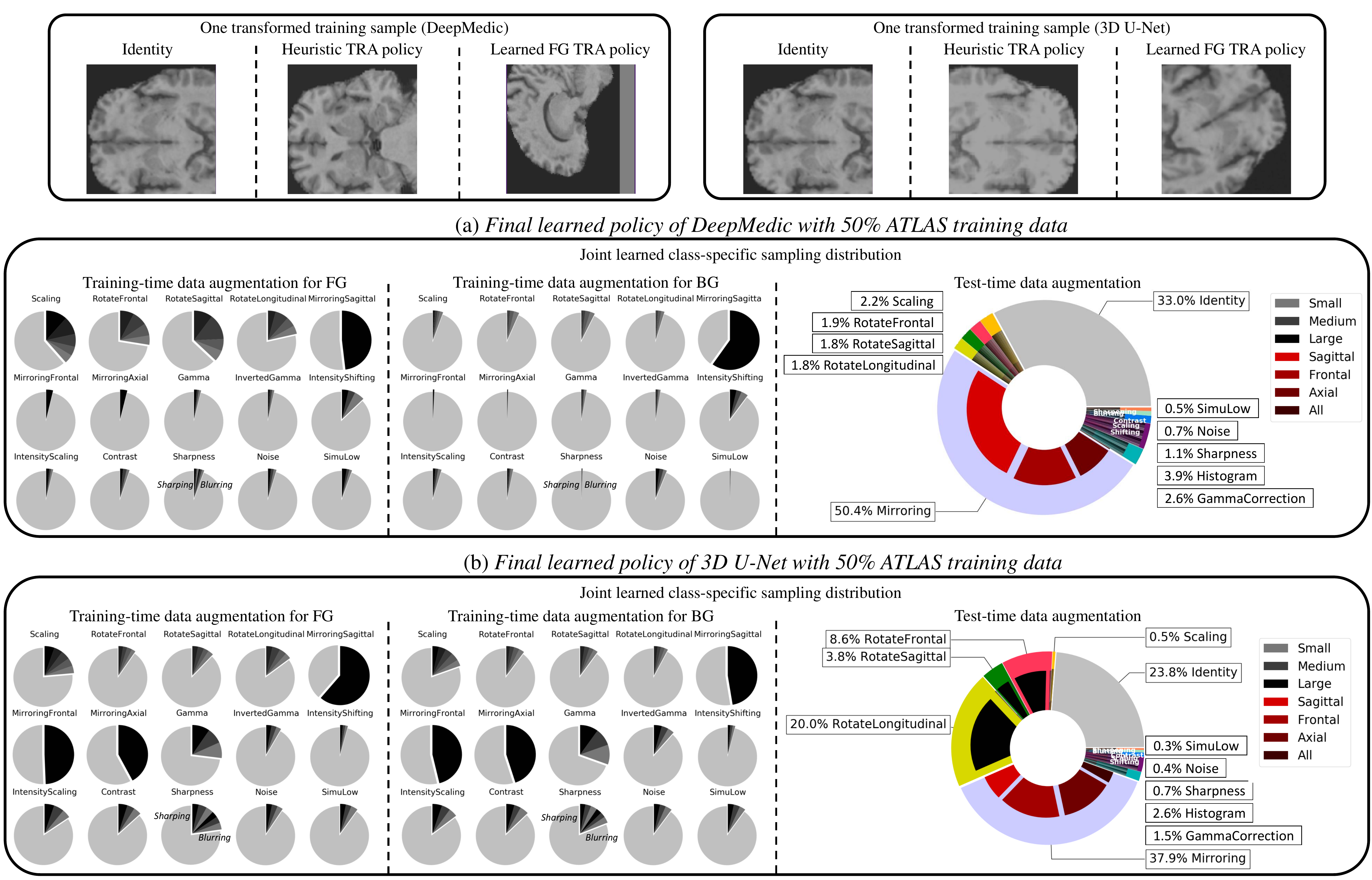}

\caption{The learned probability distributions over augmentations based on different segmentation models for brain stroke lesion segmentation with 50\% ATLAS training data. We also visualize an example of the transformed foreground training sample with different sampling distributions for TRA.} \label{figa_atlas50}
\end{figure*}

\begin{figure*}[t]
\centering
\includegraphics[width=\textwidth]{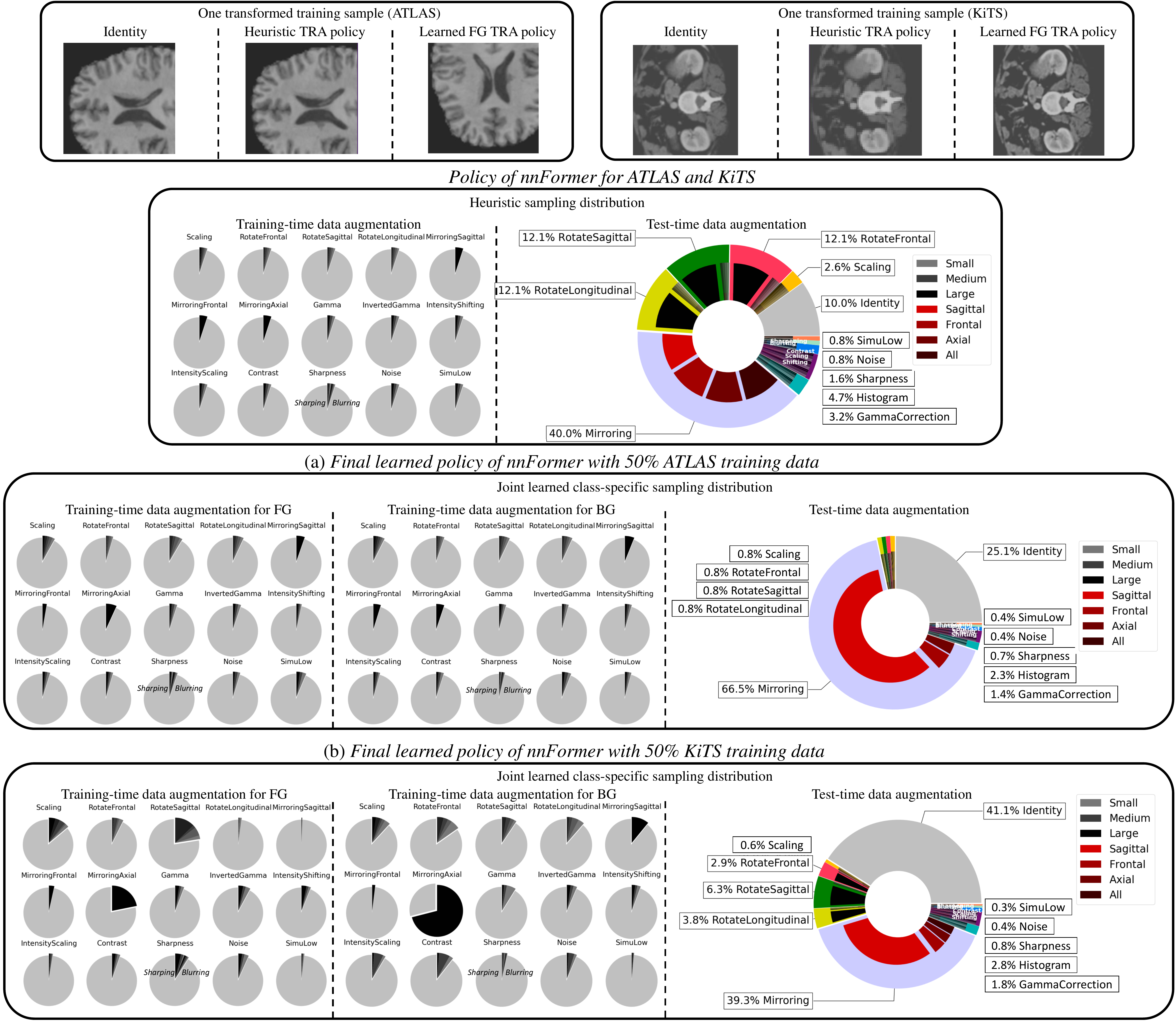}

\caption{The learned probability distributions over augmentations based on nnFormer for brain stroke lesion segmentation with 50\% ATLAS training data and 50\% KiTS training data. We also visualize an example of the transformed foreground training sample with different sampling distributions for TRA.} \label{figa_nnformer}
\end{figure*}

We summarize the learned data augmentation policies for brain stroke lesion segmentation with 50\% ATLAS training data in Fig.~\ref{figa_atlas50} and Fig.~\ref{figa_nnformer}. As brain stroke lesion is relatively small and often under-represented, we find the learned TRA policies tend to apply transformations with larger magnitude to foreground samples and transformations with smaller magnitude to background samples. Specifically, when compared with heuristic policies, the learned policies would increase the probabilities of adopting transformations for foreground samples while decrease these probabilities for background samples. We observe consistent changes in all kinds of transformations. This indicates that the segmentation models would benefit from foreground samples with more variances and background samples with limited transformations.

We find the probabilities of spatial transformations such as scaling and rotation would be largely increased. This indicates that spatial transformations make more differences to the training data distributions. We also find that the policies learned with 100$\%$ training data would often adopt larger transformations when compared with the ones learned with 50$\%$ training data. This indicates that the optimal TRA policies vary for training datasets with different amounts of samples. Specifically, we should utilize larger transformations to effectively extend the data distributions with sufficient training data. It may be because small transformations are hard to add more information to the training datasets on the three top of sufficient training data.

We find the learned TEA policies for DeepMedic increase the probabilities of flipping in sagittal planes. It might be because the initialized TRA policy for DeepMedic has large probabilities of flipping in the sagittal planes (which is designed by taking into account the symmetrical brain structure). We also notice the learned TEA policies would always largely increase the probabilities of identity for both DeepMedic and 3D U-Net. This indicates that the predictions of images in the original data distribution are fairly accurate.

\subsubsection{KiTS}

\begin{figure*}[t]
\centering
\includegraphics[width=\textwidth]{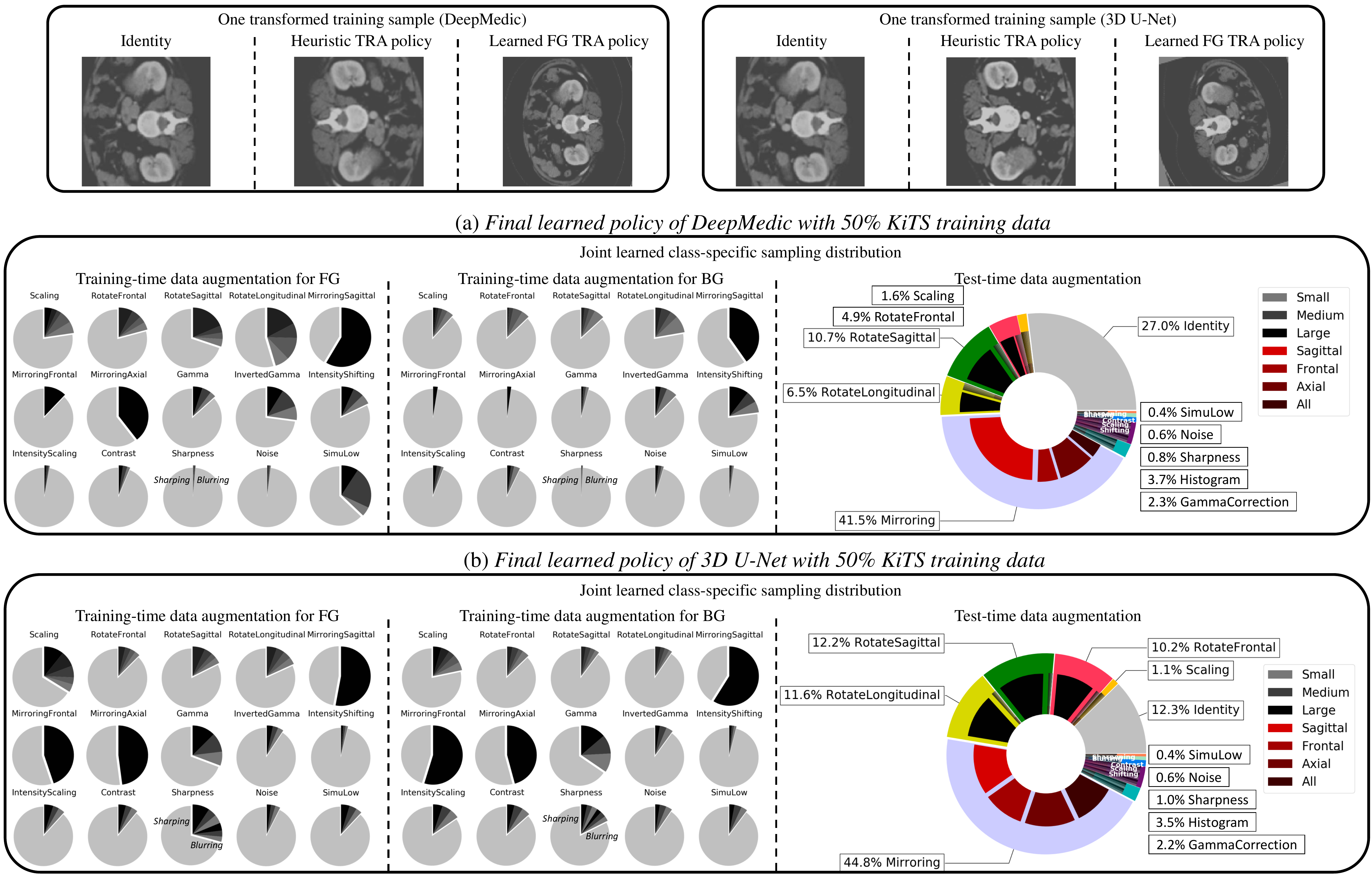}

\caption{The learned probability distributions over augmentations based on different segmentation models for kidney and kidney tumor segmentation with 50\% KiTS training data. We also visualize an example of the transformed foreground training sample with different sampling distributions for TRA.} \label{figa_kits50}
\end{figure*}

\begin{figure*}[t]
\centering
\includegraphics[width=\textwidth]{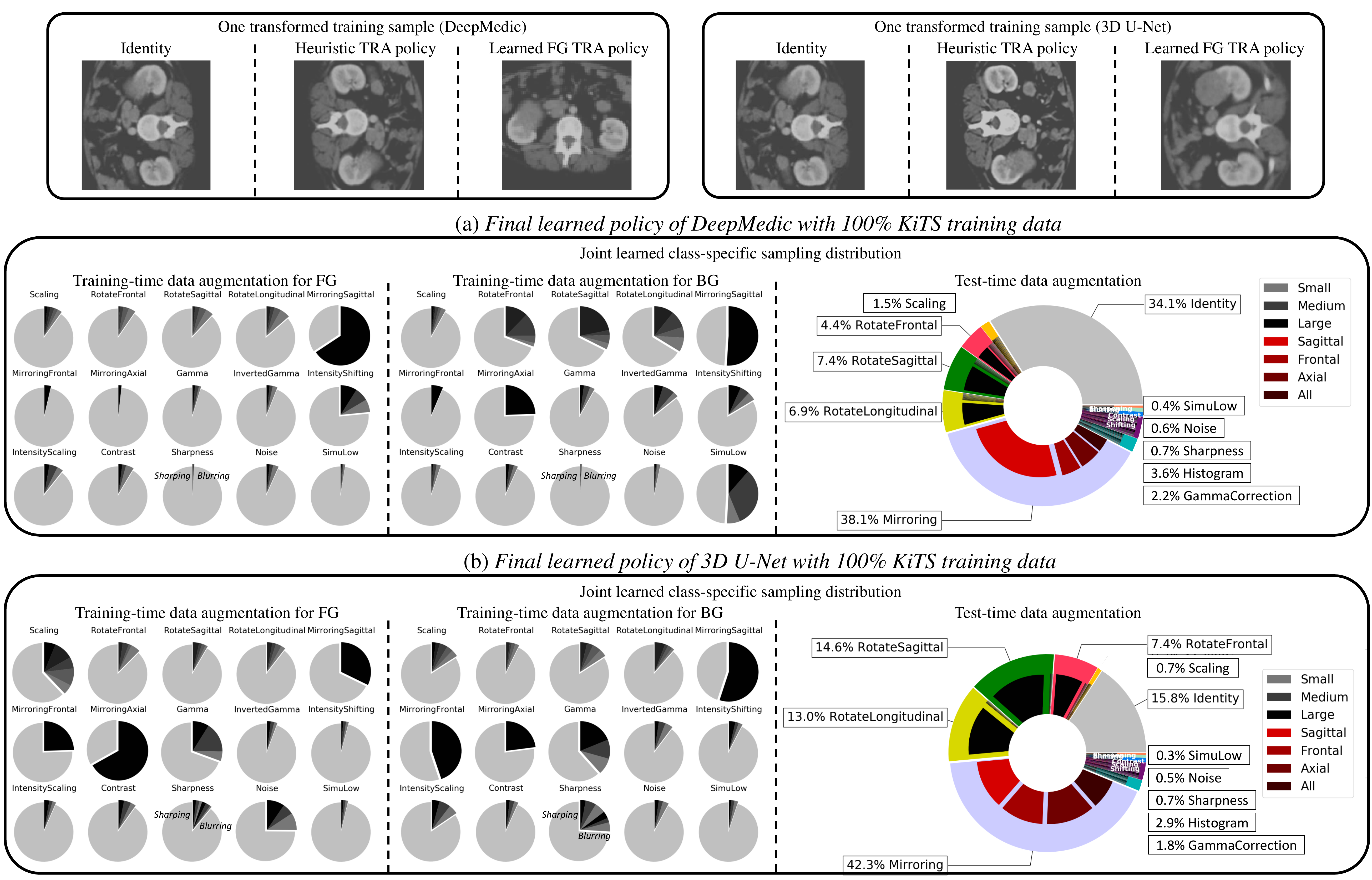}

\caption{The learned probability distributions over augmentations based on different segmentation models for kidney and kidney tumor segmentation with 100\% KiTS training data. We also visualize an example of the transformed foreground training sample with different sampling distributions for TRA.} \label{figa_kits100}
\end{figure*}

We summarize the learned data augmentation policies for kidney and kidney tumor segmentation with 50\% KiTS training data in Fig.~\ref{figa_kits50} and Fig.~\ref{figa_nnformer}. We also summarize the policies learned with 100\% KiTS training data in Fig.~\ref{figa_kits100}. We find when the segmentation models are trained with 50$\%$ training data, the learned TRA policies tend to generate larger transformations for foreground samples, similar to the case of brain stoke lesion segmentation. This might be because the kidney and kidney tumor are under-represented with less training data. In contrast, the policies learned with 100$\%$ training data do not have consistent bias towards the foreground samples. This may be because the class imbalance problem probably would not affect the learning process too much as the training datasets contain a sufficient amount of foreground samples. Under such condition, the policies are learned to generate class-specific transformations. Specifically, the learned policies increase the probabilities of scaling for foreground samples while increase the probabilities of noise transformations such as sharpening and simulating low resolution for background samples.

When compared with brain stroke lesion segmentation in T1-weighted MR images, we find the learned policies is prone to adopt intensity transformations such as gamma correction and intensity shifting and noise transformations such as adding Gaussian noise and simulating low resolution for TRA for kidney and kidney tumor segmentation in CT images. This is may because the objects in CT images have low contrast and blurred boundaries. In this case, the simulated images with varied imaging quality can help the segmentation model generalize better with unseen image conditions.

\subsubsection{Abdominal organ}

\begin{figure*}[t]
\centering
\includegraphics[width=\textwidth]{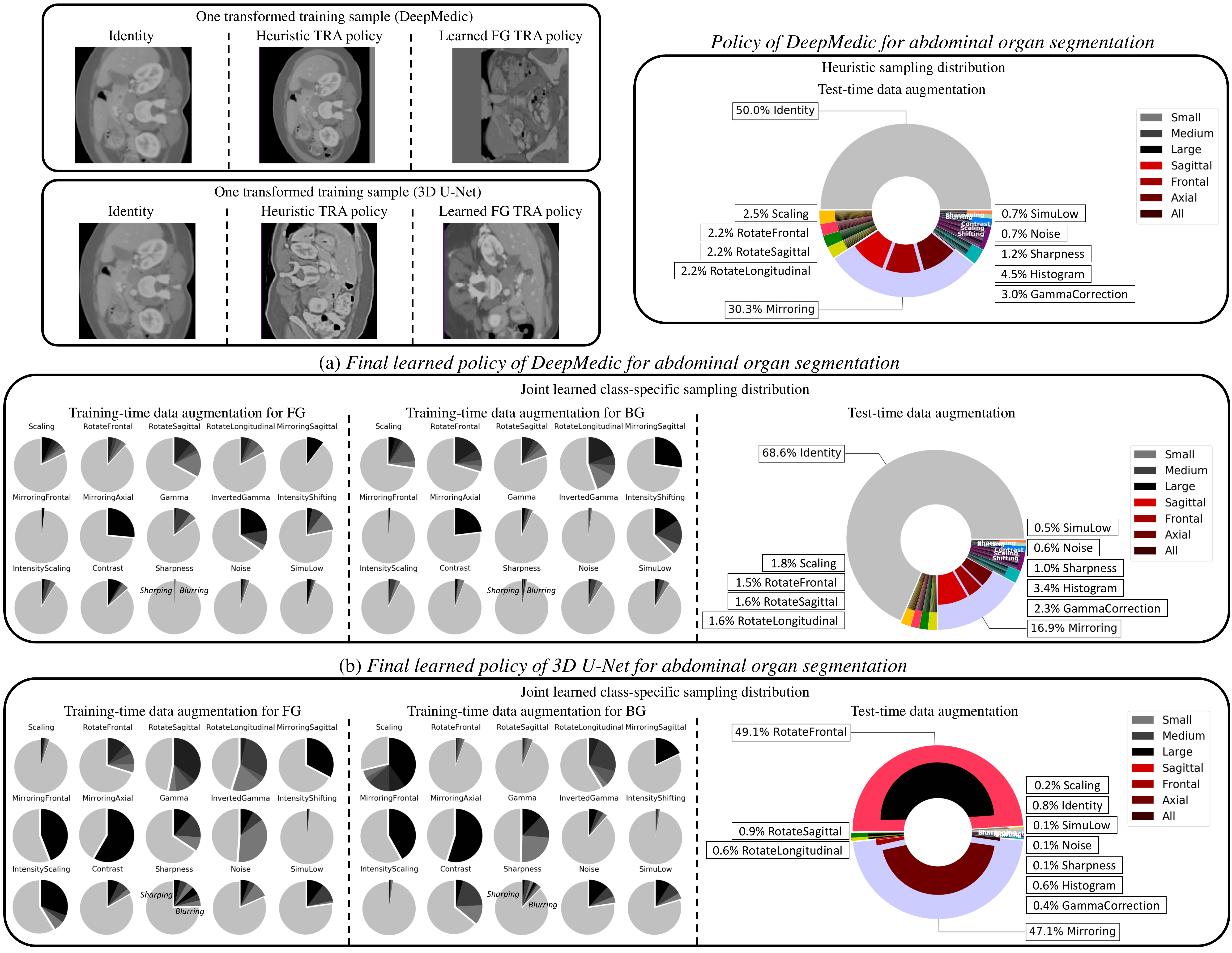}

\caption{The learned probability distributions over augmentations based on different segmentation models for abdominal organ segmentation. We also visualize an example of the transformed foreground training sample with different sampling distributions for TRA.} \label{figa_abdomen}
\end{figure*}

We summarize the learned data augmentation policies for abdominal organ segmentation in Fig.~\ref{figa_abdomen}. The learned policies for abdominal organ segmentation are quite different from the cases of brain stroke lesion and kidney tumor segmentation. This is may due to the complexity of foreground class which contains many different classes of abdominal organs. We find the learned TRA policies are prone to adopt large scaling transformations to the background samples. This maybe because some background objects which are similar to the foreground objects vary in size. The segmentation models can perform better when learned simulated background objects with varied scales. Similar to the case of kidney tumor segmentation in CT images, the learned TRA policies tend to adopt many intensity and noise transformations. We think this is also related to the low imaging quality of CT.

We find the default TEA policy would decrease the performance of the segmentation model, therefore we utilize and initialize with a TEA policy with increased probability of identity, as shown in the upper part of Fig.~\ref{figa_abdomen}. The learned TEA policy for DeepMedic further increases the probability of identity. This indicates that DeepMedic cannot perform well with transformed images. This is because the initialized TRA policies for DeepMedic do not contain large transformations. This could be also related to the network architecture of DeepMedic which contains more convolutional layers in the original size. The network design drives DeepMedic to make predictions relying more on the local features, potentially being more sensitive to noise.

\subsubsection{Cross-site prostate}

\begin{figure*}[t]
\centering
\includegraphics[width=\textwidth]{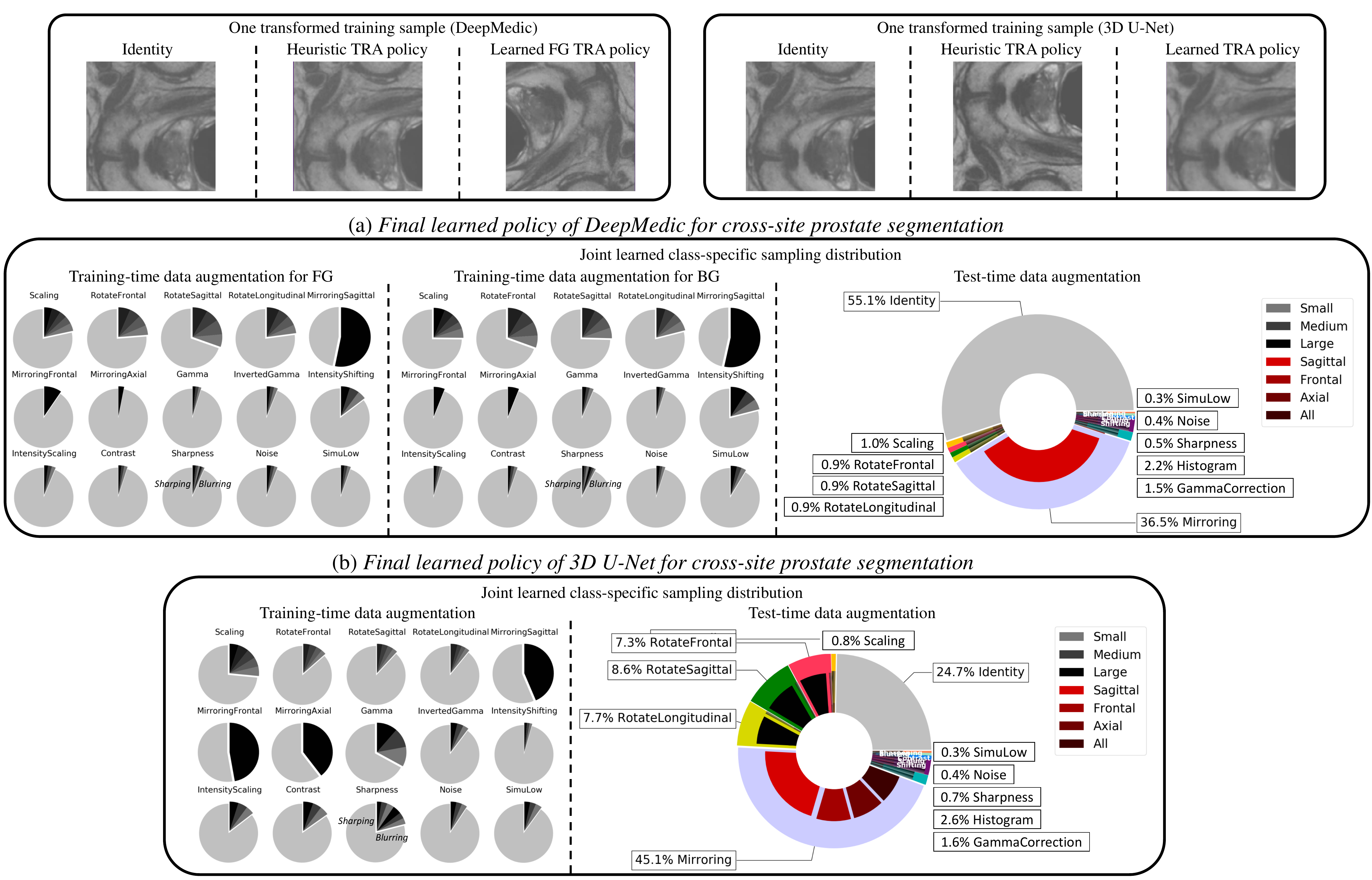}

\caption{The learned probability distributions over augmentations based on different segmentation models for cross-site prostate segmentation. We also visualize an example of the transformed foreground training sample with different sampling distributions for TRA.} \label{figa_prostate}
\end{figure*}

We summarize the learned data augmentation policies for cross-site prostate segmentation in Fig.~\ref{figa_prostate}. We find the learned TRA policies are generally very close to the initialized ones. This indicates that the default TRA policies fit this task well. We notice the learned TRA policies for background adopt more intensity transformations such as intensity shifting and noise transformations such as simulating low resolution. Those transformations might help align the MRI datasets acquired with different settings.

\subsubsection{Cross-sequence and cross-site cardiac}

\begin{figure*}[t]
\centering
\includegraphics[width=\textwidth]{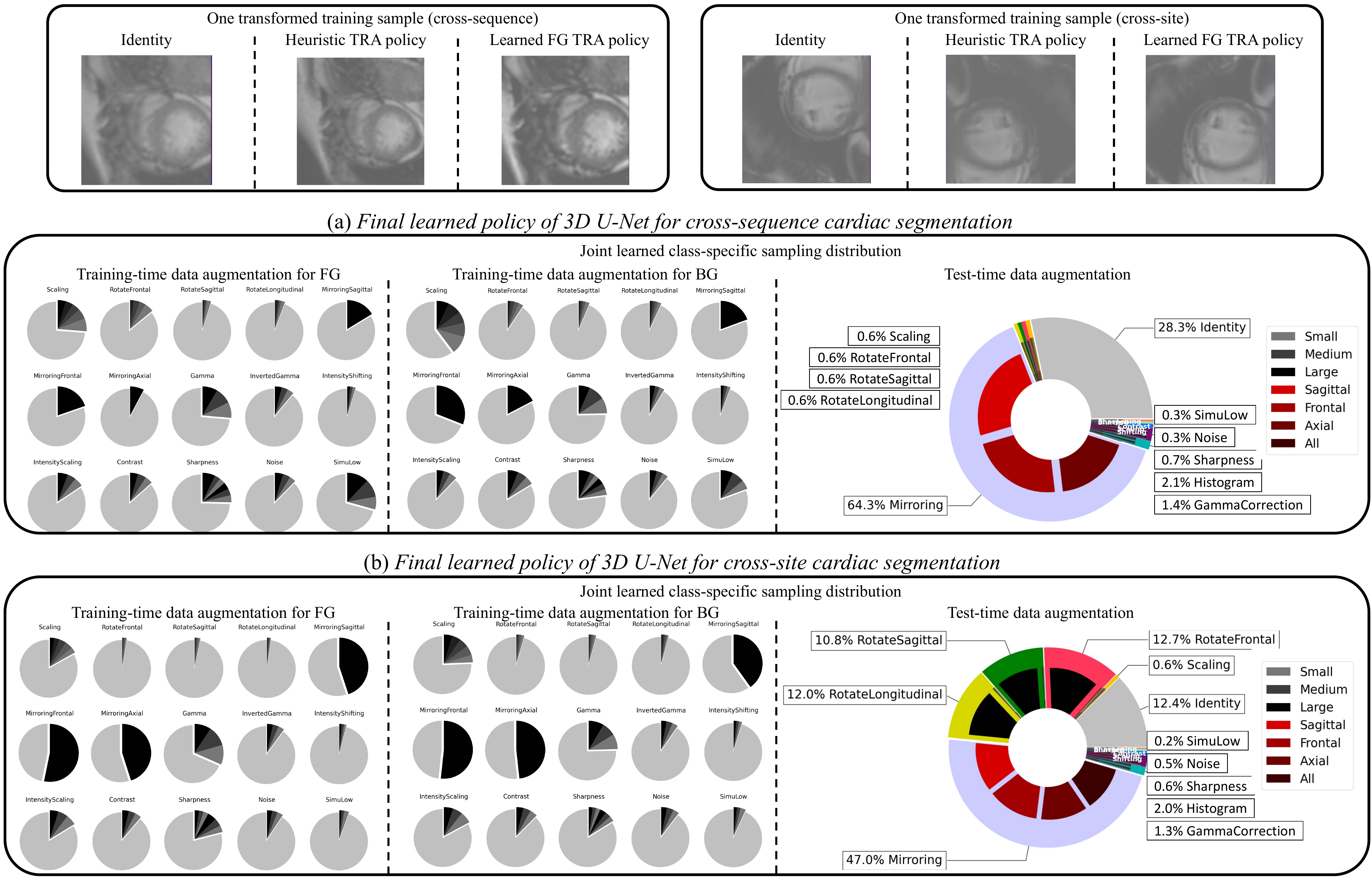}

\caption{The learned probability distributions over augmentations based on 3D U-Net for (a) cross-sequence or (b) cross-site cardiac segmentation. We also visualize an example of the transformed foreground training sample with different sampling distributions for TRA.} \label{figa_cardiac}
\end{figure*}

We summarize the learned data augmentation policies for cross-sequence or cross-site cardiac segmentation in Fig.~\ref{figa_cardiac}. Experimental details can be found in Section~\ref{sec:seq-cardiac} and Section~\ref{sec:seq-cardiac2}. We find in both cases the learned TRA polices would select large intensity transformation for FG samples while choose large spatial transformations for BG samples. This might indicate that style transformation of FG samples could help the model generalize better across domains for cardiac MR images. This finding is consistent with previous studies which are based on 2D networks~\cite{ouyang2021causality}.

\clearpage

\subsection{Logit Map Distributions}

\begin{figure}[t]
\centering
\includegraphics[width=0.48\textwidth]{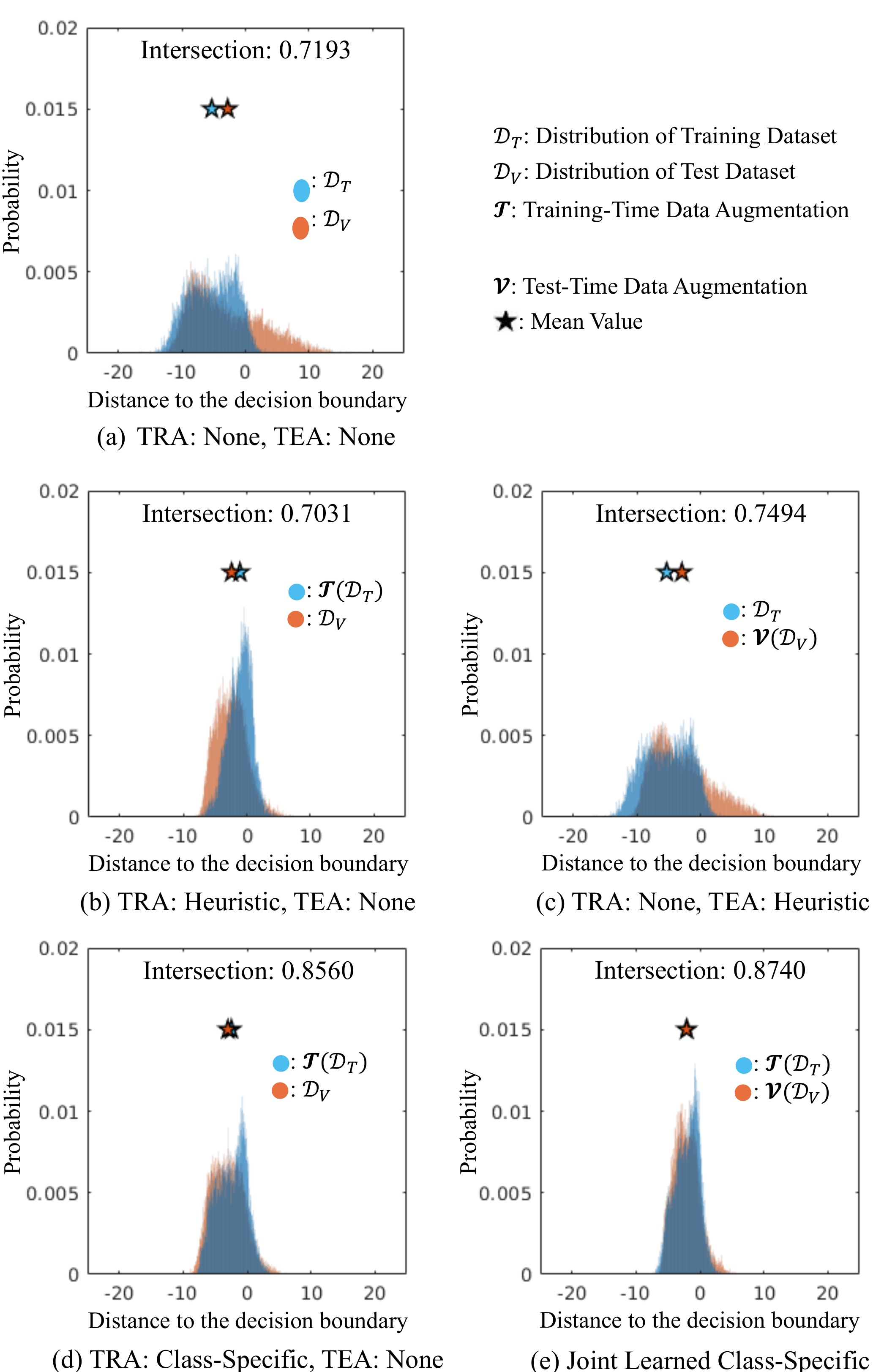}

\caption{The histograms of activations of the classification layer when processing training (blue) and test (orange) data of ATLAS with DeepMedic using different data augmentation methods. We visualize the data distributions as the distance of lesion logits to the decision boundary, which is calculated as $(z_1-z_2)/\sqrt{2}$ (logit $z_1$ for background and logit $z_2$ for lesion). TRA and TEA increase the intersection regions of the training and test data distributions. Our proposed methods can make the distribution overlap more. } \label{figa_logitmapFG}
\end{figure}

In order to illustrate the effectiveness of data augmentation methods, we visualize the activations of classification layer when models are trained and deployed under different conditions. We summarize the histograms of logit distributions when processing training and test samples of ATLAS with DeepMedic trained with 100\% training data in Fig.~\ref{figa_logitmapFG}. Specifically, we calculate the distance of logits to the decision boundary with $(z_1-z_2)/\sqrt{2}$, where $z_1$ is the logit for background and $z_2$ is the logit for lesion. To simplify the observations, we only monitor the logit distributions of lesion samples. 

We calculate the intersection regions of training and test distributions and summarize them on the top of each figure. TRA and TEA both improve the model performance by aligning the training and test data distributions, when compared with model trained without data augmentation (c.f. Fig.~\ref{figa_logitmapFG}(a)). However, they are not always optimal for the given datasets, as shown in Fig.~\ref{figa_logitmapFG}(b, c). Our proposed methods learned application-specific and class-specific TRA, therefore fit the given tasks better when compared with heuristic TRA, as shown in Fig.~\ref{figa_logitmapFG}(d). The joint optimization of TRA and TEA (c.f. Fig.~\ref{figa_logitmapFG}(e)) drive the data distributions to overlap more, thus generalize better.

\begin{figure*}[t]
\centering
\includegraphics[width=\textwidth]{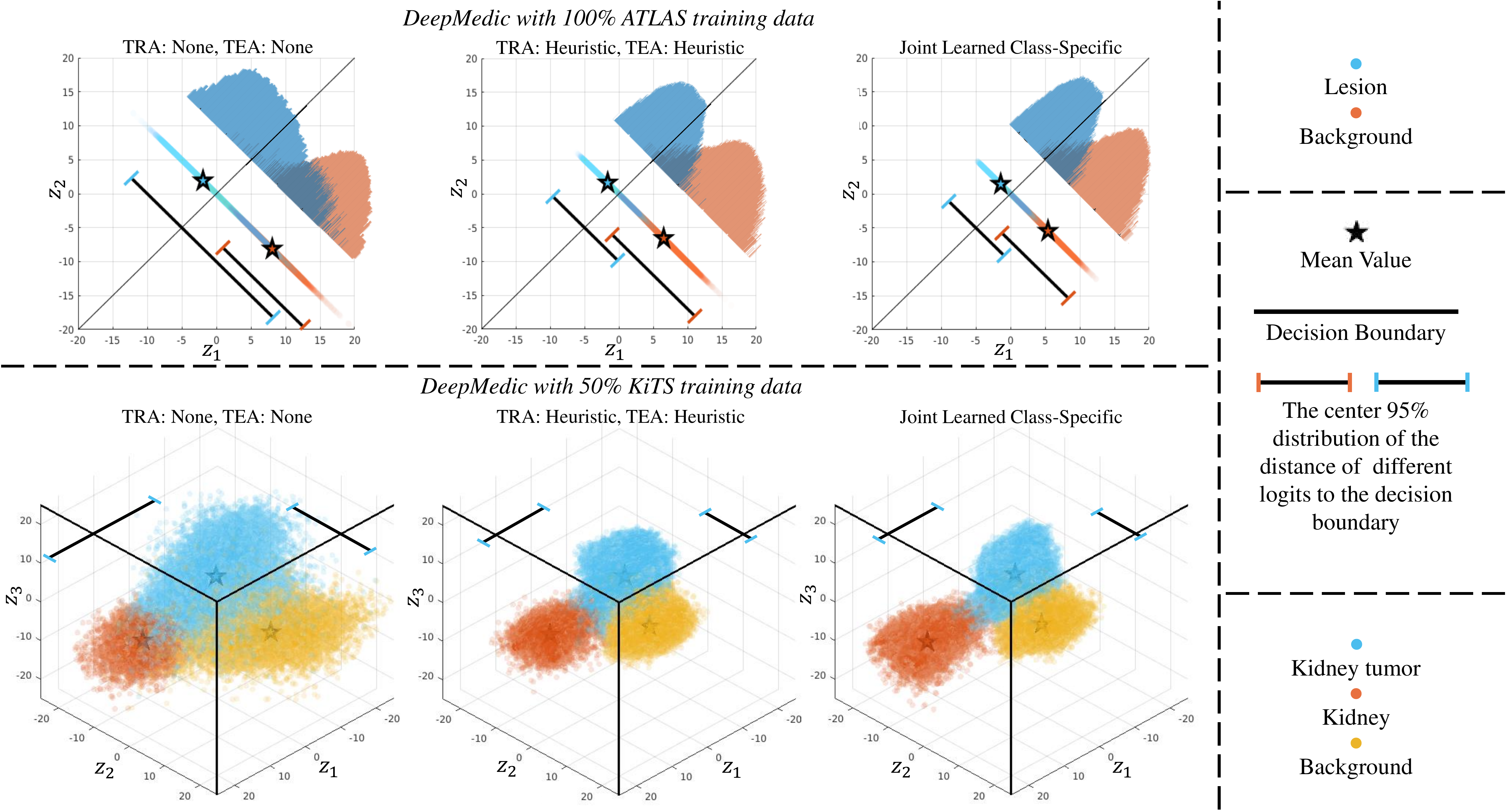}

\caption{(Uppr part) Activations of the classification layer (logit $z_1$ for background and logit $z_2$ for lesion) when processing the lesion (blue) or background samples (orange) of test data using different data augmentation methods. (Lower part) Activations of the classification layer (logit $z_1$ for backround, logit $z_2$ for kidney and logit $z_3$ for kidney tumor) when processing the kidney tumor (blue), kidney (yellow) and background samples (orange) of test data using different data augmentation methods. Compared with training without data augmentation or heuristic data augmentation, the proposed framework can make the model better separate the samples from different classes.} \label{figa_logitmap}
\end{figure*}

To obtain a better understanding of network behaviour, we also investigate the network activations when processing test samples of different classes. The logit distributions of DeepMedic trained with 100\% ATLAS training data and 50\% KiTS training data are both summarized in Fig.~\ref{figa_logitmap}. We find that data augmentation can always help the model build better representation, especially for the minority classes such as lesion and kidney tumor. Without data augmentation, the model would map the samples from the minority class across the decision boundary, causing false negatives. We observe that heuristic data augmentation can help the model map the logits of the samples from the minority classes away from the decision boundary. The proposed data augmentation strategy can help the model further reduce logit shifts of the minority classes and build a better decision boundary. Moreover, it can make the model map the logits of samples from the same class to a more compact cluster. This indicates that the model trained with our method can implicitly encourage the feature to have better inter-class separability and intra-class compactness, thus generalize well.

\clearpage

\subsection{The Decrease of Validation Loss During the Training Process}

\begin{figure}[t]
\centering
\includegraphics[width=0.48\textwidth]{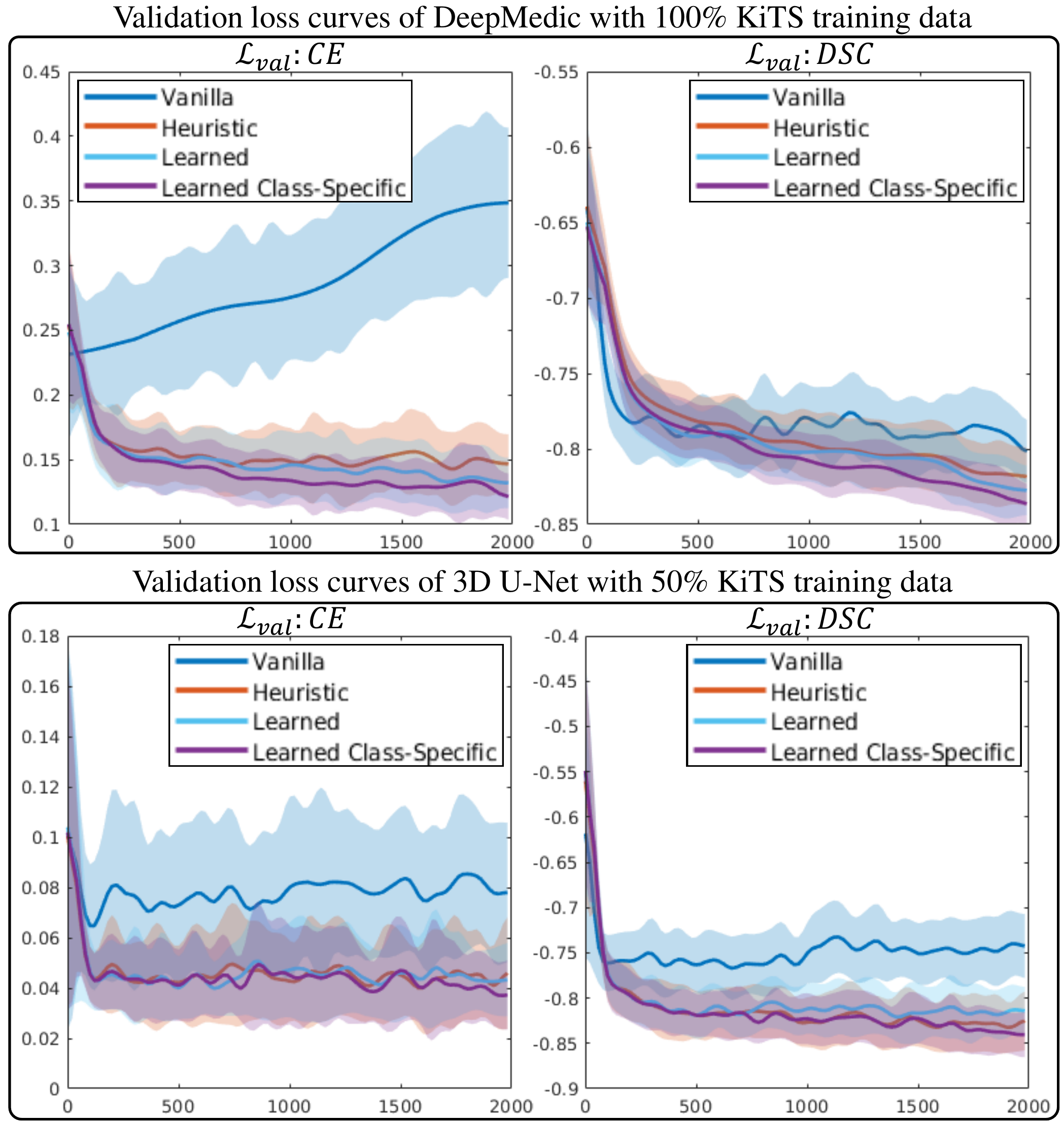}

\caption{Validation loss curves of different models trained for kidney and kidney tumor segmentation with KiTS. Compared with heuristic and learned class-agnostic data augmentation, the learned class-specific data augmentation can decrease the loss more. This may indicate that class-specific data augmentation is more effective in aligning the training and validation data distribution.} \label{fig_valloss}
\end{figure}

\begin{figure}[t]
\centering
\includegraphics[width=0.48\textwidth]{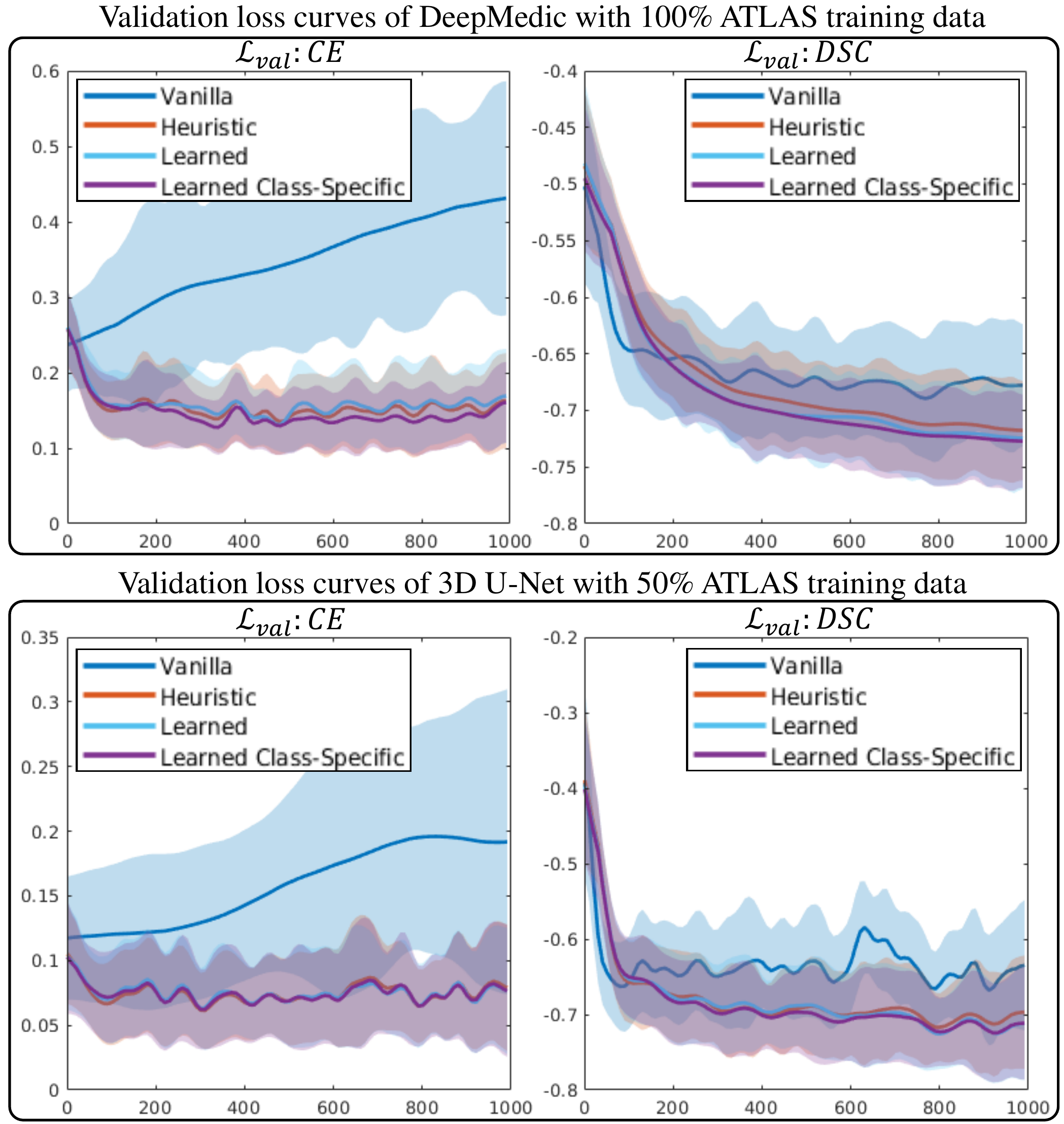}

\caption{Validation loss curves of different models trained for brain lesion segmentation with ATLAS. The learned class-specific data augmentation can decrease the loss more in comparison with the alternative methods.} \label{figa14}
\end{figure}

We visualize validation loss curves of four settings in Fig.~\ref{fig_valloss} and Fig.~\ref{figa14}. We find when the model is trained without TRA, it would be very likely to overfit the training dataset as we observe that the calculated CE of validation data would even increase during the training process, but not for DSC. This might indicate that the model becomes overconfident with the prediction of the hard cases.

Heuristic and learned TRA can help the model decrease the validation loss, while the proposed learned class-specific TRA is the most effective to decrease the validation loss. This indicates that our class-specific transformation model is more capable of mimicking the underlying data distribution and thus help the segmentation model generalize better. The validation loss curves can be utilized as a good indicator to assess the test performance of the trained models. In practice, we suggest the practitioners utilize the validation loss curves to choose the best TRA hyper-parameters for their settings.

\clearpage

\subsection{Segmentation based on a Vision Transformer}

In this study, we evaluate our data augmentation algorithms with convolutional neural network (CNN) based segmentation models including DeepMedic and 3D U-Net. Here we extend our experiments with a transformer based segmentation model, nnFormer~\cite{zhou2021nnformer, li2022transforming}. Similar to our previous experiments, we train nnFormer using 50\% training data from ATLAS and 50\% training data from KiTS. We choose a patch size of 64$\times$64$\times$64 for both applications. We initialize TRA for nnFormer using transformations with small magnitudes as we find large transformations could easily decrease the segmentation performance. We choose the same initialized TEA for nnFormer as 3D U-Net.

We summarize the quantitative results in Table~\ref{tabatlasvit} and Table~\ref{tabkitsvit}, separately. We find nnFormer perform worse than DeepMedic and 3D U-Net in both tasks. Similarly, our proposed data augmentation methods consistently improve the segmentation results with higher DSC.

\begin{table*}[t]
\centering
\caption{Evaluation of brain stroke lesion segmentation on ATLAS based on nnFormer with 50\% training data using different data augmentation methods. Best and second best results are in \textbf{bold}, with best also \underline{\textbf{underlined}}.}\label{tabatlasvit}
\newsavebox{\tableboxatlasvit}
\begin{lrbox}{\tableboxatlasvit}
\begin{tabular}{c|c|m{16mm}<{\centering}|m{5mm}<{\centering}|m{5mm}<{\centering}|m{5mm}<{\centering}|m{5mm}<{\centering}}
\hlineB{3}
\tabincell{c}{Training-time \\ augmentation} & \tabincell{c}{Test-time \\ augmentation} & 
\tabincell{c}{DSC \\ $\uparrow$} & \tabincell{c}{SEN \\ $\uparrow$} & \tabincell{c}{PRC \\ $\uparrow$} & \tabincell{c}{HD \\ $\downarrow$} & \tabincell{c}{AVG \\ Rank $\downarrow$} \\
\hlineB{1}
None & None & 54.6 & 58.8 & \underline{\textbf{62.2}} & \underline{\textbf{25.4}} & \textbf{2.5} \\
Heuristic & None & 57.0 & \underline{\textbf{67.1}} & 57.0 & 47.6 &3.0  \\
Learned~\cite{cubuk2018autoaugment, lim2019fast, li2020dada} & None & \textbf{58.2} & 65.2 & 58.0 & 45.1 & 2.8 \\
\textbf{Learned Class-Specific} & None & \underline{\textbf{58.3}} (+1.3)$^\sim$ & \textbf{65.8} & \textbf{61.4} & \textbf{35.4} & \textbf{\underline{1.8}} \\
\hline
Heuristic~\cite{kamnitsas2017efficient} & Heuristic~\cite{isensee2021nnu} & \textbf{60.5} & \underline{\textbf{65.9}} & 63.4 & 36.3 & 2.8 \\
\textbf{Learned Class-Specific} & Heuristic~\cite{isensee2021nnu} & 59.9 & 62.9 & 66.0 & \textbf{29.1} & 3.3 \\
\textbf{Learned Class-Specific} & Learned~\cite{shanmugam2020and, kim2020learning} & \textbf{60.5} & 63.5 & \underline{\textbf{67.6}} & \underline{\textbf{27.8}} & \textbf{\underline{1.8}} \\
\multicolumn{2}{c|}{\textbf{Joint Learned Class-Specific}} & \underline{\textbf{61.2}} (+0.7)$^\sim$ & \textbf{65.1} & \textbf{66.1} & 33.3 & \textbf{2.0} \\
\hline
\end{tabular}
\end{lrbox}
\scalebox{1}{\usebox{\tableboxatlasvit}}

{\raggedright \quad $^*p$-value $<$ 0.05; $^{**}p$-value $<$ 0.01; $^\sim p$-value $	\geq$ 0.05 (compared to Heuristic TRA w/o TEA or Heuristic TRA w/ Heuristic TEA) \par}

\end{table*}

\begin{table*}[t]
\centering
\caption{Evaluation of kidney and kidney tumor segmentation based on nnFormer with 50\% training data using different data augmentation methods. Best and second best results are in \textbf{bold}, with best also \underline{\textbf{underlined}}.}\label{tabkitsvit}
\newsavebox{\tableboxkitsvit}
\begin{lrbox}{\tableboxkitsvit}
\begin{tabular}{c|c|m{8mm}<{\centering}|m{8mm}<{\centering}|m{8mm}<{\centering}|m{8mm}<{\centering}|m{16mm}<{\centering}|m{8mm}<{\centering}|m{8mm}<{\centering}|m{8mm}<{\centering}|m{8mm}<{\centering}}
\hlineB{3}
\multirow{2}{*}{\tabincell{c}{Training-time \\ data augmentation}} & \multirow{2}{*}{\tabincell{c}{Test-time \\ data augmentation}} & \multicolumn{4}{c|}{Kidney} & \multicolumn{4}{c|}{Tumor} & \multirow{2}{*}{\tabincell{c}{AVG \\ Rank $\downarrow$}} \\
& & DSC $\uparrow$ & SEN $\uparrow$ & PRC $\uparrow$ & HD $\downarrow$ & DSC $\uparrow$ & SEN $\uparrow$ & PRC $\uparrow$ & HD $\downarrow$ &  \\ 
\hlineB{1}
None & None & 92.4 & 91.6 & 94.4 & 9.4 & 55.5 & 53.9 & 62.5 & 113.4 & 4.0 \\
Heuristic & None & \underline{\textbf{95.0}} & \underline{\textbf{95.4}} & \textbf{94.7} & 9.8 & \textbf{67.4} & \textbf{71.3} & 69.3 & \underline{\textbf{61.6}} & \textbf{2.0} \\
Learned~\cite{cubuk2018autoaugment, lim2019fast, li2020dada} & None & \textbf{94.9} & \textbf{95.2} & 94.6 & \textbf{7.8} & 67.0 & 68.1 & \underline{\textbf{71.2}} & 65.8 & 2.5 \\
\textbf{Learned Class-Specific} & None & \textbf{94.9} & 94.9 & \underline{\textbf{94.9}} & \underline{\textbf{7.5}} & \underline{\textbf{68.6}} (+1.2)$^\sim$ & \underline{\textbf{72.2}} & \textbf{69.9} & \textbf{64.5} & \textbf{\underline{1.5}} \\
\hline
Heuristic~\cite{kamnitsas2017efficient} & Heuristic~\cite{isensee2021nnu} & \textbf{94.6} & \textbf{94.4} & \textbf{95.0} & \textbf{5.7} & 68.2 & 69.7 & \textbf{73.8} & \underline{\textbf{40.8}} & \textbf{2.3} \\
\textbf{Learned Class-Specific} & Heuristic~\cite{isensee2021nnu} & 94.4 & 94.3 & 94.8 & 13.8 & 66.0 & 68.2 & 70.2 & 64.1 &3.8 \\
\textbf{Learned Class-Specific} & Learned~\cite{shanmugam2020and, kim2020learning} & \underline{\textbf{94.9}} & \underline{\textbf{95.0}} & 94.9 & 7.6 & \textbf{68.6} & \underline{\textbf{71.6}} & 71.5 & \textbf{54.4} & \textbf{\underline{2.0}} \\
\multicolumn{2}{c|}{\textbf{Joint Learned Class-Specific}} & \underline{\textbf{94.9}} & \textbf{94.4} & \underline{\textbf{95.4}} & \underline{\textbf{5.5}} & \underline{\textbf{69.6}} (+1.4)$^\sim$ & \textbf{70.2} & \underline{\textbf{75.4}} & 69.6 & \textbf{\underline{2.0}} \\
\hline
\end{tabular}
\end{lrbox}
\scalebox{1}{\usebox{\tableboxkitsvit}}

{\raggedright \quad $^*p$-value $<$ 0.05; $^{**}p$-value $<$ 0.01; $^\sim p$-value $	\geq$ 0.05 (compared to Heuristic TRA w/o TEA or Heuristic TRA w/ Heuristic TEA) \par}

\end{table*}

\subsection{Cross-Sequence Cardiac Segmentation}
\label{sec:seq-cardiac}

Here, we further evaluate the proposed algorithm for cross-sequence cardiac segmentation in MR images. In this experiment, we train segmentation models with short-axis cardiac MR images which are collected with different MRI sequences. We utilize 45 balanced steady-state free precession (bSSFP) MR images and 45 late gadolinium enhanced (LGE) MR images from~\cite{zhuang2022cardiac}. We resample all the MR images to an in-plane spacing of 1.25$\times$1.25 mm following~\cite{ouyang2021causality}. We report the segmentation performance of three cardiac structures including left ventricle (LV), myocardium (MYO) and right ventricle (RV). Similar to the setting of cross-site prostate segmentation, we investigate the application scenario where only a small portion of labelled data is available for the target domain (LGE MRI). We select 30 cases from bSSFP for training and 10 for testing. We select randomly select 1 case from LGE for validation and utilize the rest 44 for testing. As the cardiac MR images are highly anisotropic, we train a segmentation models based on 3D U-Net using a patch size of 128$\times$128$\times$8.

We summarize the segmentation results in Table~\ref{tabCMR}. The proposed data augmentation methods can improve the segmentation results when compared with heuristic policies for the target domain (LGE). As the training patches always contain foreground samples, the additional advantage of class-specific TRA is not significant.

\begin{table*}[t]
\centering
\caption{Evaluation of cross-sequence cardiac segmentation based on 3D U-Net using different data augmentation methods. Best and second best results are in \textbf{bold}, with best also \underline{\textbf{underlined}}.}\label{tabCMR}
\newsavebox{\tableboxCMR}
\begin{lrbox}{\tableboxCMR}
\begin{tabular}{c|c|m{8mm}<{\centering}|m{8mm}<{\centering}|m{8mm}<{\centering}|m{16mm}<{\centering}||m{8mm}<{\centering}|m{8mm}<{\centering}|m{8mm}<{\centering}|m{8mm}<{\centering}}
\hlineB{3}
bSSFP & bSSFP/LGE & \multicolumn{4}{c||}{LGE (DSC $\uparrow$)} & \multicolumn{4}{c}{bSSFP (DSC $\uparrow$)} \\
\multirow{2}{*}{\tabincell{c}{Training-time \\ data augmentation}} & \multirow{2}{*}{\tabincell{c}{Test-time \\ data augmentation}} & \multirow{2}{*}{\tabincell{c}{LV}} & \multirow{2}{*}{\tabincell{c}{MYO}} & \multirow{2}{*}{\tabincell{c}{RV}} & \multirow{2}{*}{\tabincell{c}{AVG}} & \multirow{2}{*}{\tabincell{c}{LV}} & \multirow{2}{*}{\tabincell{c}{MYO}} & \multirow{2}{*}{\tabincell{c}{RV}} & \multirow{2}{*}{\tabincell{c}{AVG}} \\
& & & & & & & & & \\
\hlineB{1}
None & None & 59.6 & 43.8 & 46.1 & 49.8 & 96.4 & 90.7 & 94.2 & 93.8 \\
Heuristic~\cite{kamnitsas2017efficient} & None & 82.0 & 68.1 & 72.2 & 74.1 & \textbf{97.2} & \underline{\textbf{92.4}} & 94.7 & \textbf{94.8} \\
Heuristic$^\ddagger$~\cite{kamnitsas2017efficient} & None & 85.1 & 74.3 & 77.4 & \textbf{78.9} & \underline{\textbf{97.3}} & 92.2 & \textbf{95.0} & \textbf{94.8} \\
Learned$^\ddagger$~\cite{cubuk2018autoaugment, lim2019fast, li2020dada} & None & \textbf{86.1} & \textbf{74.8} & \underline{\textbf{80.9}} & \underline{\textbf{80.6}} & \underline{\textbf{97.3}} & \textbf{92.3} & \underline{\textbf{95.1}} & \underline{\textbf{94.9}} \\
\textbf{Learned Class-Specific}$^\ddagger$ & None & \underline{\textbf{86.4}} & \underline{\textbf{75.6}} & \textbf{79.7} & \underline{\textbf{80.6}} (+1.7)$^{**}$ & 97.1 & 92.2 & \textbf{95.0} & \textbf{94.8} \\
\hline
Heuristic$^\ddagger$~\cite{kamnitsas2017efficient} & Heuristic~\cite{isensee2021nnu} & 86.4 & 76.6 & 81.7 & 81.6 & \textbf{97.5} & \underline{\textbf{93.0}} & \textbf{95.2} & \underline{\textbf{95.2}} \\
\textbf{Learned Class-Specific}$^\ddagger$ & Heuristic~\cite{isensee2021nnu} & 86.8 & 76.6 & 82.1 & 81.8 & \underline{\textbf{97.6}} & \textbf{92.7} & \underline{\textbf{95.3}} & \underline{\textbf{95.2}} \\
\textbf{Learned Class-Specific}$^\ddagger$ & Learned~\cite{shanmugam2020and, kim2020learning} & \textbf{87.2} & \textbf{76.9} & \textbf{82.2} & \textbf{82.1} & 97.3 & \textbf{92.7} & \underline{\textbf{95.3}} & \textbf{95.1} \\
\multicolumn{2}{c|}{\textbf{Joint Learned Class-Specific}$^\ddagger$} & \underline{\textbf{87.7}} & \underline{\textbf{77.0}} & \underline{\textbf{82.7}} & \underline{\textbf{82.5}} (+0.9)$^*$ & 97.3 & 92.3 & 94.8 & 94.8 \\
\hline
\end{tabular}
\end{lrbox}
\scalebox{1}{\usebox{\tableboxCMR}}

{\raggedright \quad $^*p$-value $<$ 0.05; $^{**}p$-value $<$ 0.01; $^\sim p$-value $	\geq$ 0.05 (compared to Heuristic$^\ddagger$ TRA w/o TEA or Heuristic$^\ddagger$ TRA w/ Heuristic TEA) \par}

{\raggedright \quad $^\ddagger$We train these models with both training data collected with bSSFP and validation data collected with LGE. \par}

\end{table*}

\subsection{Cross-Site Cardiac Segmentation}
\label{sec:seq-cardiac2}

Here, we further validate our method with cross-site cardiac segmentation where cardiac MR images are collected with 5 different sites using different scanners~\cite{campello2021multi}. This dataset contains totally available 345 cardiac short axis MR images. Each images are annotations at the end-diastolic (ED) and end-systolic (ES) phases including LV, MYO and RV. We resample all the images to 1.25$\times$1.25$\times$10 mm. Following the setting of the challenge~\cite{campello2021multi}, we utilize 175 cases from for training, 34 cases for validation and 136 for testing. In order to evaluate the generalization ability of the segmentation model when deployed on data with domain shifts, we include data which is collected with different sites from the training data in the validation and test datasets. Specifically, there are 10 cases and 40 cases collected from unseen test set in the validation and test dataset, separately. We encourage the readers to refer to the challenge paper for detailed experimental settings~\cite{campello2021multi}. Similar to the network settings in Section~\ref{sec:seq-cardiac}, we train a segmentation model based on 3D U-Net using a patch size of 128$\times$128$\times$8. 
We summarize the results in Table~\ref{tabsiteCMR}. We also compare our methods with top ranking methods in this challenge~\cite{full2021studying, zhang2021semi, ma2021histogram}. We take their results directly from the challenge report (results of vendor D in~\cite{campello2021multi}).
We observe that the proposed data augmentation strategies can improve the segmentation performance in different settings, outperforming other competitive solutions in the challenge on unseen site. We should note that all the top ranking methods are based on the same network architectures with us (nnU-Net) but utilize different hand-engineered TRA policies or normalization techniques. Therefore, the results further demonstrate that our method is superior than current heuristic data augmentation strategies.

\begin{table*}[t]
\centering
\caption{Evaluation of cross-site cardiac segmentation based on 3D U-Net using different data augmentation methods. Best and second best results are in \textbf{bold}, with best also \underline{\textbf{underlined}}.}\label{tabsiteCMR}
\newsavebox{\tableboxsiteCMR}
\begin{lrbox}{\tableboxsiteCMR}
\begin{tabular}{c|c|m{8mm}<{\centering}|m{8mm}<{\centering}|m{8mm}<{\centering}|m{16mm}<{\centering}||m{8mm}<{\centering}|m{8mm}<{\centering}|m{8mm}<{\centering}|m{16mm}<{\centering}}
\hlineB{3}
Source sites & Source/Unseen sites & \multicolumn{4}{c||}{Unseen site (DSC $\uparrow$)} & \multicolumn{4}{c}{Source sites (DSC $\uparrow$)} \\
\multirow{2}{*}{\tabincell{c}{Training-time \\ data augmentation}} & \multirow{2}{*}{\tabincell{c}{Test-time \\ data augmentation}} & \multirow{2}{*}{\tabincell{c}{LV}} & \multirow{2}{*}{\tabincell{c}{MYO}} & \multirow{2}{*}{\tabincell{c}{RV}} & \multirow{2}{*}{\tabincell{c}{AVG}} & \multirow{2}{*}{\tabincell{c}{LV}} & \multirow{2}{*}{\tabincell{c}{MYO}} & \multirow{2}{*}{\tabincell{c}{RV}} & \multirow{2}{*}{\tabincell{c}{AVG}} \\
& & & & & & & & & \\
\hlineB{1}
None & None & \underline{\textbf{90.6}} & 82.0 & 85.3 & 86.0 & 89.0 & 81.5 & \textbf{83.0} & 84.5 \\
Heuristic~\cite{kamnitsas2017efficient} & None & 90.4 & \textbf{82.7} & 87.6 & 86.9 & 90.7 & \textbf{84.0} & \textbf{87.4} & \textbf{87.4} \\
Learned~\cite{cubuk2018autoaugment, lim2019fast, li2020dada} & None & 90.4 & \underline{\textbf{82.8}} & \textbf{88.1} & \textbf{87.1} & \underline{\textbf{91.0}} & \underline{\textbf{84.1}} & \textbf{87.4} & \underline{\textbf{87.5}} \\
\textbf{Learned Class-Specific} & None & \textbf{90.5} & \textbf{82.7} & \underline{\textbf{88.5}} & \underline{\textbf{87.3}} (+0.4)$^*$ & \textbf{90.8} & \textbf{84.0} & \underline{\textbf{87.4}} & \textbf{87.4} (+0.0)$^\sim$ \\
\hline
Heuristic~\cite{kamnitsas2017efficient} & Heuristic~\cite{isensee2021nnu} & 90.8 & 83.5 & \textbf{88.8} & \textbf{87.7} & 91.0 & \textbf{84.6} & \textbf{87.8} & 87.8 \\
\textbf{Learned Class-Specific} & Heuristic~\cite{isensee2021nnu} & 90.6 & 83.2 & \underline{\textbf{89.2}} & \textbf{87.7} & \textbf{91.1} & \underline{\textbf{84.7}} & \underline{\textbf{88.0}} & \textbf{87.9} \\
\textbf{Learned Class-Specific} & Learned~\cite{shanmugam2020and, kim2020learning} & 90.5 & 83.2 & \underline{\textbf{89.2}} & \textbf{87.7} & \textbf{91.1} & \underline{\textbf{84.7}} & \underline{\textbf{88.0}} & \textbf{87.9} \\
Hand-engineered~\cite{full2021studying} (Top1) & Heuristic~\cite{isensee2021nnu} & \textbf{90.9} & \underline{\textbf{83.8}} & 88.2 & 87.6 & ------ & ------ & ------ & ------ \\
Hand-engineered~\cite{zhang2021semi} (Top2) & Heuristic~\cite{isensee2021nnu} & 90.3 & 82.7 & 87.1 & 86.7 & ------ & ------ & ------ & ------ \\
Hand-engineered~\cite{ma2021histogram} (Top3) & Heuristic~\cite{isensee2021nnu} & 89.8 & 82.4 & 87.0 & 86.4 & ------ & ------ & ------ & ------ \\
\multicolumn{2}{c|}{\textbf{Joint Learned Class-Specific}$^\ddagger$} & \underline{\textbf{91.1}} & \textbf{83.7} & \textbf{88.8} & \underline{\textbf{87.9}} (+0.2)$^\sim$ & \underline{\textbf{91.4}} & \underline{\textbf{84.7}} & \underline{\textbf{88.0}} & \underline{\textbf{88.0}} (+0.2)$^{**}$ \\
\hline
\end{tabular}
\end{lrbox}
\scalebox{1}{\usebox{\tableboxsiteCMR}}

{\raggedright \quad $^*p$-value $<$ 0.05; $^{**}p$-value $<$ 0.01; $^\sim p$-value $	\geq$ 0.05 (compared to Heuristic TRA w/o TEA or Heuristic TRA w/ Heuristic TEA) \par}

\end{table*}

\subsection{Sensitivity Analysis of the Size of Validation Dataset}

We optimize the data augmentation policies based on the model performance on a set of held-out dataset. In other words, we choose the TRA and TEA policies which can help the model perform well on this validation dataset. Here, we investigate the effects of validation data size and optimize the policies with varied amounts of validation data. Specifically, we optimize the joint learning of class-specific TRA and TEA based on DeepMedic with 50\% training data using different amounts the validation samples. 

We summarize the quantitative results in Table.~\ref{tabkitssens}. The results show that the proposed data augmentation framework is capable of improving the segmentation accuracy with different amounts of validation samples. 

Initially, the probability of specific transformations for TRA would be largely increased when we only reduce the size the validation data. As a result, the performance of the segmentation model is unstable. Specifically, with less validation samples, the segmentation model would achieve higher sensitivity and is prone to over-segmentation when trained using training data with large variance. This is probably because the learned policies would bias towards specific kinds of transformations which can benefit the segmentation of the small portion of validation data.  
Therefore, here we choose smaller learning rate for optimizing TRA $\beta$ when less validation data is available, in order to reduce risks of overfitting to specific transformations. In this way, we observe that the segmentation model can bring stable improvements. We suggest the practitioners also reduce $\beta$ with small validation data.

\begin{table*}[t]
\centering
\caption{Evaluation of kidney and kidney tumor segmentation based on DeepMedic with 50\% training data optimized using different amounts of validation samples. Best and second best results are in \textbf{bold}, with best also \underline{\textbf{underlined}}.}\label{tabkitssens}
\newsavebox{\tableboxkitssens}
\begin{lrbox}{\tableboxkitssens}
\begin{tabular}{c|c|m{14mm}<{\centering}|m{7mm}<{\centering}|m{7mm}<{\centering}|m{7mm}<{\centering}|m{7mm}<{\centering}|m{6mm}<{\centering}|m{16mm}<{\centering}|m{7mm}<{\centering}|m{7mm}<{\centering}|m{6mm}<{\centering}|m{7mm}<{\centering}}
\hlineB{3}
\multirow{3}{*}{\tabincell{c}{Training-time \\ data augmentation}} & \multirow{3}{*}{\tabincell{c}{Test-time \\ data augmentation}} & \multirow{3}{*}{\tabincell{c}{Validation \\ cases}} & \multirow{3}{*}{$\beta$} &  \multicolumn{4}{c|}{Kidney} & \multicolumn{4}{c|}{Tumor} & \multirow{2}{*}{\tabincell{c}{AVG \\ Rank \\ $\downarrow$ }} \\
& & & & DSC $\uparrow$ & SEN $\uparrow$ & PRC $\uparrow$ & HD $\downarrow$ & \tabincell{c}{DSC \\ $\uparrow$} & SEN $\uparrow$ & PRC $\uparrow$ & HD $\downarrow$ \\
\hlineB{1}
Heuristic~\cite{kamnitsas2017efficient} & Heuristic~\cite{isensee2021nnu} & ------ & ------ & 95.8 & 95.0 & \underline{\textbf{97.1}} & 11.0 & 70.5 & 70.5 & 78.5 & 58.7 & 4.8 \\
\multicolumn{2}{c|}{\textbf{Joint Learned Class-Specific}} & 28 (100\%) & 1e-3 & 95.8 & 94.9 & \textbf{97.0} & 11.0 & \underline{\textbf{73.3}} (+2.8)$^{**}$ & \textbf{73.5} & \underline{\textbf{79.7}} & \underline{\textbf{48.4}} & \textbf{\underline{1.3}} \\
\multicolumn{2}{c|}{\textbf{Joint Learned Class-Specific}} & 14 (50\%) & 5e-4 & \textbf{95.9} & \underline{\textbf{95.3}} & 96.8 & 9.7 & \textbf{73.2} (+2.7) & 72.5 & \textbf{79.3} & 55.8 & \textbf{2.8} \\
\multicolumn{2}{c|}{\textbf{Joint Learned Class-Specific}} & 6 (20\%) & 5e-4 & \underline{\textbf{96.1}} & \underline{\textbf{95.3}} & \underline{\textbf{97.1}} & \underline{\textbf{8.5}} & 73.0 (+2.5) & \underline{\textbf{75.0}} & 77.4 & 58.4 & 3.3 \\
\multicolumn{2}{c|}{\textbf{Joint Learned Class-Specific}} & 3 (10\%) & 5e-4 & 95.8 & \textbf{95.2} & 96.9 & \textbf{8.9} & 72.6 (+2.1)$^{*}$ & 73.1 & 79.1 & \textbf{52.9} & 3.0 \\
\hline
\end{tabular}
\end{lrbox}
\scalebox{0.95}{\usebox{\tableboxkitssens}}

{\raggedright \quad $^*p$-value $<$ 0.05; $^{**}p$-value $<$ 0.01; $^\sim p$-value $	\geq$ 0.05 (compared with Heuristic TRA w/ Heuristic TEA) \par}

\end{table*}

\subsection{Cross-Site Prostate Segmentation When Trained with Small Patches}

We found when the patches always contain foreground samples, the class-specific constraints could decrease the segmentation performance. This is what we observe in the experiments of cross-site prostate segmentation when training with patch size of 64$\times$64$\times$32. Here, we conduct experiments for cross-site prostate segmentation with smaller patches. Specifically, we train the segmentation models with patch sizes of 48$\times$48$\times$24 and keep the rest settings the same.

We summarize the segmentation results in Table~\ref{tabprostates}. The results show that the proposed class-specific TRA shows better results than heuristic policies and class-agnostic TRA. This demonstrates that class-specific TRA can work well with small patches when the regions of interest (ROIs) are relatively large.

However, we notice that the segmentation models trained with smaller patches would generally perform worse when compared with models trained with large patches. This is because the models trained with smaller patches have less context information but the local features cannot generalize well. Therefore, we remind the readers to strike a balance between large context information and class-specific constraints to achieve better performance when dealing with similar cases.

\begin{table*}[t]
\centering
\caption{Evaluation of cross-site prostate segmentation based on 3D U-Net size using different data augmentation methods. The models are trained and deployed using samples which are cropped with small patch size (48$\times$48$\times$24). Best and second best results are in \textbf{bold}, with best also \underline{\textbf{underlined}}.}\label{tabprostates}
\newsavebox{\tableboxprostates}
\begin{lrbox}{\tableboxprostates}
\begin{tabular}{c|c|m{16mm}<{\centering}|m{6mm}<{\centering}|m{6mm}<{\centering}|m{6mm}<{\centering}||m{6mm}<{\centering}|m{6mm}<{\centering}|m{6mm}<{\centering}|m{6mm}<{\centering}|m{6mm}<{\centering}}
\hlineB{3}
Site A & Site A/B & \multicolumn{4}{c||}{Site B} & \multicolumn{4}{c|}{Site A} & \multirow{3}{*}{\tabincell{c}{AVG \\ Rank \\ $\downarrow$}} \\
\multirow{2}{*}{\tabincell{c}{Training-time \\ data augmentation}} & \multirow{2}{*}{\tabincell{c}{Test-time \\ data augmentation}} & \multirow{2}{*}{\tabincell{c}{DSC \\ $\uparrow$}} & \multirow{2}{*}{\tabincell{c}{SEN \\ $\uparrow$}} & \multirow{2}{*}{\tabincell{c}{PRC \\ $\uparrow$}} & \multirow{2}{*}{\tabincell{c}{HD \\ $\downarrow$}} & \multirow{2}{*}{\tabincell{c}{DSC \\ $\uparrow$}} & \multirow{2}{*}{\tabincell{c}{SEN \\ $\uparrow$}} & \multirow{2}{*}{\tabincell{c}{PRC \\ $\uparrow$}} & \multirow{2}{*}{\tabincell{c}{HD \\ $\downarrow$}} \\
& & & & & & & & & \\
\hlineB{1}
None & None & 55.3 & 51.8 & \underline{\textbf{71.3}} & \underline{\textbf{22.6}} & \textbf{85.2} & 83.3 & \underline{\textbf{89.1}} & 40.7 & 3.0 \\
Heuristic~\cite{isensee2021nnu} & None & 58.1 & 81.4 & 46.9 & 87.6 & 82.9 & \textbf{89.1} & 78.7 & 42.7 & 4.5 \\
Heuristic$^\ddagger$~\cite{isensee2021nnu} & None & 67.9 & \textbf{86.3} & 57.9 & 56.3 & 83.5 & \underline{\textbf{89.3}} & 79.6 & \underline{\textbf{29.4}} & 2.8 \\
Learned$^\ddagger$~\cite{cubuk2018autoaugment, lim2019fast, li2020dada} & None & \textbf{68.3} & \underline{\textbf{86.9}} & 57.9 & 75.5 & 83.1 & 88.6 & 79.3 & 56.6 & \textbf{2.5} \\
\textbf{Learned Class-Specific}$^\ddagger$ & None & \underline{\textbf{74.5}} (+6.6)$^{**}$ & 83.8 & \textbf{68.9} & \textbf{46.1} & \underline{\textbf{86.9}} & 88.9 & \textbf{85.9} & \textbf{33.2} & \textbf{\underline{2.0}} \\
\hline
\end{tabular}
\end{lrbox}
\scalebox{1}{\usebox{\tableboxprostates}}

{\raggedright \quad $^*p$-value $<$ 0.05; $^{**}p$-value $<$ 0.01; $^\sim p$-value $	\geq$ 0.05 (compared to Heuristic$^\ddagger$ TRA w/o TEA) \par}

{\raggedright \quad $^\ddagger$We train these models with both training data from site A and validation data from site B. \par}

\end{table*}

\subsection{Cross-Validated Segmentation Results}

Most of our experiments choose fixed data split and report the model performance on a separate test set. Here, we extend the experiments with three-fold cross-validation and report the model performance on the whole dataset, to further validate our algorithms. Specifically, we train DeepMedic for kidney and kidney tumor segmentation with three different 70 cases and report the segmentation results on the rest data. We summarize the results in Table~\ref{tabkitsvitcv}, which are consistent with our previous experiments. Our proposed methods bring significant improvements to the segmentation of kidney tumor in terms of DSC. 

\begin{table*}[t]
\centering
\caption{Three-fold evaluation of kidney and kidney tumor segmentation based on DeepMedic with 50\% training data using different data augmentation methods. Best and second best results are in \textbf{bold}, with best also \underline{\textbf{underlined}}.}\label{tabkitsvitcv}
\newsavebox{\tableboxkitscv}
\begin{lrbox}{\tableboxkitscv}
\begin{tabular}{c|c|m{8mm}<{\centering}|m{8mm}<{\centering}|m{8mm}<{\centering}|m{8mm}<{\centering}|m{16mm}<{\centering}|m{8mm}<{\centering}|m{8mm}<{\centering}|m{8mm}<{\centering}|m{8mm}<{\centering}}
\hlineB{3}
\multirow{2}{*}{\tabincell{c}{Training-time \\ data augmentation}} & \multirow{2}{*}{\tabincell{c}{Test-time \\ data augmentation}} & \multicolumn{4}{c|}{Kidney} & \multicolumn{4}{c|}{Tumor} & \multirow{2}{*}{\tabincell{c}{AVG \\ Rank $\downarrow$}} \\
& & DSC $\uparrow$ & SEN $\uparrow$ & PRC $\uparrow$ & HD $\downarrow$ & DSC $\uparrow$ & SEN $\uparrow$ & PRC $\uparrow$ & HD $\downarrow$ \\
\hlineB{1}
None & None & 91.5 & 88.7 & \underline{\textbf{96.7}} & \underline{\textbf{10.4}} & 35.9 & 32.2 & 56.1 & 93.6 & 4.0 \\
Heuristic & None & \textbf{94.4} & \underline{\textbf{94.3}} & 95.3 & 18.8 & 63.4 & 66.2 & 70.6 & 73.7 & 3.0 \\
Learned~\cite{cubuk2018autoaugment, lim2019fast, li2020dada} & None & \underline{\textbf{94.5}} & \textbf{94.2} & \textbf{95.5} & 15.1 & \textbf{66.7} & \textbf{67.8} & \textbf{74.8} & \textbf{62.3} & \textbf{2.0} \\
\textbf{Learned Class-Specific} & None & 94.0 & 93.5 & \textbf{95.5} & \textbf{14.1} & \underline{\textbf{67.6}} (+4.2)$^{**}$ & \underline{\textbf{69.2}} & \underline{\textbf{75.0}} & \underline{\textbf{62.2}} & \textbf{\underline{1.0}} \\
\hline
Heuristic~\cite{kamnitsas2017efficient} & Heuristic~\cite{isensee2021nnu} & \underline{\textbf{94.7}} & \underline{\textbf{94.1}} & \underline{\textbf{96.2}} & 13.1 & 66.9 & 67.4 & 75.9 & 60.5 & 4.0 \\
\textbf{Learned Class-Specific} & Heuristic~\cite{isensee2021nnu} & \textbf{94.4} & 93.7 & \textbf{96.1} & \underline{\textbf{11.5}} & 69.2 & \underline{\textbf{69.4}} & \textbf{79.3} & \textbf{45.9} & \textbf{2.0} \\
\textbf{Learned Class-Specific} & Learned~\cite{shanmugam2020and, kim2020learning} & \textbf{94.4} & 93.7 & \textbf{96.1} & \underline{\textbf{11.5}} & \textbf{69.3} & \underline{\textbf{69.4}} & \underline{\textbf{79.5}} & \underline{\textbf{43.3}} & \textbf{\underline{1.3}} \\
\multicolumn{2}{c|}{\textbf{Joint Learned Class-Specific}} & \underline{\textbf{94.7}} & \textbf{94.0} & 96.1 & \textbf{11.9} & \underline{\textbf{69.6}} (+2.7)$^*$ & \textbf{69.0} & \textbf{79.3} & 46.7 & 2.3 \\
\hline
\end{tabular}
\end{lrbox}
\scalebox{1}{\usebox{\tableboxkitscv}}

{\raggedright \quad $^*p$-value $<$ 0.05; $^{**}p$-value $<$ 0.01; $^\sim p$-value $	\geq$ 0.05 (compared to Heuristic TRA w/o TEA or Heuristic TRA w/ Heuristic TEA) \par}

\end{table*}

\subsection{Segmentation Results with Post-Processing}

In this study, we investigate the problem of class imbalance in medical image segmentation and conduct experiments with datasets containing small objects. Specifically, the tasks of brain lesion and kidney tumor segmentation are challenging because the positions of these objects are quite random. The segmentation model could make false positive predictions which are far from the ground truth locations. As a result, the distance based evaluation metric, such as HD, is unstable to represent the quality of segmentation results. For example, HD would be large due to small positive predictions which are distant from the ROIs. In addition, the HD penalty of failing to make any predictions in a volume is large. This is the reason why our proposed methods do not always lead to the best HD in the experiments.
  
In practice, the false positive predictions could be easily eliminated with some simple post-processing techniques. We adopt a component-based post-processing approach where we only keep the largest component within the segmentation results but suppress the other predictions. We apply the post-processing approach to the segmentation results of ATLAS and KiTS based on 3D U-Net with 50\% training data and summarize the quantitative results in Table~\ref{tabatlaspost} and Table~\ref{tabkitspp}, separately. The results show that our methods always achieve both the best DSC and HD in all settings. 

We notice that this component-based post-processing approach would decrease the segmentation performance of brain lesion segmentation in terms of DSC, when compared to results in Table~\ref{tabatlas}. This is because the post-processing would introduce false negatives predictions when multiple brain lesions exist in an image. This is the reason why we do not utilize post-processing for all the experiments and is not the focus of this study. We think more advanced post-processing could be effective to improve segmentation results in that case.

\begin{table*}[t]
\centering
\caption{Evaluation of brain stroke lesion segmentation on ATLAS based on 3D U-Net with 50\% training data using different data augmentation methods. The results are calculated with post-processing. Best and second best results are in \textbf{bold}, with best also \underline{\textbf{underlined}}.}\label{tabatlaspost}
\newsavebox{\tableboxatlaspost}
\begin{lrbox}{\tableboxatlaspost}
\begin{tabular}{c|c|m{16mm}<{\centering}|m{5mm}<{\centering}|m{5mm}<{\centering}|m{16mm}<{\centering}|m{5mm}<{\centering}}
\hlineB{3}
\tabincell{c}{Training-time \\ augmentation} & \tabincell{c}{Test-time \\ augmentation} & 
\tabincell{c}{DSC \\ $\uparrow$} & \tabincell{c}{SEN \\ $\uparrow$} & \tabincell{c}{PRC \\ $\uparrow$} & \tabincell{c}{HD \\ $\downarrow$} & \tabincell{c}{AVG \\ Rank $\downarrow$} \\
\hlineB{1}
None & None & 52.3 & 48.9 & 69.5 & 32.8 &  3.8 \\
Heuristic & None & 57.8 & \textbf{57.1} & \textbf{69.8} & \textbf{24.0} & \textbf{2.3} \\
Learned~\cite{cubuk2018autoaugment, lim2019fast, li2020dada} & None & \textbf{58.0} & 55.8 & 68.1 & 25.5 & 3.0 \\
\textbf{Learned Class-Specific} & None & \underline{\textbf{61.2}} (+3.4)$^\sim$ & \underline{\textbf{58.7}} & \underline{\textbf{75.5}} & \underline{\textbf{21.6}} (-2.4)$^\sim$ & \textbf{\underline{1.0}} \\
\hline
Heuristic~\cite{kamnitsas2017efficient} & Heuristic~\cite{isensee2021nnu} & \textbf{60.6} & \textbf{59.0} & \textbf{73.0} & 23.3 & \textbf{2.5} \\
\textbf{Learned Class-Specific} & Heuristic~\cite{isensee2021nnu} & 59.2 & 58.0 & 71.5 & \textbf{22.8} & 2.8 \\
\textbf{Learned Class-Specific} & Learned~\cite{shanmugam2020and, kim2020learning} & 59.0 & 56.6 & 71.3 & \textbf{22.8} & 3.5 \\
\multicolumn{2}{c|}{\textbf{Joint Learned Class-Specific}} & \underline{\textbf{60.8}} (+0.2)$^\sim$ & \underline{\textbf{59.2}} & \underline{\textbf{73.6}} & \underline{\textbf{22.3}} (-0.5)$^\sim$ & \textbf{\underline{1.0}} \\
\hline
\end{tabular}
\end{lrbox}
\scalebox{1}{\usebox{\tableboxatlaspost}}

{\raggedright \quad $^*p$-value $<$ 0.05; $^{**}p$-value $<$ 0.01; $^\sim p$-value $	\geq$ 0.05 (compared to Heuristic TRA w/o TEA or Heuristic TRA w/ Heuristic TEA) \par}

\end{table*}

\begin{table*}[t]
\centering
\caption{Evaluation of kidney and kidney tumor segmentation based on 3D U-Net with 50\% training data using different data augmentation methods. The results are calculated with post-processing. Best and second best results are in \textbf{bold}, with best also \underline{\textbf{underlined}}.}\label{tabkitspp}
\newsavebox{\tableboxkitspp}
\begin{lrbox}{\tableboxkitspp}
\begin{tabular}{c|c|m{8mm}<{\centering}|m{8mm}<{\centering}|m{8mm}<{\centering}|m{8mm}<{\centering}|m{16mm}<{\centering}|m{8mm}<{\centering}|m{8mm}<{\centering}|m{16mm}<{\centering}|m{8mm}<{\centering}}
\hlineB{3}
\multirow{2}{*}{\tabincell{c}{Training-time \\ data augmentation}} & \multirow{2}{*}{\tabincell{c}{Test-time \\ data augmentation}} & \multicolumn{4}{c|}{Kidney} & \multicolumn{4}{c|}{Tumor} & \multirow{2}{*}{\tabincell{c}{AVG \\ Rank $\downarrow$}} \\
& & DSC $\uparrow$ & SEN $\uparrow$ & PRC $\uparrow$ & HD $\downarrow$ & DSC $\uparrow$ & SEN $\uparrow$ & PRC $\uparrow$ & HD $\downarrow$ \\
\hlineB{1}
None & None & 95.2 & 93.7 & \underline{\textbf{97.3}} & 5.8 & 43.9 & 39.1 & 62.3 & 98.8 & 4.0 \\
Heuristic & None & \textbf{96.5} & \textbf{96.1} & 96.9 & \textbf{2.7} & 75.0 & 74.6 & 78.1 & 20.6 & 3.0  \\
Learned~\cite{cubuk2018autoaugment, lim2019fast, li2020dada} & None & 96.4 & 95.9 & 96.9 & 2.8 & \textbf{78.9} & \textbf{80.4} & \textbf{80.3} & \textbf{17.4} & \textbf{2.0} \\
\textbf{Learned Class-Specific} & None & \underline{\textbf{96.6}} & \underline{\textbf{96.3}} & \textbf{97.0} & \underline{\textbf{2.5}} & \underline{\textbf{80.0}} (+5.0)$^*$ & \underline{\textbf{80.6}} & \underline{\textbf{81.5}} & \underline{\textbf{12.8}} (-7.8)$^\sim$ & \textbf{\underline{1.0}} \\
\hline
Heuristic~\cite{kamnitsas2017efficient} & Heuristic~\cite{isensee2021nnu} & \textbf{96.8} & \textbf{96.3} & \underline{\textbf{97.2}} & \textbf{2.4} & 78.2 & 78.3 & 80.5 & 16.5 & 4.0 \\
\textbf{Learned Class-Specific} & Heuristic~\cite{isensee2021nnu} & 96.6 & 96.1 & \textbf{97.1} & 2.6 & \textbf{79.9} & \textbf{80.1} & \textbf{82.2} & \textbf{12.9} & \textbf{2.0} \\
\textbf{Learned Class-Specific} & Learned~\cite{shanmugam2020and, kim2020learning} & 96.6 & 96.1 & \textbf{97.1} & 2.6 & \textbf{79.9} & \textbf{80.1} & \textbf{82.2} & \textbf{12.9} & \textbf{2.0} \\
\multicolumn{2}{c|}{\textbf{Joint Learned Class-Specific}} & \underline{\textbf{96.9}} & \underline{\textbf{96.6}} & \underline{\textbf{97.2}} & \underline{\textbf{2.3}} & \underline{\textbf{80.5}} (+2.3)$^\sim$ & \underline{\textbf{80.6}} & \underline{\textbf{82.5}} & \underline{\textbf{12.8}} (-3.7)$^\sim$ & \textbf{\underline{1.0}} \\
\hline
\end{tabular}
\end{lrbox}
\scalebox{0.97}{\usebox{\tableboxkitspp}}

{\raggedright \quad $^*p$-value $<$ 0.05; $^{**}p$-value $<$ 0.01; $^\sim p$-value $	\geq$ 0.05 (compared to Heuristic TRA w/o TEA or Heuristic TRA w/ Heuristic TEA) \par}

\end{table*}

\subsection{Domain Generalized Prostate Segmentation}

To further validate the generalization of the presented learning scheme, we test our learned models on test data which is collected from unseen domains, as an extension of the 3D U-Net results in Table.~\ref{tabprostate}. We train a 3D U-Net model with data from site A~\cite{bloch2015nci} and optimize the data augmentation policies using validation data from site B~\cite{lemaitre2015computer} following the settings in Section~\ref{sec:experiments} for prostate segmentation. Then we test the models on unseen data which is collected from other unseen data sources. Specifically, we test data on totally 67 unseen samples including 30 cases from~\cite{bloch2015nci}, 13, 12 and 12 cases from ~\cite{litjens2014evaluation}. We resample all the images to a voxel spacing of 0.8$\times$0.8$\times$1.5 mm. We encourage the readers to refer to~\cite{liu2020saml} for more details. We summarize the results in Table.~\ref{tabprostatedg}. We find that the proposed methods can work well when test data has different appearance from the training and validation datasets. This indicates that the learning scheme of our methods is robust to domain shifts.

\begin{table*}[t]
\centering
\caption{Evaluation of prostate segmentation with unseen data based on 3D U-Net using different data augmentation methods. Best and second best results are in \textbf{bold}, with the best also \underline{\textbf{underlined}}.}\label{tabprostatedg}
\newsavebox{\tableboxprostatedg}
\begin{lrbox}{\tableboxprostatedg}
\begin{tabular}{c|c|m{16mm}<{\centering}|m{5mm}<{\centering}|m{5mm}<{\centering}|m{8mm}<{\centering}|m{5mm}<{\centering}}
\hlineB{3}
\tabincell{c}{Training-time \\ augmentation} & \tabincell{c}{Test-time \\ augmentation} & 
\tabincell{c}{DSC \\ $\uparrow$} & \tabincell{c}{SEN \\ $\uparrow$} & \tabincell{c}{PRC \\ $\uparrow$} & \tabincell{c}{HD \\ $\downarrow$} & \tabincell{c}{AVG \\ Rank $\downarrow$} \\
\hlineB{1}
None & None & 68.3 & 59.6 & \textbf{\underline{89.2}} & \textbf{\underline{14.5}} & 2.5 \\
Heuristic$^\ddagger$~\cite{kamnitsas2017efficient} & None & 75.8 & 75.5 & 71.5 & 63.4 & 3.5 \\
Learned$^\ddagger$~\cite{cubuk2018autoaugment, lim2019fast, li2020dada} & None & \textbf{79.7} & \textbf{\underline{86.2}} & 76.5 & 57.9 & \textbf{2.3} \\
\textbf{Learned Class-Specific}$^\ddagger$ & None & \textbf{\underline{80.8}} (+5.0)$^{**}$ & 84.7 & \textbf{79.4} & \textbf{52.8} & \textbf{\underline{1.8}} \\
\hline
Heuristic$^\ddagger$~\cite{kamnitsas2017efficient} & Heuristic$^\ddagger$~\cite{isensee2021nnu} & 81.0 & \textbf{85.8} & 79.3 & 39.8 & 3.3 \\
Learned$^\ddagger$~\cite{cubuk2018autoaugment, lim2019fast, li2020dada} & Heuristic~\cite{isensee2021nnu} & \textbf{83.4} & \textbf{\underline{86.6}} & \textbf{82.2} & \textbf{35.8} & \textbf{\underline{1.8}} \\
Learned$^\ddagger$~\cite{cubuk2018autoaugment, lim2019fast, li2020dada} & Learned~\cite{shanmugam2020and, kim2020learning} & 83.3 & \textbf{\underline{86.6}} & 82.0 & 37.4 & \textbf{2.8} \\
\multicolumn{2}{c|}{\textbf{Joint Learned}$^\ddagger$} & \textbf{\underline{84.3}} (+3.3)$^{**}$ & 85.5 & \textbf{\underline{84.3}} & \textbf{\underline{27.0}} & \textbf{\underline{1.8}} \\
\hline
\end{tabular}
\end{lrbox}
\scalebox{1}{\usebox{\tableboxprostatedg}}

{\raggedright \quad $^*p$-value $<$ 0.05; $^{**}p$-value $<$ 0.01; $^\sim p$-value $	\geq$ 0.05 (compared to Heuristic$^\ddagger$ TRA w/o TEA or Heuristic$^\ddagger$ TRA w/ Heuristic TEA) \par}

{\raggedright \quad $^\ddagger$We train these models with both training data from site A and validation data from site B. \par}

\end{table*}

\bibliographystyle{ieee}
\bibliography{reference}


\end{document}